\documentclass[10pt,twocolumn,letterpaper]{article}

\usepackage[pagebackref=true,breaklinks=true,letterpaper=true,colorlinks,bookmarks=false]{hyperref}
\usepackage{cvpr}
\usepackage{times}
\usepackage{epsfig}
\usepackage{graphicx}
\usepackage{amsmath}
\usepackage{amssymb}
\usepackage{graphicx}  \graphicspath{{figures/}}
\usepackage[normalem]{ulem}
\usepackage{makecell}
\usepackage{rotating}
\usepackage{booktabs}
\usepackage{gensymb}
\usepackage{tabularx}
\usepackage{enumitem}
\usepackage[ruled,vlined,linesnumbered]{algorithm2e}
\usepackage[dvipsnames,svgnames,x11names]{xcolor}
\usepackage{subfig}
\usepackage[toc,page]{appendix}

\cvprfinalcopy 

\def\cvprPaperID{705} 

\newcolumntype{Y}{>{\centering\arraybackslash}X}

\ifcvprfinal\fi
\begin{document}
	
	\title{NRMVS: Non-Rigid Multi-View Stereo}
	
	\def\authorspace{\qquad}
	\def\affilspace{\quad}
	\author{		
		Matthias Innmann$^{1,2}$\footnotemark[1] \authorspace 
		Kihwan Kim$^1$ \authorspace 
		Jinwei Gu$^{1,3}$\footnotemark[1] \authorspace 
		Matthias Nie{\ss}ner$^{4}$ \authorspace 
		Charles Loop$^1$ \\ 
		Marc Stamminger$^{2}$ \authorspace 
		Jan Kautz$^1$ \\ \\
		$^1$NVIDIA \affilspace 
		$^2$University of Erlangen-Nuremberg \affilspace
		$^3$SenseTime \affilspace
		$^4$Technical University of Munich 
	}	
	\maketitle
	
	
	\begin{abstract}
		Scene reconstruction from unorganized RGB images is an important task in many computer vision applications.
		Multi-view Stereo (MVS) is a common solution in photogrammetry applications for the dense reconstruction of a static scene. 
		The static scene assumption, however, limits the general applicability of MVS algorithms, as many day-to-day scenes undergo non-rigid motion, e.g., clothes, faces, or human bodies. 
		In this paper, we open up a new challenging direction: dense 3D reconstruction of scenes with non-rigid changes observed from arbitrary, sparse, and wide-baseline views. We formulate the problem as a joint optimization of deformation and depth estimation, using deformation graphs as the underlying representation.
		We propose a new sparse 3D to 2D matching technique, together with a dense patch-match evaluation scheme to estimate deformation and depth with photometric consistency. 
		We show that creating a dense 4D structure from a few RGB images with non-rigid changes is possible, and demonstrate that our method can be used to interpolate novel deformed scenes from various combinations of these deformation estimates derived from the sparse views.
	\end{abstract}
	
	\begin{figure}[t]
		\centering
		\includegraphics[width=.99\linewidth]{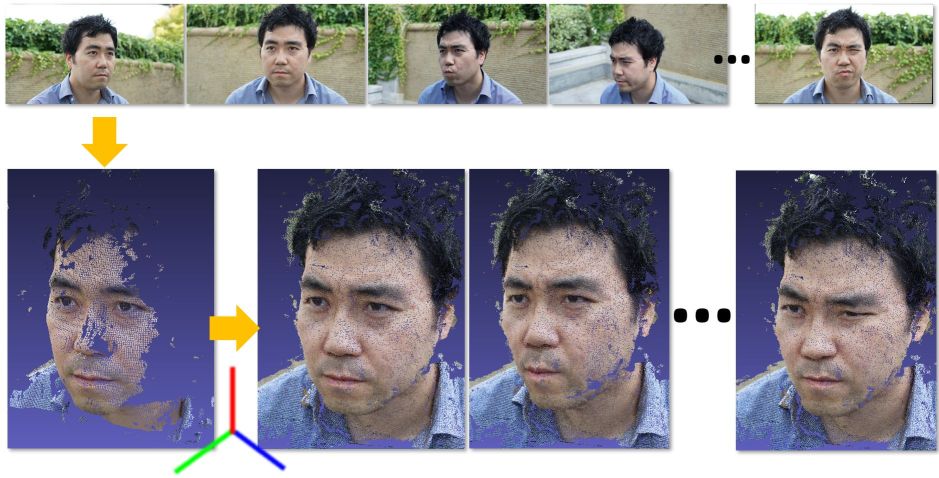}
		\parbox[h]{.24\columnwidth}{\centering \scriptsize (a) Canonical view}
		\parbox[h]{.24\columnwidth}{\centering \scriptsize (b) Updated canonical view}
		\parbox[h]{.48\columnwidth}{\centering \scriptsize (c) 3D point cloud for each view}
		\caption{We take a small number of unordered input images captured from wide-baseline views (top row), and reconstruct the 3D geometry of objects undergoing non-rigid deformation. We first triangulate a canonical surface from a pair of views with minimal deformation (a). Then we compute the deformation from the canonical view to the other views, and reconstruct point clouds for both original canonical views (b) and other remaining views (c)}
		\label{fig:teaser}
	\end{figure}

	\begin{figure*}
		\centering
		\includegraphics[width=.95\linewidth]{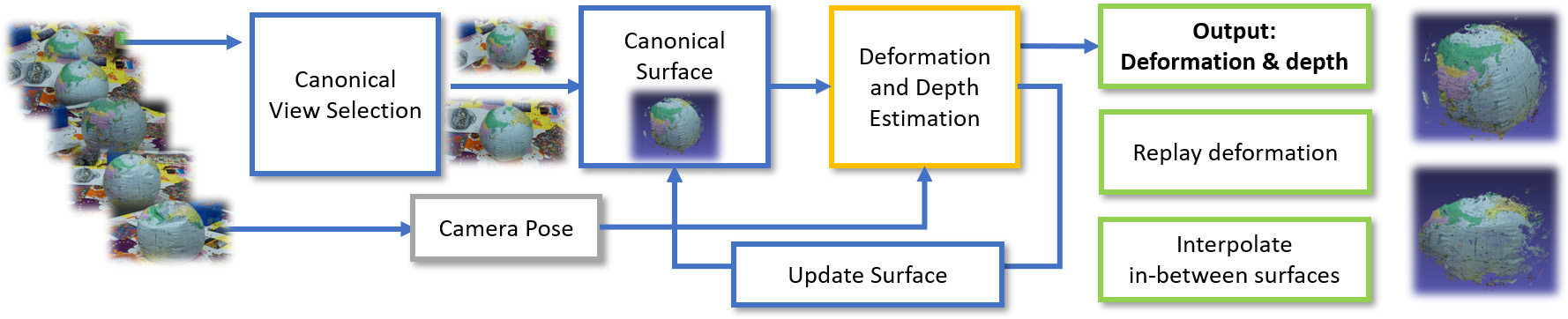}
		\caption{\textbf{Overview of our framework}: We first reconstruct the initial point cloud from the views with minimal deformation, which we call canonical views. Then, we estimate depths for the remaining views by estimating a plausible deformation from a joint optimization between depth, deformation, and dense photometric consistency. With these computed depths and deformations, we also demonstrate an interpolation  deformation between input views.}
		\label{fig:overview}
	\end{figure*}

	\section{Introduction}
	\footnotetext[1]{The authors contributed to this work when they were at NVIDIA.}
	
	Multi-view stereo algorithms \cite{Bleyer11bmvc, Campbell08eccv, Furukawa07cvpr, Galliani15iccv, schoenberger2016mvs, Seitz2006mvs} have played an important role in 3D reconstruction and scene understanding for applications such as augmented reality, robotics, and autonomous driving. However, if a scene contains motion, such as non-stationary rigid objects or non-rigid surface deformations, the assumption of an epipolar constraint is violated~\cite{Hartley04book}, causing algorithms to fail in most cases. 
	Scenes with rigidly-moving objects have been reconstructed by segmenting foreground objects from the background, and treating these regions independently \cite{Zhang11-multibody, MulVSLAM_Abhijit_ICCV2011, Wang15}. However, the reconstruction of scenes with surfaces undergoing deformation is still a challenging task.
	
	To solve this problem for sparse point pairs, various non-rigid structure from motion (NRSfM) methods have been introduced~\cite{Jensen18}. 
	These methods often require either dense views (video frames)~\cite{Ansari17} for the acquisition of dense correspondences with flow estimation, or prior information to constrain the problem~\cite{Dai17}.
	Newcombe et al.~\cite{Newcombe15cvpr} and Innmann et al.~\cite{Innmann16eccv} recently demonstrated solutions for the 3D reconstruction of arbitrary, non-rigid, dynamic scenes using a dense stream of known metric depths captured from a commercial depth camera. 
	However, there are many common scenarios one may encounter for which this method cannot be used, for example, when capturing scenes that contain non-rigid changes that are neither acquired as video nor captured by stereo or depth cameras, but rather come from independent single views.
	
	In this paper, we are specifically interested in \emph{dense 3D scene reconstruction with dynamic non-rigid deformations} acquired from images with wide spatial baselines, and sparse, unordered samples in time. This requires two solutions: (1) a method to compute the most plausible deformation from millions of potential deformations between given wide-baseline views, and (2) a dense surface reconstruction algorithm that satisfies a photometric consistency constraint between images of surfaces undergoing non-rigid changes.
	
	In our solution, we first compute a \emph{canonical surface} from views that have minimal non-rigid changes (Fig.~\ref{fig:teaser}(a)), and then estimate the deformation between the canonical pose and other views by joint optimization of depth and photometric consistency. This allows the expansion of 3D points of canonical views (Fig.~\ref{fig:teaser}(b)). 
	Then, through the individual deformation fields estimated from each view to the canonical surface, we can reconstruct a dense 3D point cloud of each single view  (Fig.~\ref{fig:teaser}(c)).
	A brief overview of our entire framework is described in Fig.~\ref{fig:overview}.\\
	
	\noindent Our contributions are as follows:
	\begin{itemize}[topsep=0pt,noitemsep]
		\item The first non-rigid MVS pipeline that densely reconstructs dynamic 3D scenes with non-rigid changes from wide-baseline and sparse RGB views.
		\item A new formulation to model non-rigid motion using a deformation graph~\cite{sumner2007embedded} and the approximation of the inverse-deformation used for the joint optimization with photometric consistency.
		\item Patchmatch-based~\cite{Bleyer11bmvc} dense sample propagation on top of an existing MVS pipeline~\cite{schoenberger2016mvs}, which allows flexible implementation depending on different MVS architectures.
	\end{itemize}

	\section{Related Work}
	\label{sec:related}
	
	\noindent \textbf{Dynamic RGB-D Scene Reconstruction.}
	A prior step to full dynamic scene reconstruction is dynamic template tracking of 3D surfaces.
	The main idea is to track a shape template over time while non-rigidly deforming its surface \cite{deAguiar2008,allain2015efficient,l0Norigid2015,hernandez2007non,li2008global,li2009robust,li2012temporally,gall2008driftfree,zollhoefer2014deformable}.
	Jointly tracking and reconstructing a non-rigid surface is significantly more challenging.
	In this context, researchers have developed an impressive line of works based on RGB-D or depth-only input~\cite{zeng2013templateless,mitra2007dynamic,tevs2012animation,bojsen2012tracking,dou2013scanning,Dou_2015_CVPR,Malleson2014,wang2016capturing,Li13}.
	DynamicFusion~\cite{Newcombe15cvpr} jointly optimizes a Deformation Graph \cite{sumner2007embedded}, then fuses the deformed surface with the current depth map.
	Innmann et al.~\cite{Innmann16eccv} follows up on this work by using an as-rigid-as-possible regularizer to represent deformations~\cite{sorkine2007rigid}, and incorporate RGB features in addition to a dense depth tracking term.
	Fusion4D~\cite{dou2016fusion4d} brings these ideas a level further by incorporating a high-end RGB-D capture setup, which achieves very impressive results.
	More recent RGB-D non-rigid fusion frameworks include KillingFusion~\cite{slavcheva2017killingfusion} and 
	SobolevFusion~\cite{slavcheva2018sobolevfusion}, which allow for implicit topology changes using advanced regularization techniques.
	This line of research has made tremendous progress in the recent years; but given the difficulty of the problem, all these methods either rely on depth data or calibrated multi-camera rigs.
	
	\noindent \textbf{Non-Rigid Structure from Motion.} 
	Since the classic structure from motion solutions tend to work well for many real world applications~\cite{Triggs:1996, Kanade1153, Tomasi92}, many recent efforts have been devoted to computing the 4D structure of sparse points in the spatio-temporal domain, which we call non-rigid structure from motion (NRSfM)~\cite{Jensen18, kumar17, Rabaud08, garg2013dense}. 
	However, most of the NRSfM methods consider the optimization of sparse points rather than a dense reconstruction, and often require video frames for dense correspondences~\cite{Ansari17} or prior information~\cite{Dai17}. 
	Scenes with rigidly moving objects (e.g., cars or chairs) have been reconstructed by segmenting rigidly moving regions~\cite{Zhang11-multibody, MulVSLAM_Abhijit_ICCV2011, Ladick10, Wang15}.
	In our work, we focus on a new scenario of reconstructing scenes with non-rigid changes from a few images, and estimate deformations that satisfy each view.
	
	\noindent \textbf{Multi-View Stereo and Dense Reconstruction.} 
	Various MVS approaches for dense 3D scene reconstruction have been introduced in the last few decades~\cite{Furukawa07cvpr, Galliani15iccv, schoenberger2016mvs, Seitz2006mvs}. 
	While many of these methods work well for static scenes, they often reject regions that are not consistent with the epipolar geometry~\cite{Hartley04book}, e.g., if the scene contains changing regions. 
	Reconstruction failure can also occur if the ratio of static to non-rigid parts present in the scene is too low~\cite{Lv18eccv}. 
	A recent survey~\cite{Schoeps2017CVPR} on MVS shows that COLMAP~\cite{schoenberger2016mvs} performs the best among state-of-the-art methods. Therefore, we adopt COLMAP's Patchmatch framework for dense photometric consistency.

	\section{Approach}
	The input to our non-rigid MVS method is a set of images of a scene taken from unique (wide-baseline) locations at different times. An overview of our method is shown in Fig.~\ref{fig:overview}.
	We do not assume any knowledge of temporal order, i.e., the images are an unorganized collection. 
	However, we assume there are at least two images with minimal deformation, and the scene contains sufficient background in order to measure the ratio of non-rigidity and to recover the camera poses. We discuss the details later in Sec~\ref{sec:implementation}.
	The output of our method is an estimate of the deformation within the scene from the canonical pose to every other view, as well as a depth map for each view.
	After the canonical view selection, we reconstruct an initial canonical surface that serves as a template for the optimization.
	Given another arbitrary input image and its camera pose, we estimate the deformation between the canonical surface and the input.
	Furthermore, we compute a depth map for this processed frame using a non-rigid variant of PatchMatch.
	Having estimated the motion and the geometry for every input image, we recompute the depth for the entire set of images to maximize the growth of the canonical surface.
	
	\subsection{Modeling Deformation in Sparse Observations}
	\label{sec:deformation_model}
	
	To model the non-rigid motion in our scenario, we use the well known concept of deformation graphs \cite{sumner2007embedded}.  Each graph node represents a rigid body transform, similar to the as-rigid-as-possible deformation model \cite{sorkine2007rigid}. These transforms are locally blended to deform nearby space.
	
	Given a point $\mathbf{v} \in \mathbb{R}^3$, the deformed version $\hat{\mathbf{v}}$ of the point is computed as:
	\[
	\hat{\mathbf{v}} = \sum_{i=1}^k w_i(\mathbf{v}) \left[ \mathbf{R}_i (\mathbf{v} - \mathbf{g}_i) + \mathbf{g}_i + \mathbf{t}_i \right],
	\]
	where $\mathbf{R}_i$ and $\mathbf{t}_i$ represent the rotation and translation of a rigid body transform about position $\mathbf{g}_i$ of the $i$-nearest deformation node, and $k$ is the user-specified number of nearest neighbor nodes (we set to $k = 4$ throughout our paper).
	The weights $w_i$ are defined as:
	\[
	w_i(\mathbf{v}) = \frac{1}{\sum_{j=1}^k w_j (\mathbf{v})} \left( 1 - \frac{\|\mathbf{v} - \mathbf{g}_i \|_2}{\| \mathbf{v} - \mathbf{g}_{k+1} \|_2}\right)^2.
	\]
	For a complete description of deformation graphs, we refer to the original literature~\cite{sumner2007embedded}. 
	
	When projecting points between different images, we also need to invert the deformation. 
	The exact inverse deformation can be derived given known weights:
	\[
	\mathbf{v} = \left(\sum_{i=1}^k w_i(\mathbf{v}) \mathbf{R}_i \right)^{-1} \left[ \hat{\mathbf{v}} + \sum_{i=1}^k w_i(\mathbf{v}) \left[ \mathbf{R}_i \mathbf{g}_i - \mathbf{g}_i - \mathbf{t}_i \right] \right]
	\]
	However, because we do not know the weights a priori, which requires the nearest neighbor nodes and their distances, this becomes a non-linear problem. 
	Since this computationally expensive step is necessary at many stages of our pipeline, we introduce an approximate solution:
	\[
	\mathbf{v} \approx \left(\sum_{i=1}^k \hat{w}_i(\mathbf{\hat v}) \mathbf{R}_i \right)^{-1} \left[ \hat{\mathbf{v}} + \sum_{i=1}^k \hat{w}_i(\mathbf{\hat v}) \left[ \mathbf{R}_i \mathbf{g}_i - \mathbf{g}_i - \mathbf{t}_i \right] \right],
	\]
	where the weights $\hat{w}_i$ are given by
	\[
	\hat{w}_i(\hat{\mathbf{v}}) = \frac{1}{\sum_{j=1}^k \hat{w}_i (\hat{\mathbf{v}})} \left( 1 - \frac{\|\mathbf{\hat v} - (\mathbf{g}_i + \mathbf{t}_i) \|_2}{\|\mathbf{\hat v} - (\mathbf{g}_{k+1} + \mathbf{t}_{k+1}) \|_2}\right)^2.
	\]
	Note that our approximation can be computed directly and efficiently, without leading to any error of observable influence in our synthetic experiments. 
	
	\begin{figure}[t]
		\centering
		\includegraphics[width=.9\linewidth]{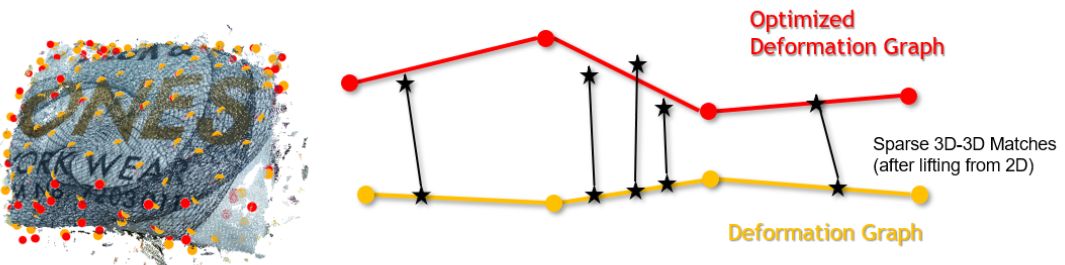}
		\caption{\textbf{Deformation nodes and correspondences:} (Left) shows the deformation nodes at $t_0$ (orange), and another set of nodes at $t_1$ (red) overlaid in the canonical view. (Right) we show the relationship between deformation nodes from two views and sparse 3D matches (after lifting) in the context of a non-rigid change. Note that we only show the sparse matching for simpler visualization while there is also a dense term for photometric consistency that drives the displacement of deformation nodes together with the sparse matches.}
		\label{fig:node-viz}
		\vspace{-.2cm}
	\end{figure}
	
	\subsection{Non-rigid Photometric Consistency and Joint Optimization}
	\label{sec:photometric-conssitency}
	With the deformation model in hand, we next estimate the depth of the other views by estimating deformations that are photometrically consistent with the collection images  and subject to  constraints on the geometry. 
	This entire step can be interpreted as a non-rigid version of a multi-view stereo framework.\\
	\noindent \textbf{Canonical View Selection} 
	From the set of input images, we select two views with a minimal amount of deformation.
	We run COLMAP's implementation of PatchMatch \cite{schoenberger2016mvs} 
	to acquire an initial temple model of the canonical pose.
	Based on this template, we compute the deformation graph by distributing a user-specified number of nodes on the surface.
	To this end, we start with all points of the point cloud as initial nodes.
	We iterate over all nodes, and for each node remove all its neighbors within a given radius.
	The process is repeated with a radius that is increased by 10\%, until we have reached the desired number of nodes.
	In our experiments, we found that 100 to 200 nodes are sufficient to faithfully reconstruct the motion. Fig.~\ref{fig:node-viz}(left) shows an example of the node distribution.\\
	\noindent \textbf{Correspondence Association}
	For sparse global correspondences, we detect SIFT keypoints \cite{lowe1999object} in each image and match descriptors for every pair of images to compute a set of feature tracks $\{ \mathbf{u}_i \}$. A \textit{feature track} represents the same 3D point and is computed by connecting each keypoint with each of its matches. We reject inconsistent tracks, i.e.,~if there is a path from a keypoint $\mathbf{u}^{(j)}_i$ in image $i$ to a different keypoint $\mathbf{u}^{(k)}_i$ with $j \neq k$ in the same image. 
	
\begin{figure}[t]
		\def\size{0.42\linewidth}
		\centering
		\subfloat[Iteration 1]{\includegraphics[width=\size]{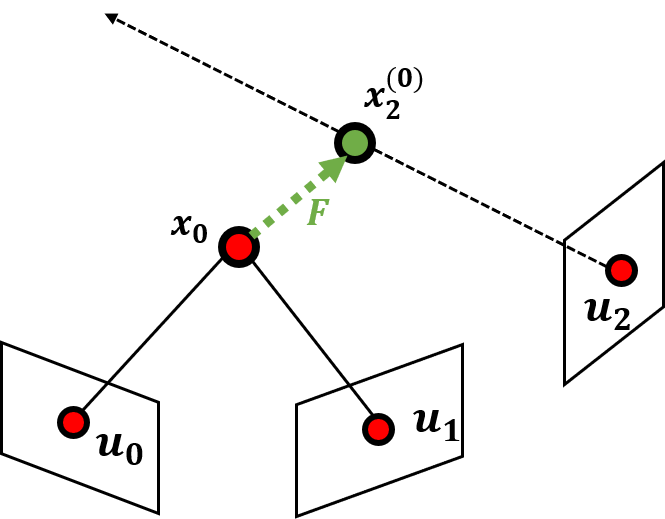}}
		\hfill
		\subfloat[Iteration 2]{\includegraphics[width=\size]{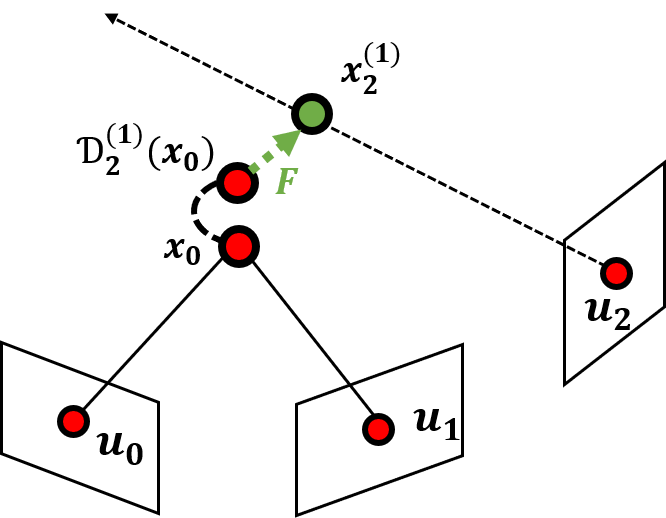}}
		\caption{\textbf{Sparse correspondence association}: In iteration $i$, we transform the 3D point $\mathbf{x}_0$ according to the previous estimate of the deformation $\mathbf{D}_2^{(i-1)}$ and project $\mathbf{D}_2^{(i - 1)}(\mathbf{x}_0)$ onto the ray defined by $\mathbf{u}_2$. The projection is used to define a force $F$ pulling the point towards the ray.}
		\label{fig:concept1}
		\vspace{-.3cm}
	\end{figure}

	We lift keypoints $\mathbf{u}_i$ to 3D points $\mathbf{x}_i$, if there is a depth value in at least one processed view, compute its coordinates in the canonical pose $\mathbf{D}_i^{-1}(\mathbf{x}_i)$ and apply the current estimate of our deformation field $\mathbf{D}_j$ for frame $j$ to these points. 
	To establish a sparse 3D-3D correspondence $(\mathbf{D}_i^{-1}(\mathbf{x}_i), \mathbf{x}_j)$ between the canonical pose and the current frame $j$ for the correspondence set $S$, we project $\mathbf{D}_j(\mathbf{D}_i^{-1}(\mathbf{x}_i))$ to the ray of the 2D keypoint $\mathbf{u}_j$ (see Fig.~\ref{fig:concept1}).
	To mitigate ambiguities and to constrain the problem, we also aim for dense photometric consistency across views.
	Thus, for each point of the template of the canonical pose, we also add a photometric consistency constraint with a mask $C_i \in \{ 0, 1 \}$.

	\noindent \textbf{Deformation and Depth Estimation}
	In our main iteration (see also Algorithm \ref{alg:algorithm1}), we estimate the deformation $\hat{\mathbf{D}}$ between the canonical pose and the currently selected view by minimizing the joint optimization problem:
	\begin{align}
	&E = w_\text{sparse} E_\text{sparse} + w_\text{dense} E_\text{dense} + w_\text{reg} E_\text{reg} \\
	&E_\text{sparse} = \sum_{(i, j) \in S} \| \hat{\mathbf{D}}(\mathbf{x}_i) - \mathbf{x}_j \|_2^2 \nonumber\\ 
	&E_\text{dense} = \sum_r \sum_s \sum_i C_i \cdot (1 - \rho_{r, s}(\hat{\mathbf{D}}(\mathbf{x}_i), \hat{\mathbf{D}}(\mathbf{n}_i), \mathbf{x}_i, \mathbf{n}_i))^2  \nonumber \\
	&E_\text{reg} = \sum_{j=1}^m \sum_{k \in N(j)} \| \mathbf{R}_j (\mathbf{g}_k - \mathbf{g}_j) + \mathbf{g}_j + \mathbf{t}_j - (\mathbf{g}_k + \mathbf{t}_k) \|_2^2 \nonumber
	\end{align}
	
	To measure photometric consistency $\rho_{r,s}$ between a reference image $r$, i.e.~the canonical pose, and a source view $s$, we use the bilaterally weighted adaption of normalized cross-correlation (NCC) as defined by Schoenberger et al. \cite{schoenberger2016mvs}. Throughout our pipeline, we employ COLMAP's default settings, i.e. a window of size $11 \times 11$. 
	The regularizer $E_\text{reg}$ as defined in \cite{sumner2007embedded} ensures a smooth deformation result.
	To ensure non-local convergence, we solve the problem in a coarse-to-fine manner using an image pyramid with 3 levels in total. 
	
	Both the sparse and dense matches are subject to outliers. 
	In the sparse case, these outliers manifest as incorrect keypoint matches across images.
	For the dense part, outliers mainly occur due to occlusions, either because of the camera pose or because of the observed deformation.
	
	To reject outliers in both cases, we reject correspondences with the highest residuals calculated from the result of the non-linear solution.
	We re-run the optimization until a user-specified maximum error is satisfied.
	This rejection is run in a 2-step process.
	First, we only solve for the deformation considering the sparse 3D-3D matches.
	Second, we fix the retained 3D-3D matches and solve the joint optimization problem, discarding only dense correspondences, resulting in a consistency map $C_i \in \{ 0, 1 \}$.
	
	We iterate this process (starting with the correspondence association) until we reach convergence.
	In our experiments, we found that 3 to 5 iterations suffice to ensure a converged state.
	
	To estimate the depth for the currently processed view, we then run a modified, non-rigid variant of COLMAP's PatchMatch \cite{schoenberger2016mvs}.
	Instead of simple homography warping, we apply the deformation to the point and its normal.
	
	\subsection{Implementation Details}
	\label{sec:implementation}
	In this section, we provide more details on our implementation of the NRMVS framework (Fig.~\ref{fig:overview}).
	Algorithm~\ref{alg:algorithm1} shows the overall method, introduced in Sec.~\ref{sec:deformation_model}, and Sec.~\ref{sec:photometric-conssitency}.
	
	Given input RGB images, we first pre-process the input.
	To estimate the camera pose for the images, we use the SfM implementation of Agisoft Photoscan \cite{photoscan}. Our tests showed accurate results for scenes containing at least 60\% static background. A recent study~\cite{Lv18eccv} shows that 60$\sim$90\% of static regions in a scene results in less than $0.02$ degree RPE~\cite{Sturm12iros} error for standard pose estimation techniques (see more discussion in the appendix~\ref{sec:pose}.
	Given the camera pose, we triangulate sparse SIFT matches \cite{lowe1999object}, i.e.,~we compute the 3D position of the associated point by minimizing the reprojection error. 
	We consider matches with a reprojection error of less than 1 pixel to be successfully reconstructed (static inliers).
	The ratio of static inliers to the number of total matches is a simple yet effective indication of the non-rigidity in the scene.
	We pick the image pair with the highest ratio to indicate the minimum amount of deformation and use these as the canonical views to bootstrap our method.
	
	Two important aspects of our main iteration are described in more detail:
	Our method to filter sparse correspondences (line 16 in Algorithm~\ref{alg:algorithm1}) is given in Algorithm~\ref{alg:filter}.
	The hierarchical optimization algorithm (line 17 in Algorithm~\ref{alg:algorithm1}) including filtering for dense correspondences is given in Algorithm~\ref{alg:optimization}.
	
	The joint optimization in our framework is a computationally expensive task. 
	The deformation estimation, which strongly dominates the overall run-time, is CPU intensive, while the depth computation runs on the GPU. 
	Specifically, for the face example shown in Fig.~\ref{fig:teaser} (6 images with 100 deformation nodes) the computation time needed is approximately six hours (Intel i7-6700 3.4 GHz, NVIDIA GTX 980Ti).
	More details about the computational expense will be discussed in the appendix~\ref{sec:performance}.
	
	\begin{algorithm}
		\KwData{RGB input images $\{ \mathbf{I}_k \}$}
		\KwResult{Deformations $\{ \mathbf{D}_k \}$, depth $\{ d_k \}$}
		$P := \{ 1, \ldots, k \}, Q := \emptyset$ \;
		$\{ \mathbf{C}_k \}$ = PhotoScanEstimateCameraPoses()\;
		$(i, j)$ = selectCanonicalViews()\;
		$(d_i^{(0)}, \mathbf{n}_i^{(0)}, d_j^{(0)}, \mathbf{n}_j^{(0)})$ = ColmapPatchMatch($\mathbf{I}_i, \mathbf{I}_j$)\;
		$\mathbf{D}_i^{(0)} = \mathbf{D}_j^{(0)}$ = initDeformationGraph($d_i^{(0)}, d_j^{(0)}$)\;
		$\{ \mathbf{u}_k \}$ = computeFeatureTracks()\;
		$Q := Q \cup \{ i, j \}$\;
		\While{$Q \neq P$}{
			$l$ = nextImage($P \setminus Q$)\;
			$\{ \mathbf{x}_k \}$ = liftKeyPointsTo3D($\{ \mathbf{u}_k \}_{k \in Q}$) \;
			$\{ \mathbf{x}_i \} = \mathbf{D}_k^{-1}(\{ \mathbf{x}_k \})$ \;
			$\mathbf{D}_l^{(1)} = \mathbf{Id}$ \;
			\For{$m = 1$ \KwTo $N$}{
				$\{ \mathbf{\hat{x}}_l^{(m)} \} = \mathbf{D}_l^{(m)}(\{ \mathbf{x}_i \})$ \;
				$\{ \mathbf{x}_l^{(m)} \}$ = projToRays($\{ \mathbf{\hat{x}}_l^{(m)} \}, \{ \mathbf{u}_l \})$ \;
				$\{ (\mathbf{\tilde{x}}_i, \mathbf{\tilde{x}}_l^{(m)}) \}$ = filter($\mathbf{D}_l^{(m)}, \{ (\mathbf{x}_i, \mathbf{x}_l^{(m)}) \}$)\;
				$\mathbf{D}_l^{(m + 1)}$ = solve($\mathbf{D}_l^{(m)}, \{ (\mathbf{\tilde{x}}_i, \mathbf{\tilde{x}}_l^{(m)}) \}$, $\mathbf{I}_i, d_i^{(0)}, \mathbf{n}_i^{(0)}, \mathbf{I}_j, d_j^{(0)}, \mathbf{n}_j^{(0)}, \mathbf{I}_l$)
			}
			$\mathbf{D}_l = \mathbf{D}_l^{(m + 1)}$ \;
			$Q := Q \cup \{ l \}$\;
			$(d_l^{(0)}, \mathbf{n}_l^{(0)})$ = NRPatchMatch($\{ \mathbf{I}_k, \mathbf{D}_k \}_{k \in Q}$)\;
		}
		$\{ (d_k, \mathbf{n}_k) \}_{k \in Q}$ = NRPatchMatch($\{ \mathbf{I}_k, \mathbf{D}_k \}_{k \in Q}$)\;
		\caption{Non-rigid multi-view stereo}
		\label{alg:algorithm1}
	\end{algorithm}
	\vspace{-0.2cm}
	\begin{algorithm}
		\SetKwFunction{FMain}{filter}
		\SetKwProg{Fn}{Function}{:}{}
		\KwData{Threshold $d_\text{max}$, Ratio $\tau \in (0, 1)$}
		\Fn{\FMain{$\mathbf{D}_l, \{ (\mathbf{x}_i, \mathbf{x}_l) \}$}}{
			\While{true}{
				$\mathbf{D}_l^\ast$ = solve($\mathbf{D}_l, \{ (\mathbf{x}_i, \mathbf{x}_l) \}$)\;
				$\{ r_k \} = \{ \| \mathbf{D}_l^\ast(\mathbf{x}_i) - \mathbf{x}_l \|_2 \}$\;
				$e_\text{max} = \max \{ r_k \} $\;
				\If{$e_\text{max} < d_\text{max}$}{
					break\;
				}
				$d_\text{cut} := \max \{ d_\text{max}, \tau \cdot e_\text{max} \}$ \;
				$\{ (\mathbf{x}_i, \mathbf{x}_l) \} := \{ (\mathbf{x}_i, \mathbf{x}_l) : r_k < d_\text{cut} \} $\;
			}
			\Return $\{ (\mathbf{x}_i, \mathbf{x}_l) \}$\;
		}
		\caption{Filtering of sparse correspondences}
		\label{alg:filter}
	\end{algorithm}
	\vspace{-0.2cm}
	\begin{algorithm}
		\SetKwFunction{FMain}{solve}
		\SetKwProg{Fn}{Function}{:}{}
		\KwData{Threshold $\rho_\text{max}$, Ratio $\tau \in (0, 1)$}
		\Fn{\FMain{$\mathbf{D}_l, \{ (\mathbf{x}_i, \mathbf{x}_l) \}, \mathbf{I}_i, d_i,     \mathbf{n}_i, \mathbf{I}_j, d_j, \mathbf{n}_j, \mathbf{I}_l$}}{
			$\hat{\mathbf{D}}_l = \mathbf{D}_l$\;
			\For{$m = 1$ \KwTo levels}{
				$\rho_\text{cut} := \tau \cdot (1 - \text{NCC}_\text{min}) = \tau \cdot 2$ \;
				$C_p := 1 \quad \forall p$\;
				\While{true}{
					$\mathbf{D}_l^\ast$ = solveEq1($\hat{\mathbf{D}_l}$) \;
					$\{ r_p \} = \{ C_p \cdot~(1~-~\rho(\mathbf{D}_l^\ast(\mathbf{x}_p), \mathbf{D}_l^\ast(\mathbf{n}_p), \mathbf{x}_p, \mathbf{n}_p)) \}$\;
					$e_\text{max} = \max \{ r_p \} $\;
					\If{$e_\text{max} < \rho_\text{max}$}{
						$\hat{\mathbf{D}}_l := \mathbf{D}_l^\ast$\;
						break\;
					}
					\If{$m = levels$}{
						$\hat{\mathbf{D}}_l := \mathbf{D}_l^\ast$\;
					}
					$C_p := 0 \quad \forall p : r_p > \rho_\text{cut}$\;
					$\rho_\text{cut} := \max \{ \rho_\text{max}, \tau \cdot \rho_\text{cut} \}$
				}
			}
			\Return $\hat{\mathbf{D}}_l$
		}
		\caption{Solving the joint problem}
		\label{alg:optimization}
	\end{algorithm}
	\vspace{-0.2cm}

	\begin{table*}[t]
		\caption{\textbf{Evaluation for ground truth data:} (a) using COLMAP, i.e., assuming a static scene, (b) applying our dense photometric optimization on top of an implementation of non-rigid ICP (NRICP), and (c) using different variants of our algorithm. S denotes \emph{sparse}, D denotes \emph{dense}, photometric objective. $N$ equals the number of iterations for sparse correspondence association (see paper for more details). We compute the mean relative error (MRE) for all reconstructed values as well as the overall completeness. The last row (w/o filter) shows the MRE, with disabled rejection of outlier depth values, i.e., a completeness of 100 \%.
		}
		\vspace{-.2cm}
		\begin{tabularx}{\textwidth}{@{}rccccccc@{}}
			\toprule
			& & \multicolumn{6}{c}{Ours} \\
			\cmidrule{4-8}
			& COLMAP \cite{schoenberger2016mvs} & NRICP~\cite{Li09} & S ($N=1$) & S ($N=10$) & D & S ($N=1$) + D &  S ($N=10$) + D \\ 
			\midrule 
			Completeness & 68.74 \% & 99.30 \%  & 97.24 \%  & 97.71 \% & 96.41 \% & 98.76 \% &  \textbf{98.99} \% \\
			\hline
			MRE & 2.11 \% & 0.53 \%  & 1.48 \% & 1.50 \% & 2.37 \% & 1.12 \% & \textbf{1.11} \%  \\ 
			\hline 
			\hline
			MRE w/o filter & 6.78 \% & 0.74 \%  & 2.16 \% & 2.05 \% & 3.32 \% & 1.63 \% & \textbf{1.34} \%  \\ 
			\bottomrule 
		\end{tabularx} 
		\label{tab:eval}
		\vspace{-0.2cm}
	\end{table*}

	\begin{figure*}
		\centering
		\newlength\exlen
		\setlength\exlen{.15\linewidth}
		\def\colorbarSize{1.5cm}
		\captionsetup[subfloat]{labelformat=empty}
		\subfloat{\parbox[t]{.02\linewidth}{\begin{sideways}\centering \footnotesize \, \, \, \,  Input\end{sideways}}}
		\hfill
		\subfloat{\includegraphics[trim=11cm 7cm 11cm 4cm, clip=true,width=\exlen]{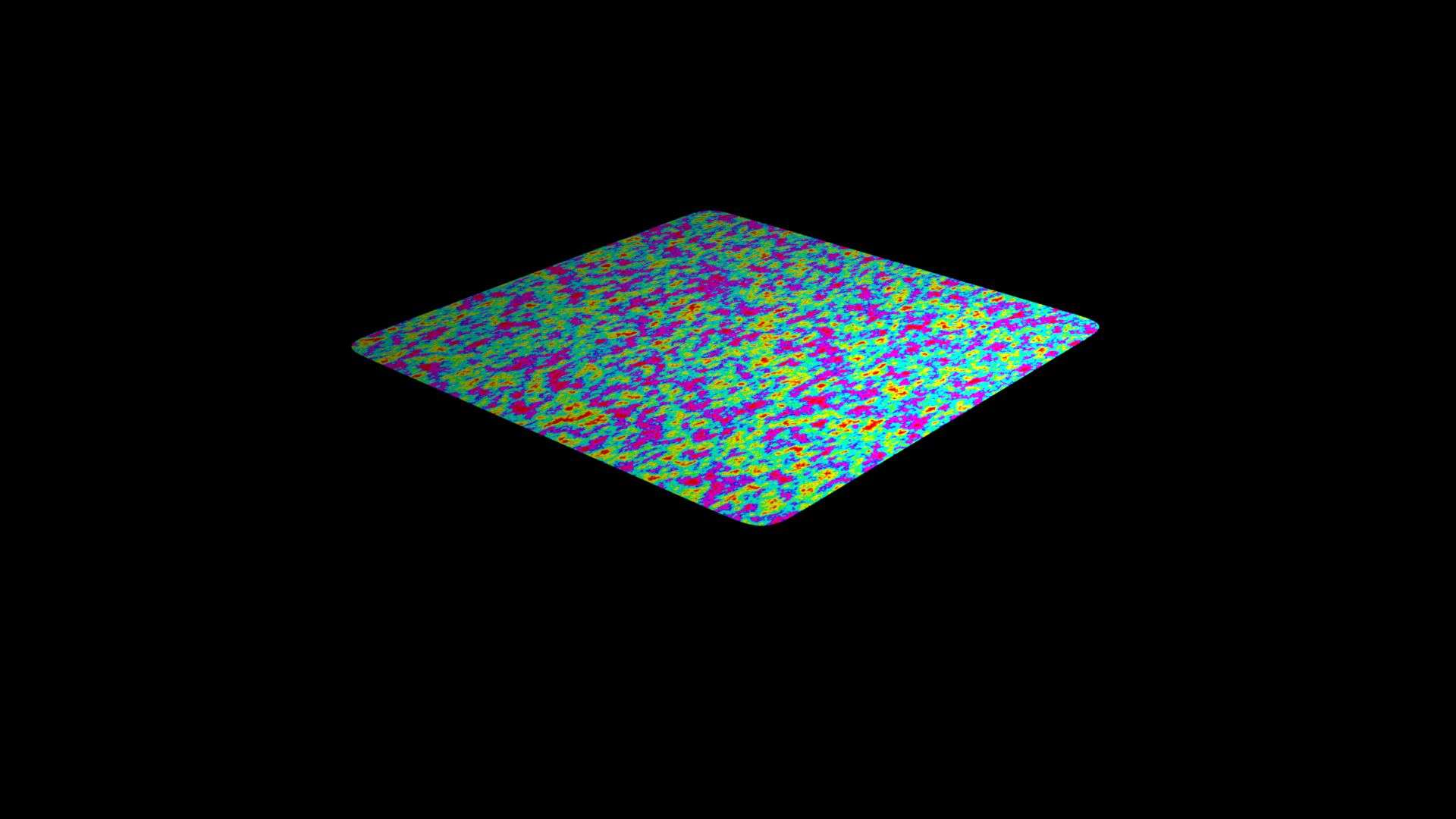}} 
		\hfill
		\subfloat{\includegraphics[trim=12cm 6cm 10cm 5cm, clip=true,width=\exlen]{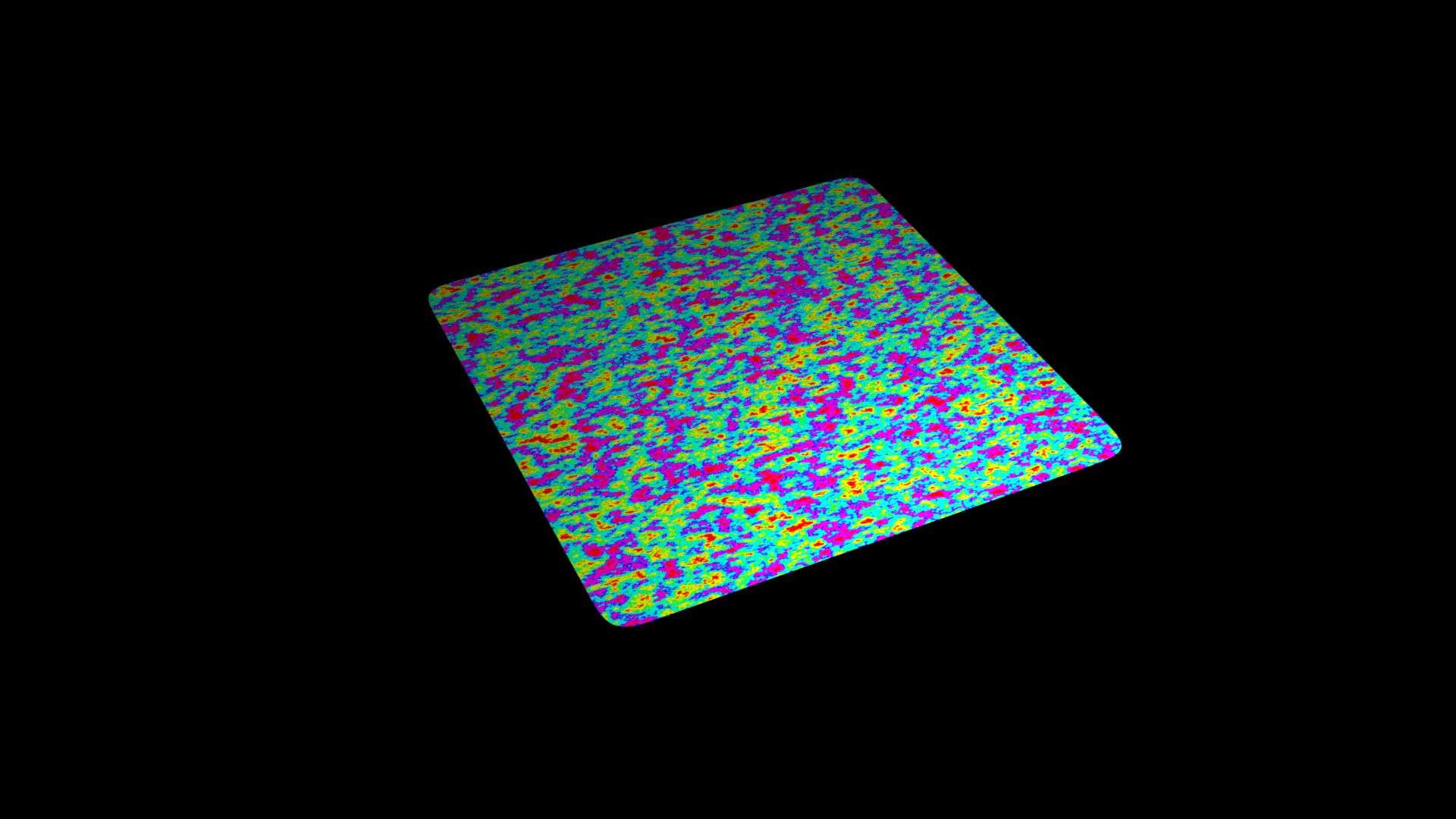}} 
		\hfill
		\subfloat{\includegraphics[trim=7.7cm 4.9cm 7.7cm 2.2cm, clip=true,width=\exlen]{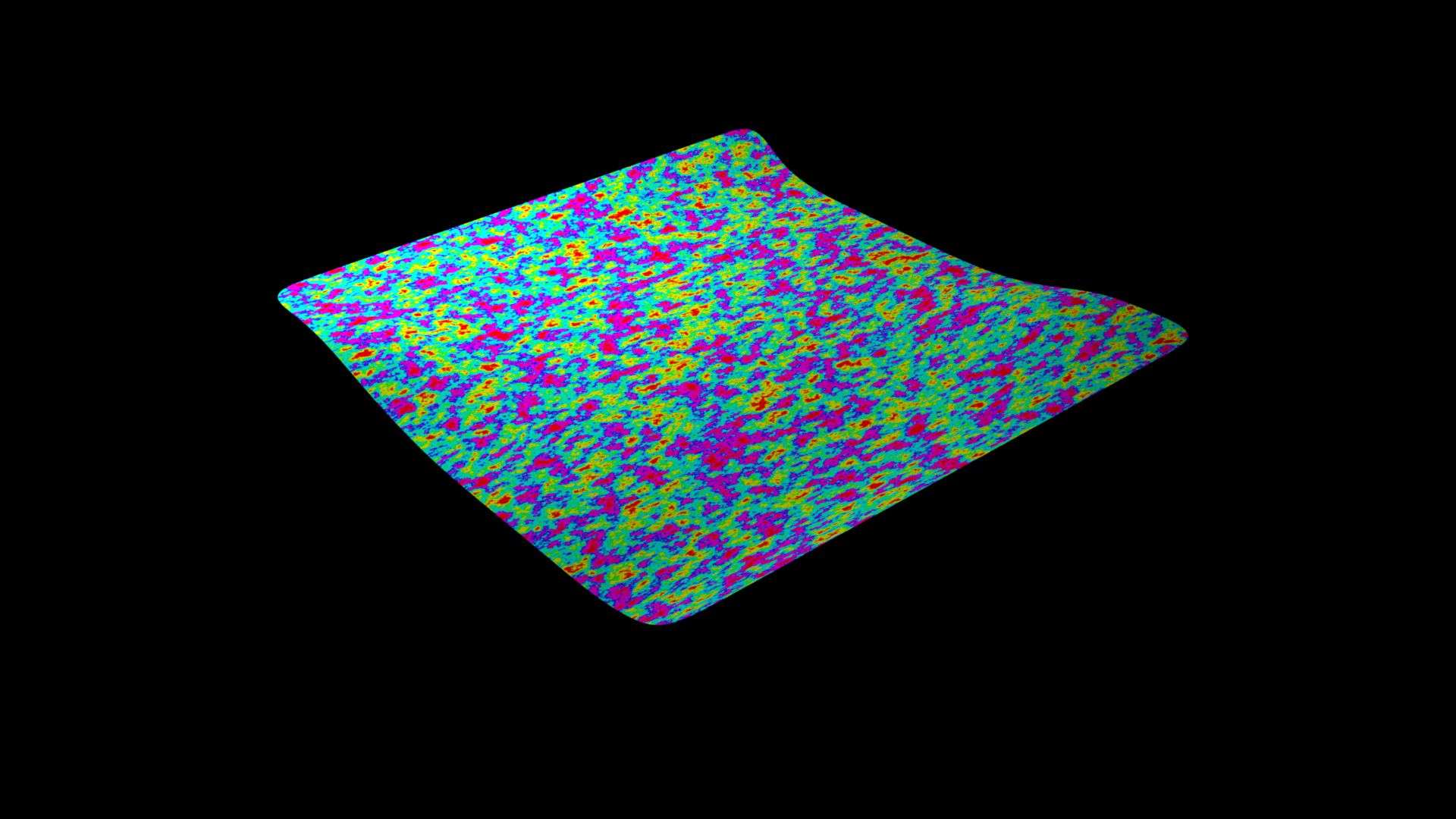}} %
		\hfill
		\subfloat{\includegraphics[trim=12.5cm 5.5cm 9.5cm 5.5cm, clip=true,width=\exlen]{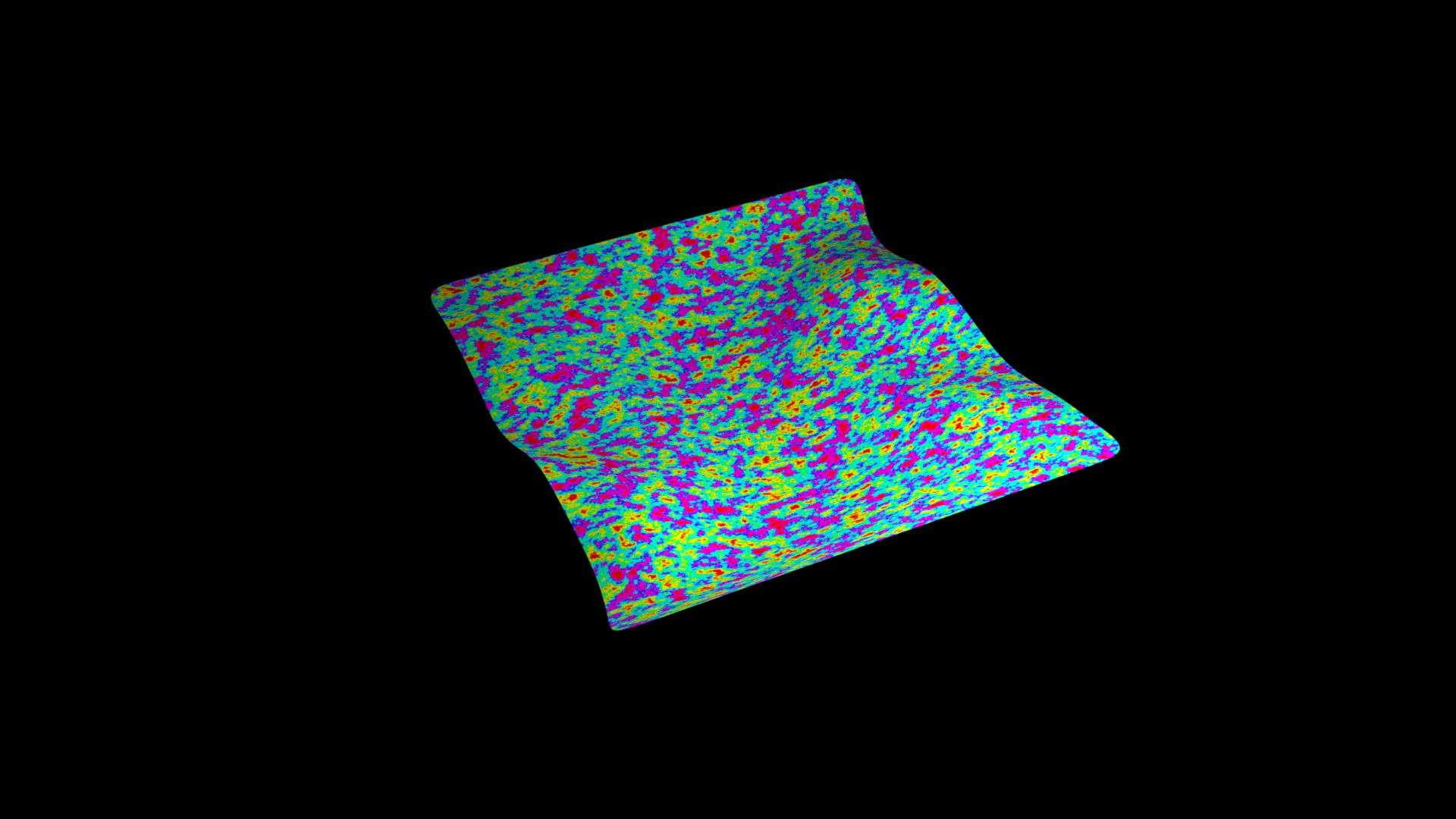}} %
		\hfill
		\subfloat{\includegraphics[trim=11cm 7cm 11cm 4cm, clip=true,width=\exlen]{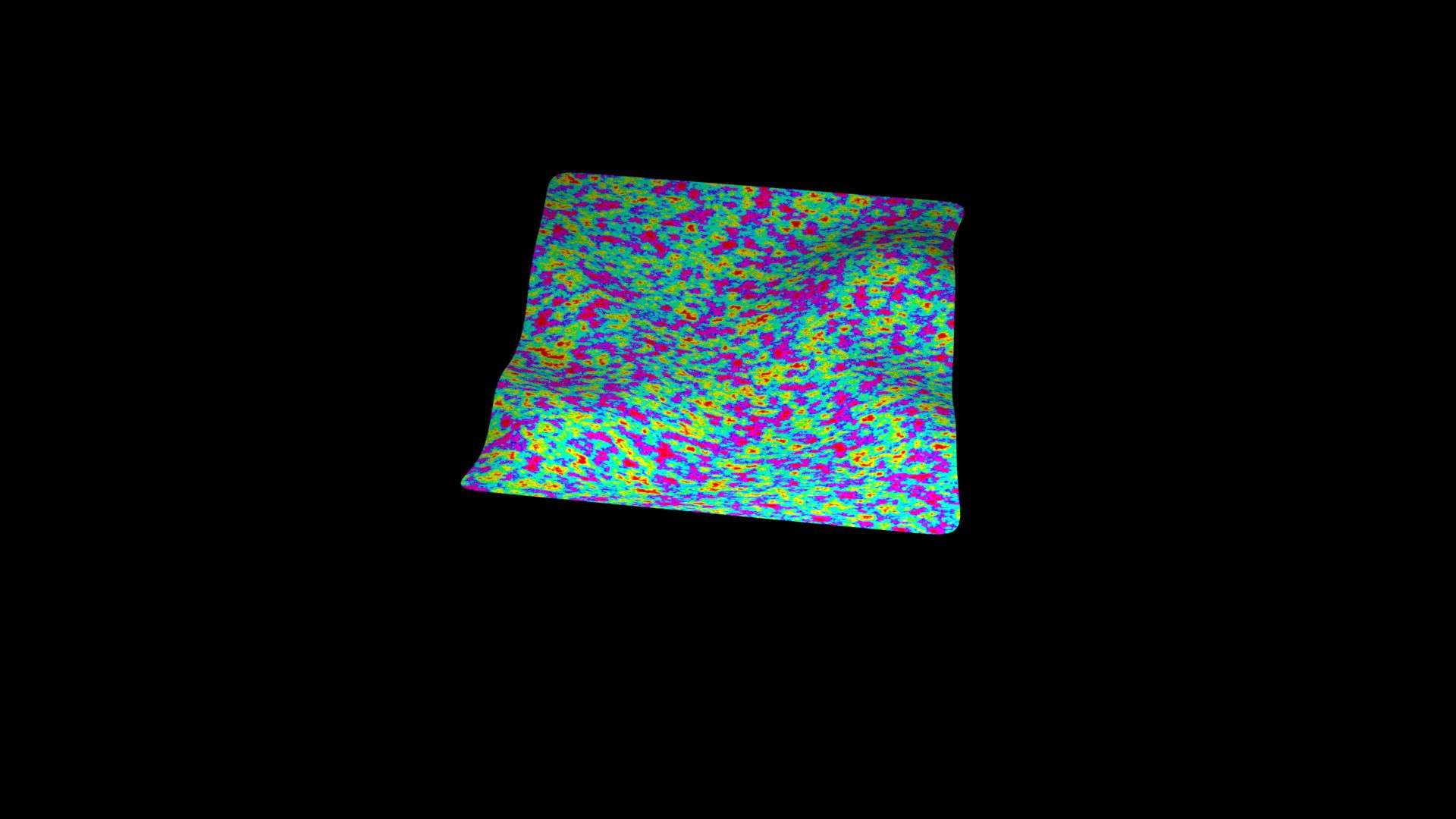}} %
		\hfill
		\subfloat{\includegraphics[trim=11cm 7cm 11cm 4cm, clip=true,width=\exlen]{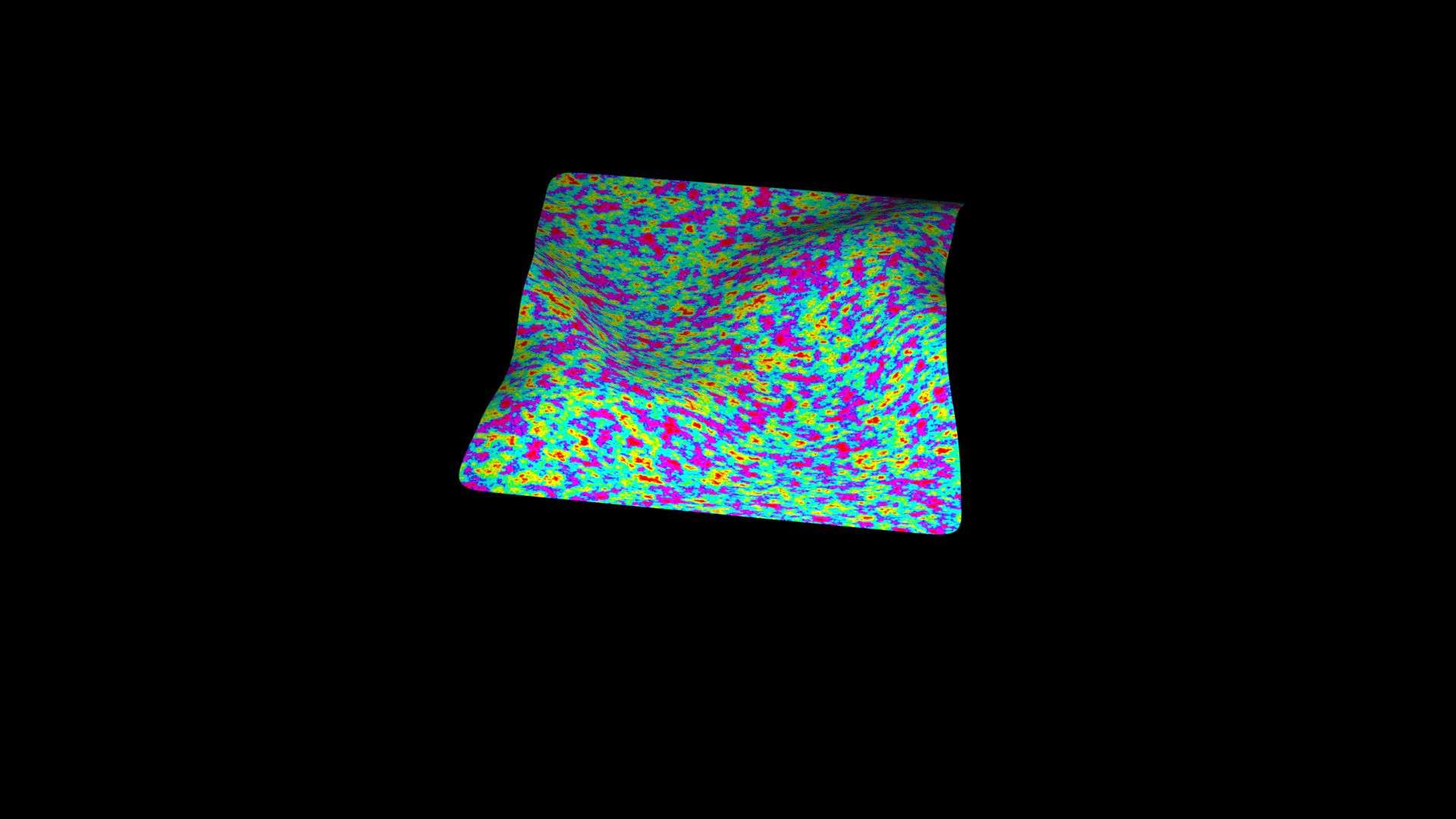}}
		\hfill
		\subfloat{\includegraphics[height=\colorbarSize]{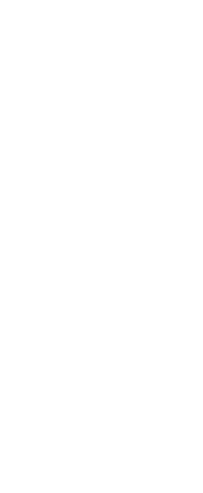}}  \\ %
		\vspace{-0.3cm}
		\subfloat{\parbox[t]{.02\linewidth}{\begin{sideways}\centering \footnotesize \quad \, 3D points\end{sideways}}}
		\hfill
		\subfloat{\includegraphics[trim=9cm 7cm 8cm 5cm, clip=true,width=\exlen]{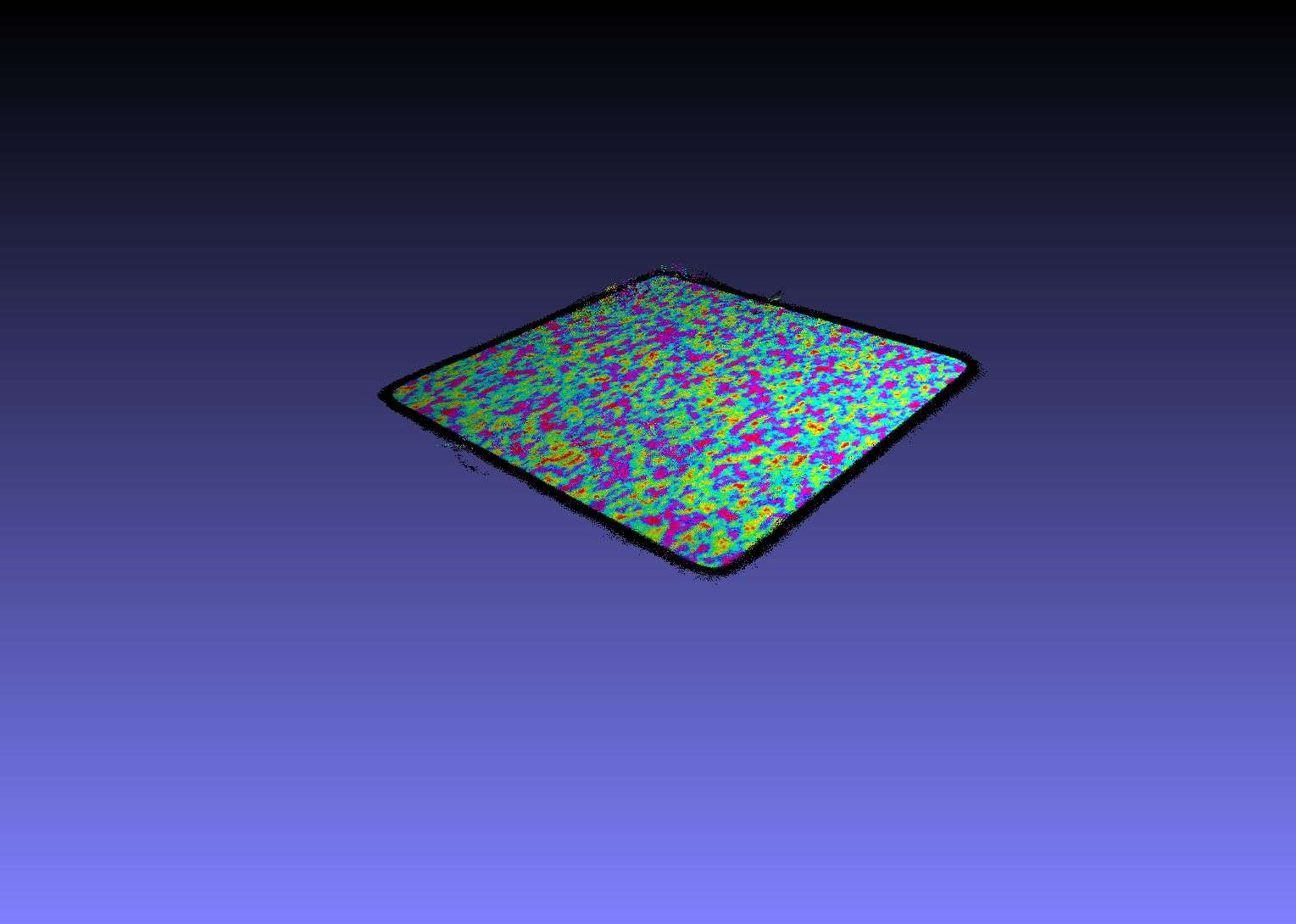}} 
		\hfill
		\subfloat{\includegraphics[trim=10cm 6cm 7cm 6cm, clip=true,width=\exlen]{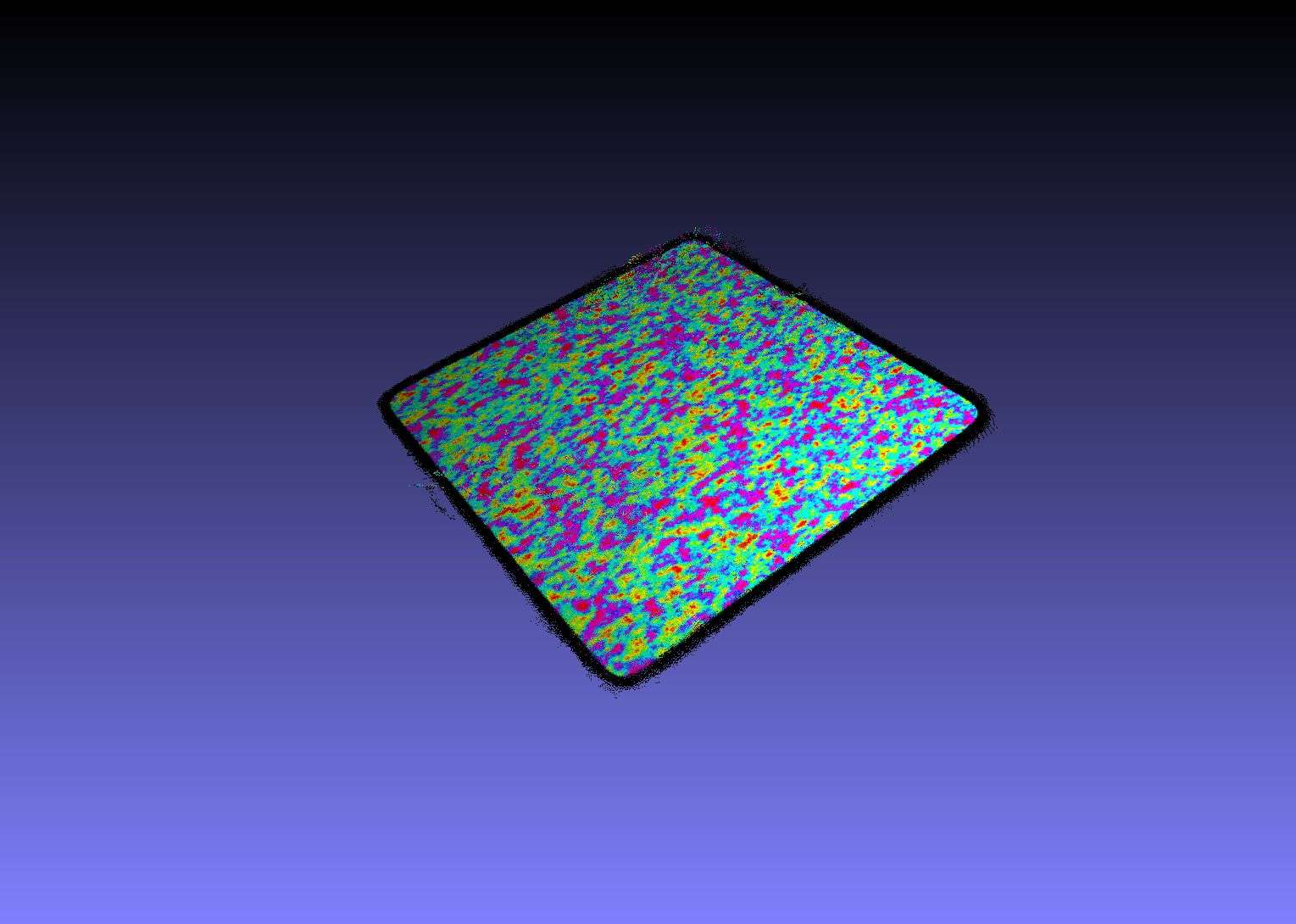}} 
		\hfill
		\subfloat{\includegraphics[trim=9cm 6cm 8cm 6cm, clip=true,width=\exlen]{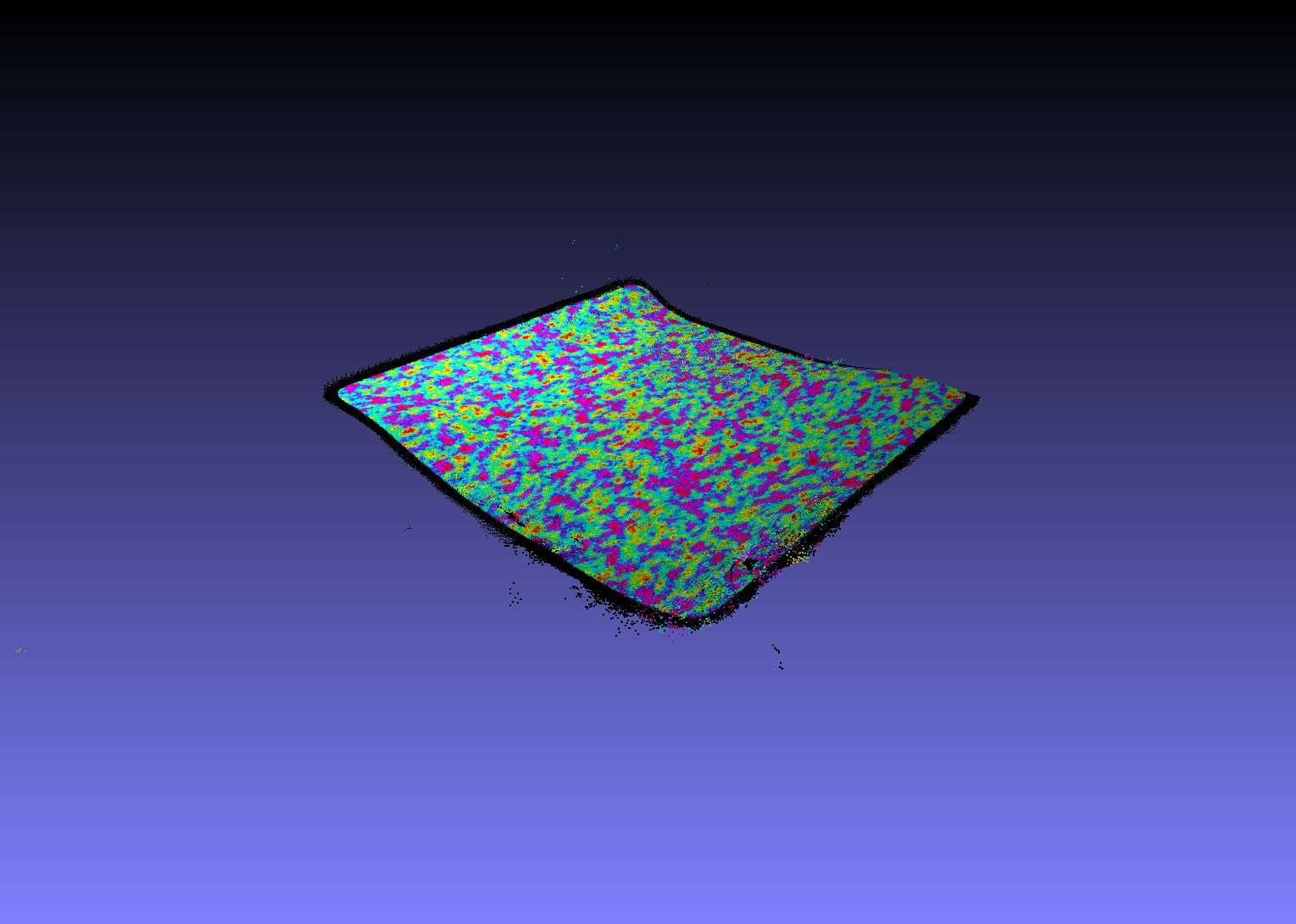}} %
		\hfill
		\subfloat{\includegraphics[trim=9cm 6cm 8cm 6cm, clip=true,width=\exlen]{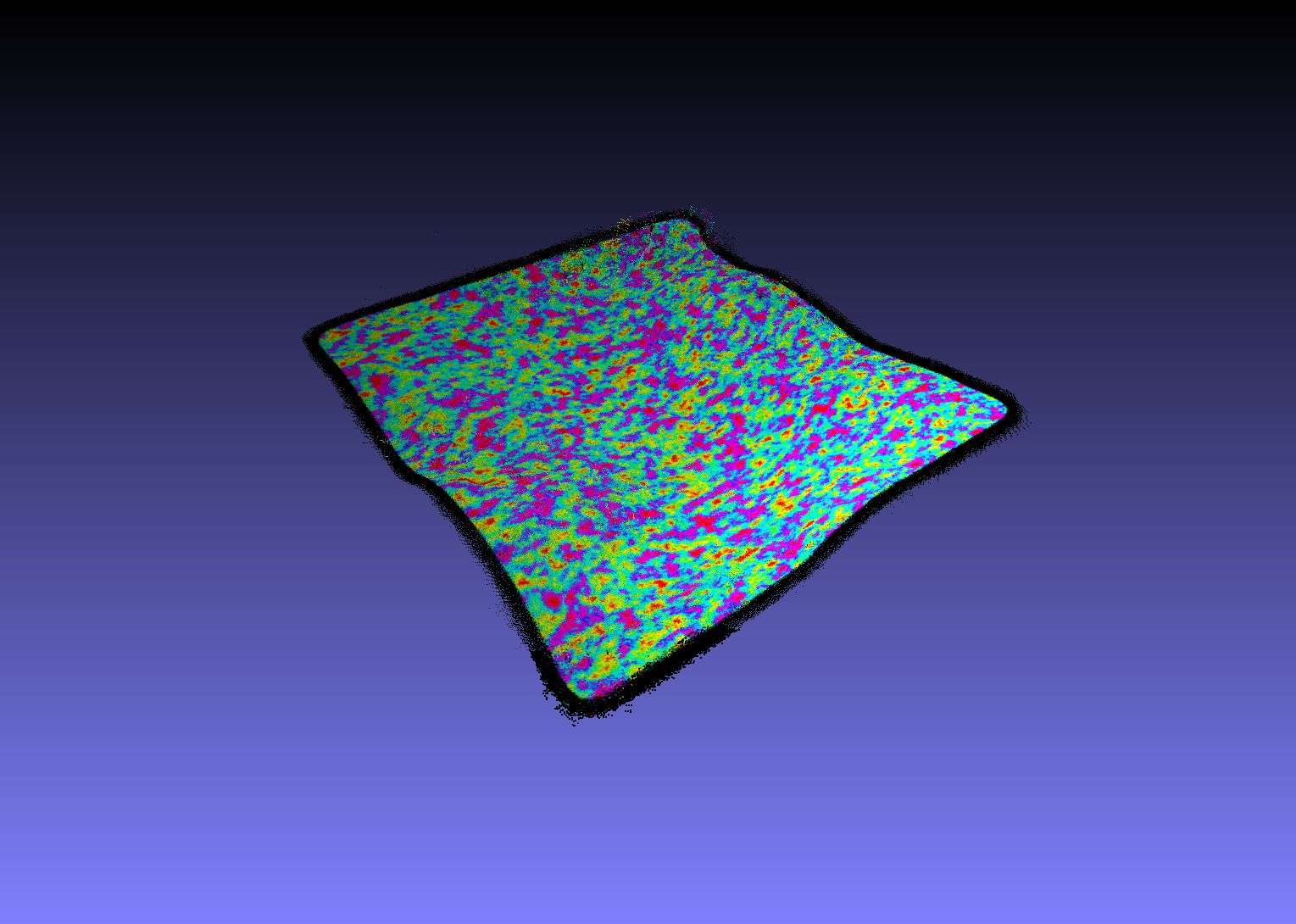}} %
		\hfill
		\subfloat{\includegraphics[trim=8.5cm 6cm 8.5cm 6cm, clip=true,width=\exlen]{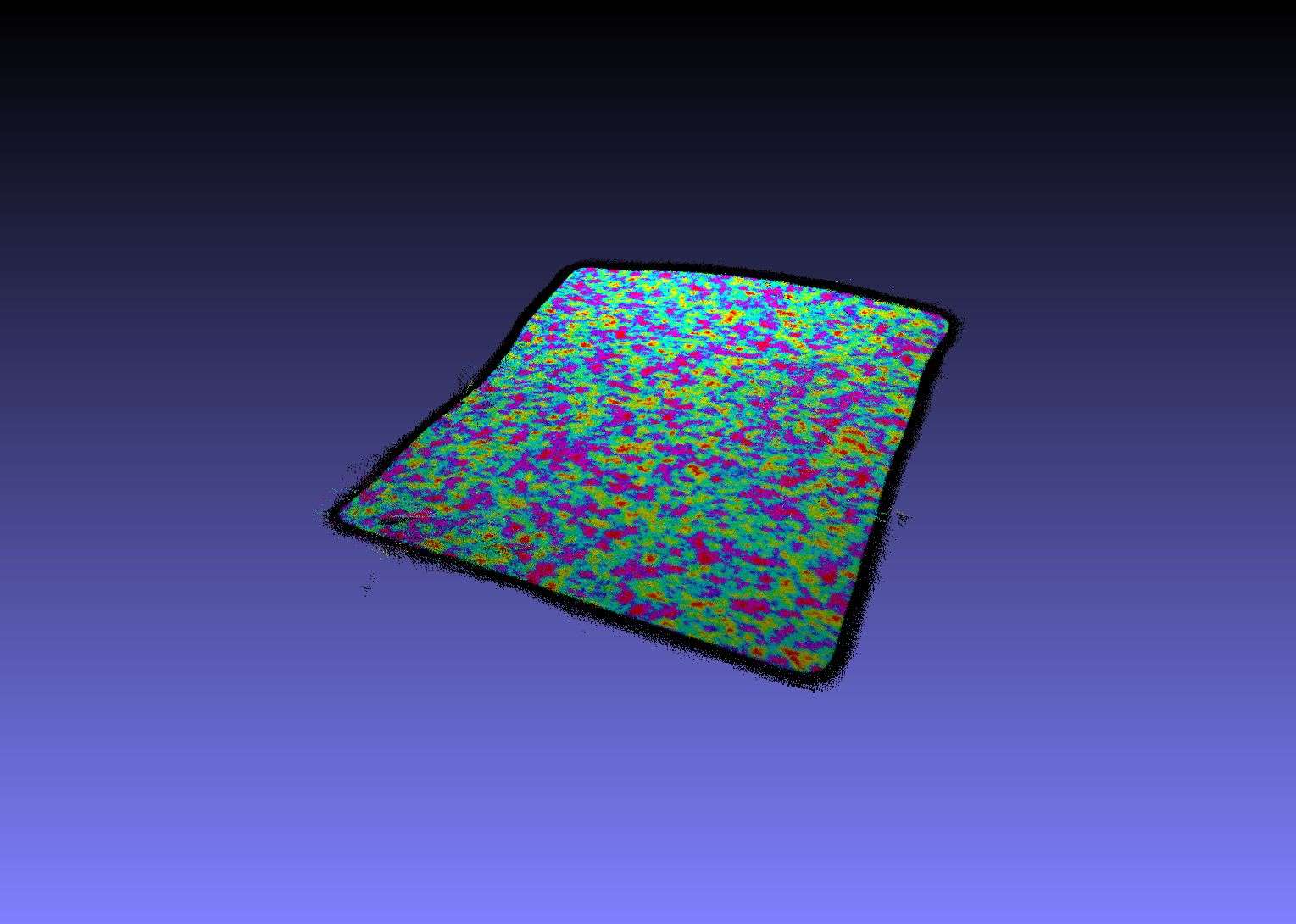}} %
		\hfill
		\subfloat{\includegraphics[trim=7cm 6cm 10cm 6cm, clip=true,width=\exlen]{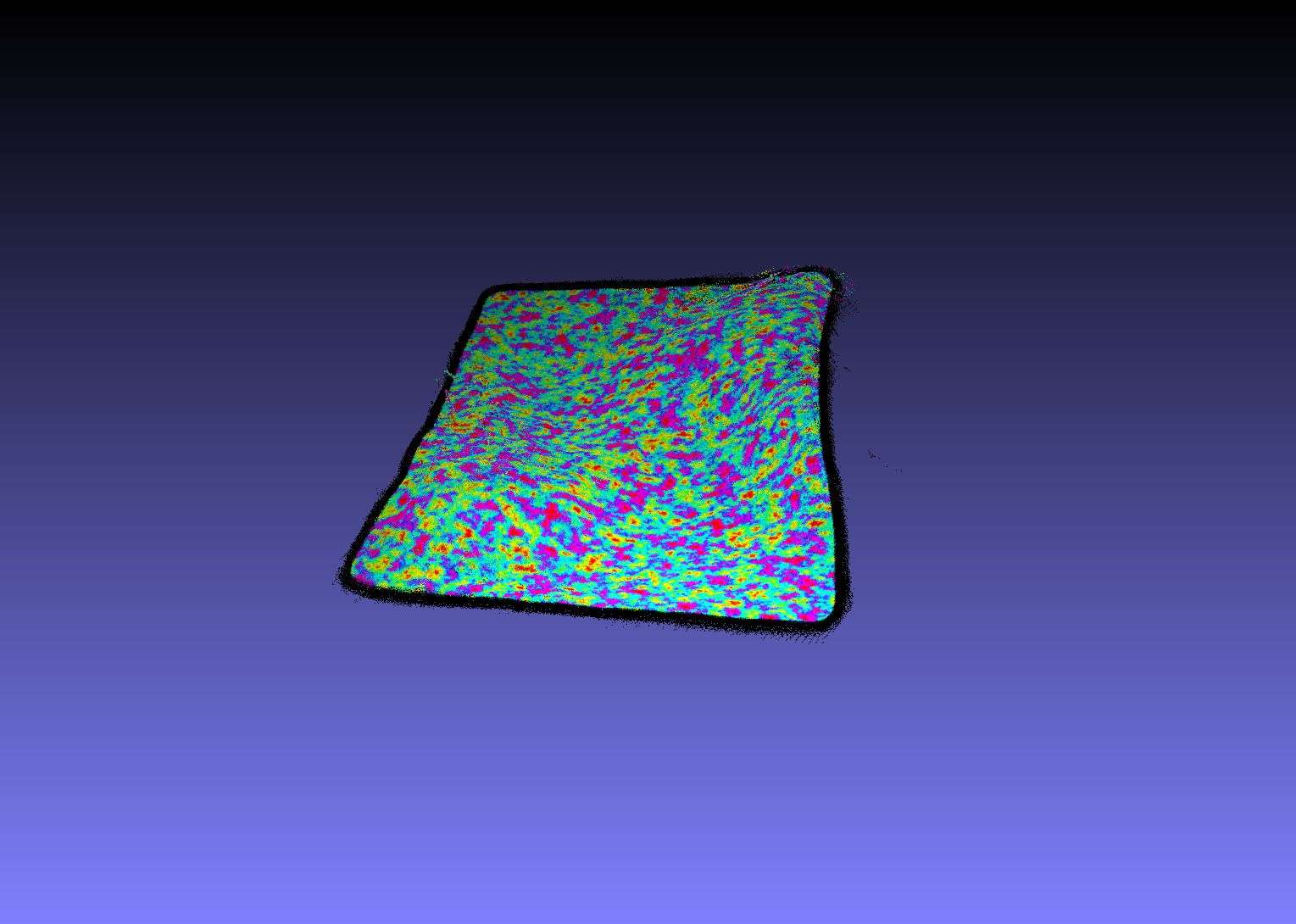}}
		\hfill
		\subfloat{\includegraphics[height=\colorbarSize]{gt_eval_colorbar_dummy}} \\ %
		\vspace{-0.3cm}
		\subfloat{\parbox[t]{.02\linewidth}{\begin{sideways}\centering \footnotesize Relative error \end{sideways}}}
		\hfill
		\subfloat[\scriptsize{[0.51 \% // 97.4 \%]}]{\includegraphics[trim=5.5cm 3.5cm 5.5cm 2cm, clip=true,width=\exlen]{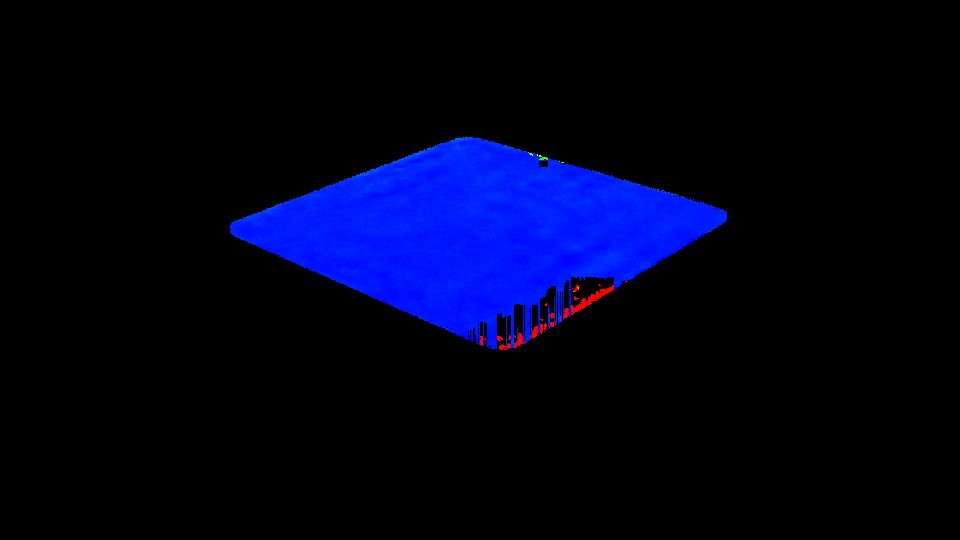}} 
		\hfill
		\subfloat[\scriptsize{[0.08 \% // 100 \%]}]{\includegraphics[trim=6cm 3cm 5cm 2.5cm, clip=true,width=\exlen]{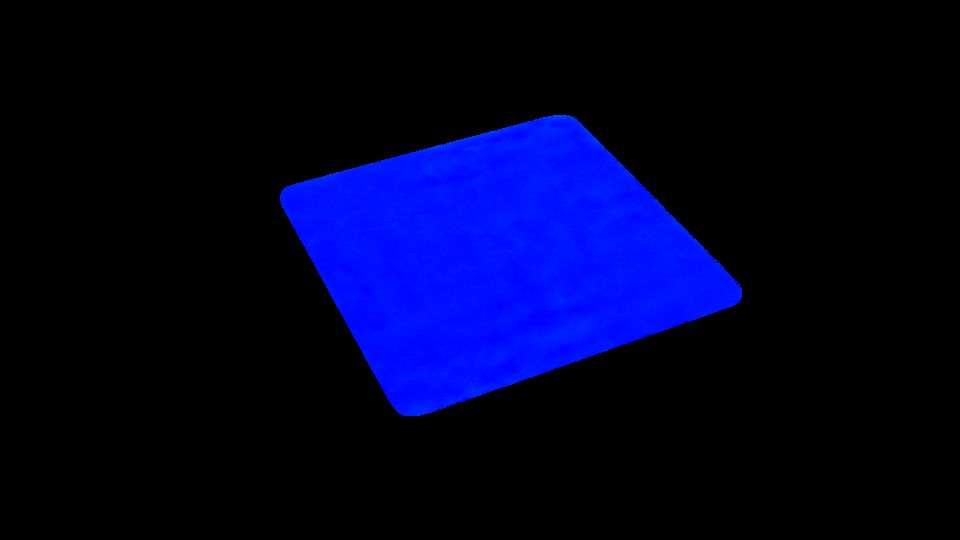}} 
		\hfill
		\subfloat[\scriptsize{[1.23 \% // 99.25 \%]}]{\includegraphics[trim=3.85cm 2.45cm 3.85cm 1.1cm, clip=true,width=\exlen]{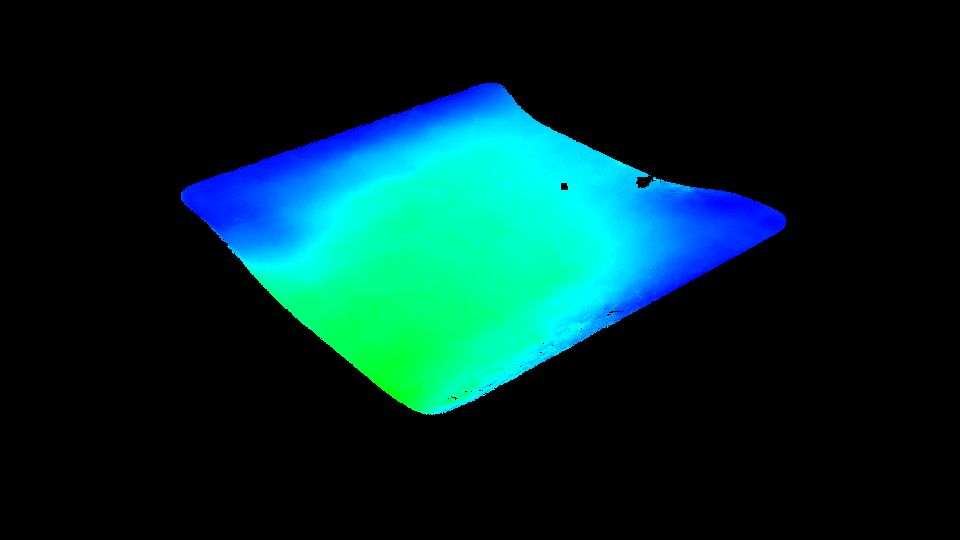}} %
		\hfill
		\subfloat[\scriptsize{[1.31 \% // 99.44 \%]}]{\includegraphics[trim=6.25cm 2.75cm 4.75cm 2.75cm, clip=true,width=\exlen]{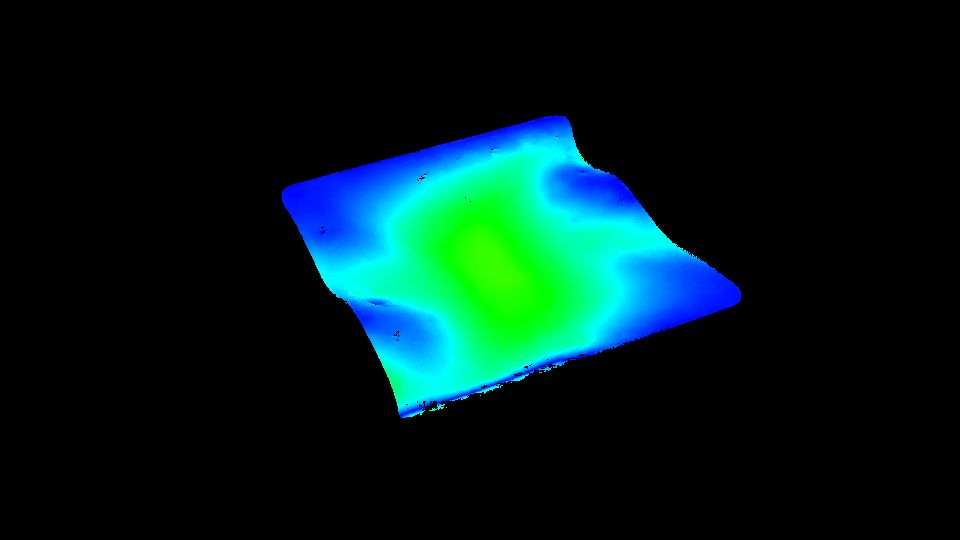}} %
		\hfill
		\subfloat[\scriptsize{[0.97 \% // 99.98 \%]}]{\includegraphics[trim=5.5cm 3.5cm 5.5cm 2cm, clip=true,width=\exlen]{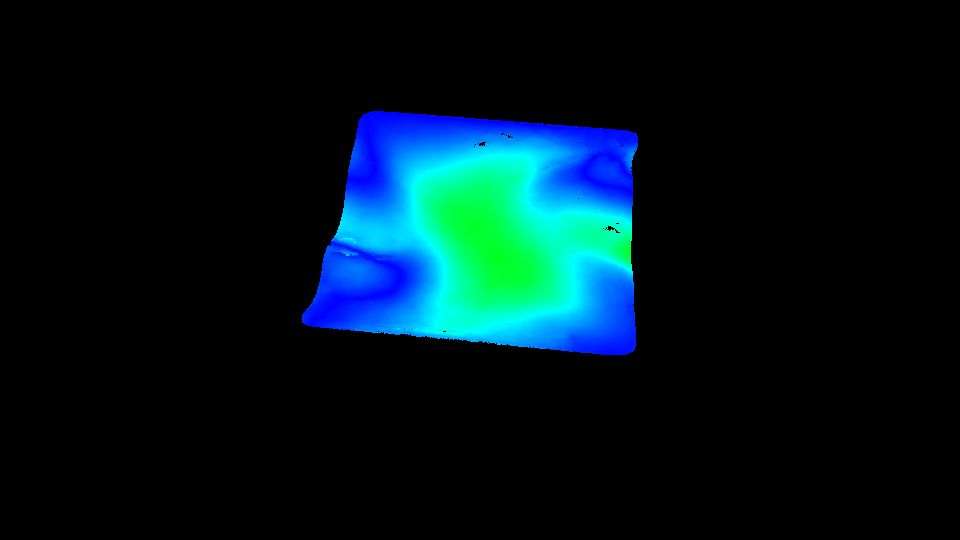}} %
		\hfill
		\subfloat[\scriptsize{[1.00 \% // 99.14 \%]}]{\includegraphics[trim=5.5cm 3.5cm 5.5cm 2cm, clip=true,width=\exlen]{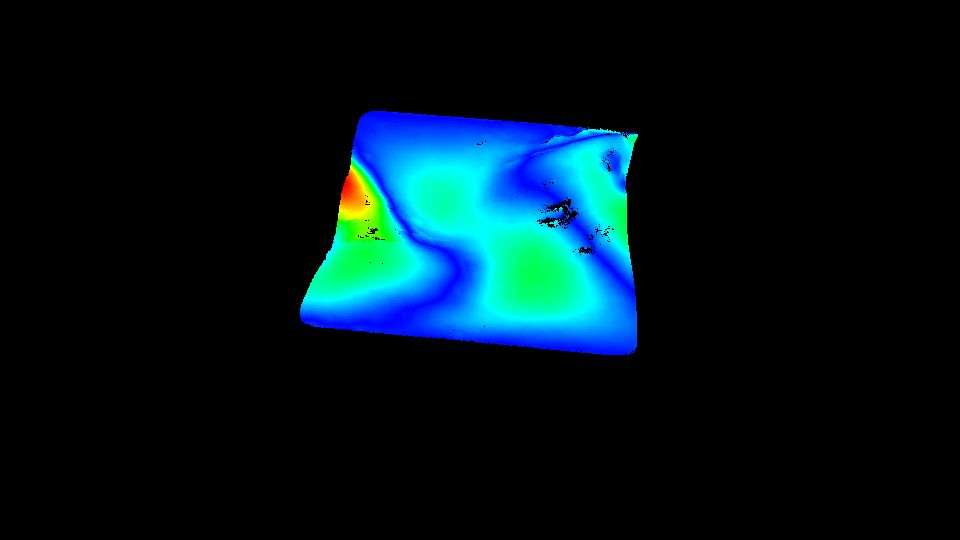}} 
		\hfill
		\subfloat{\includegraphics[height=\colorbarSize]{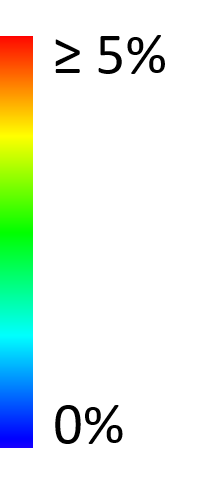}}
		\caption{\textbf{Quantitative evaluation with synthetic data:} We created images of a deforming surface with 10 different views. For the evaluation, we randomly chose six examples from the set and reconstructed the surface. The first row shows the input images. The first two columns show the chosen canonical views. The results of the reconstructed surface (with the point cloud propagated to each view) are shown in the second row. In the third row, we visualize the relative depth error compared to the ground truth. We also show the mean relative depth error value (\%) and the completeness (\%). The overall quantitative evaluation including a comparison to other baselines are shown in Table~\ref{tab:eval}.}
		\label{fig:results-synth}
		\vspace{-.2cm}
	\end{figure*}
	
	\begin{figure*}
		\def\vspaceSame{\vspace{-1em}}
		\def\vspaceScene{\vspace{-0.2cm}}
		\centering
		\subfloat{\parbox[t]{.02\linewidth}{\begin{sideways}\centering \footnotesize \qquad Input images\end{sideways}}}
		\hfill
		\subfloat{\includegraphics[trim=3cm 0.6cm 1.8cm 0cm, clip=true,width=.16\linewidth]{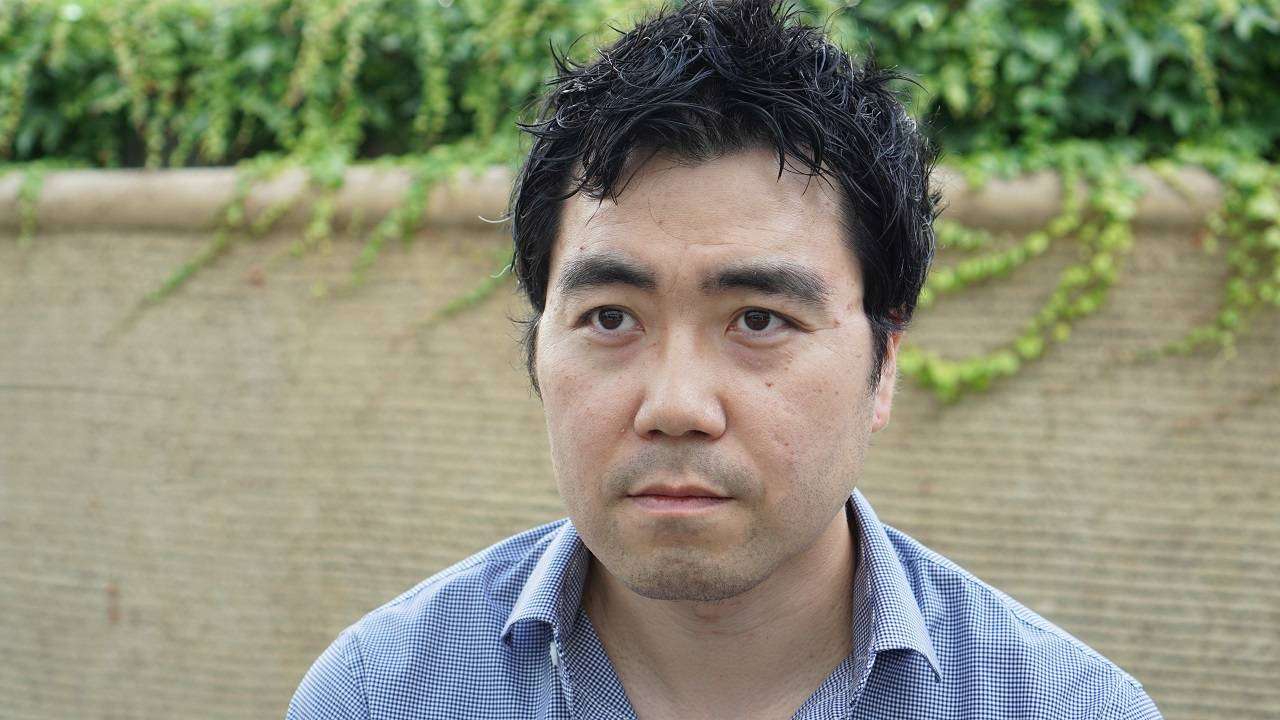}}
		\hfill
		\subfloat{\includegraphics[trim=2cm 0.3cm 2.8cm 0.3cm, clip=true,width=.16\linewidth]{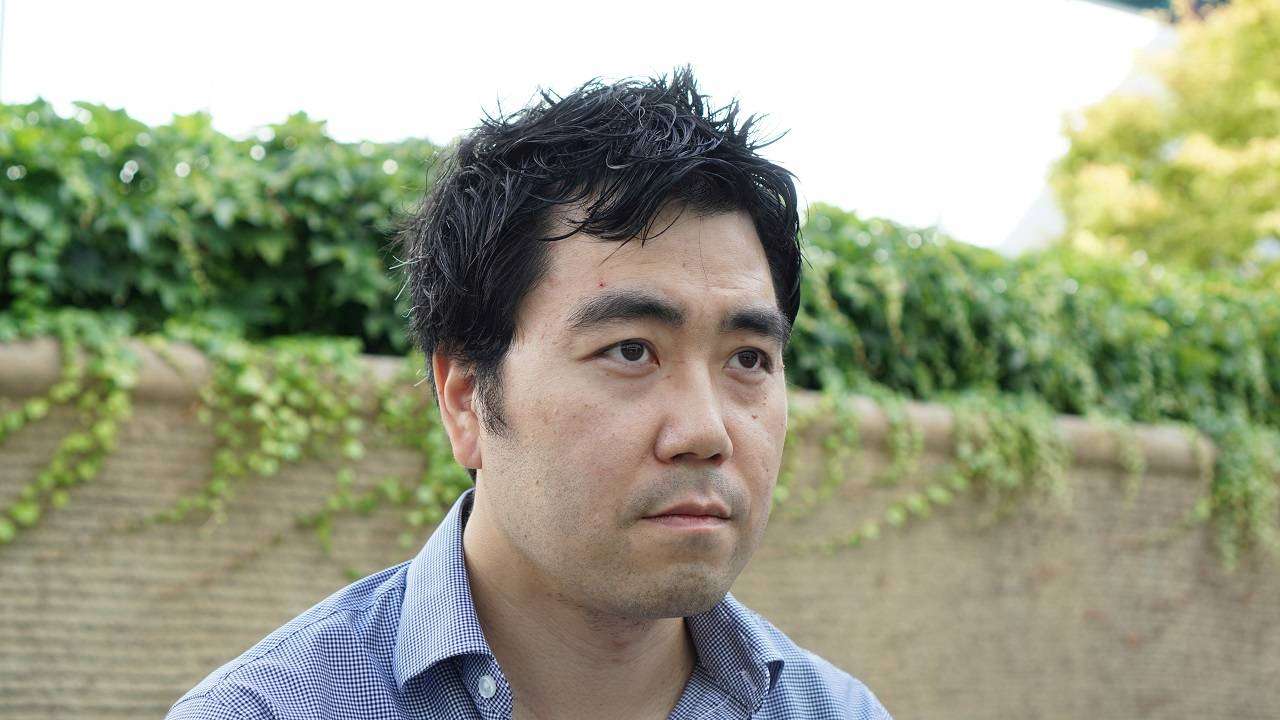}} 
		\hfill
		\subfloat{\includegraphics[trim=2.5cm 0.6cm 2.3cm 0cm, clip=true,width=.16\linewidth]{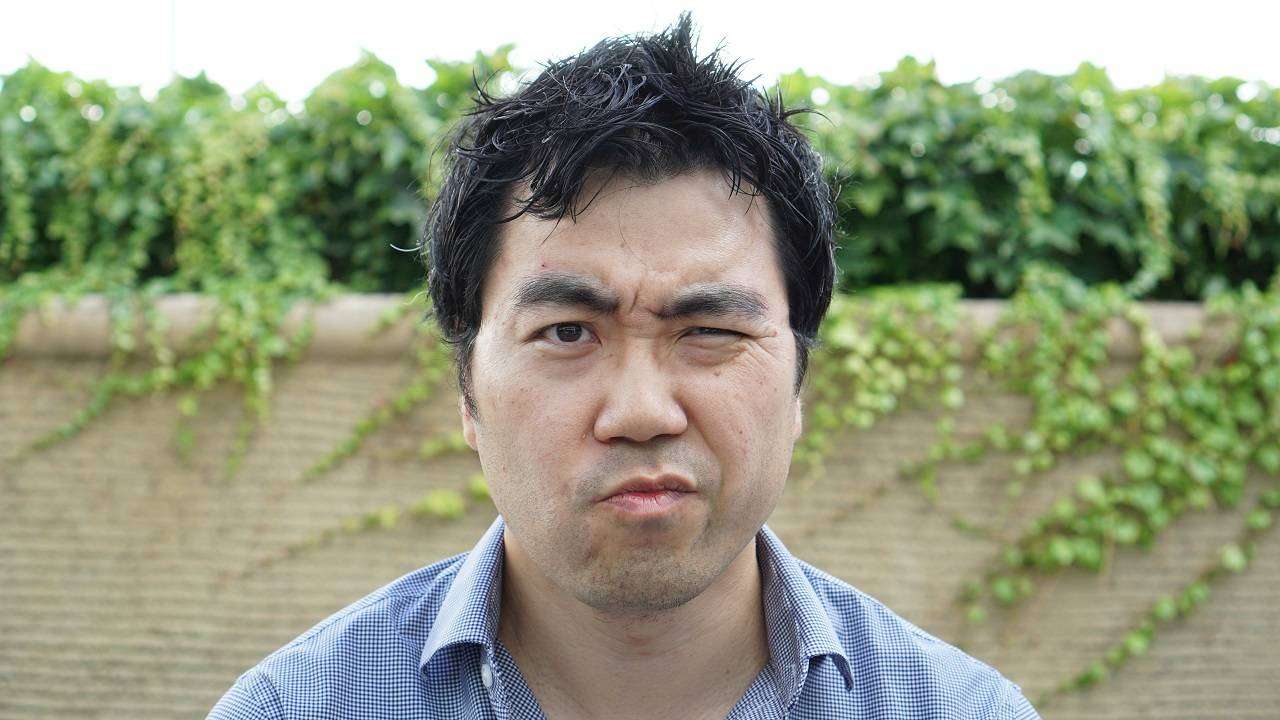}}
		\hfill
		\subfloat{\includegraphics[trim=2.5cm 0.3cm 2.3cm 0.3cm, clip=true,width=.16\linewidth]{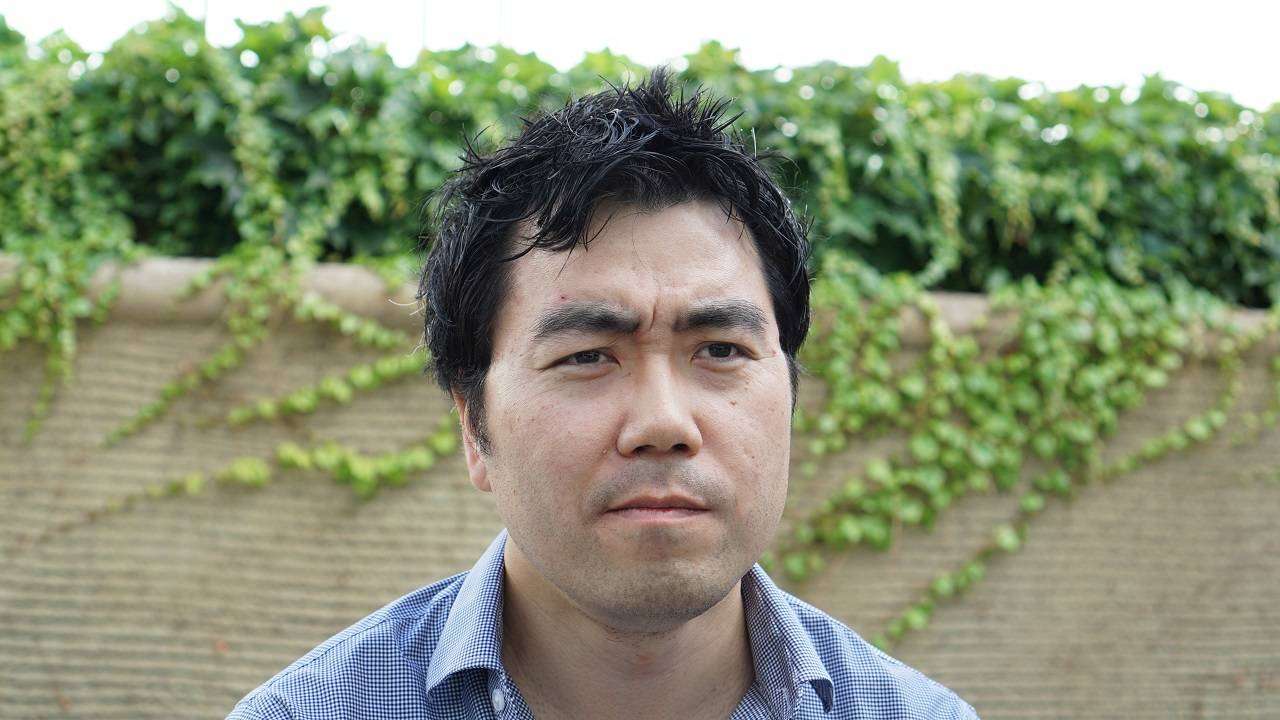}}
		\hfill
		\subfloat{\includegraphics[trim=2.5cm 0.6cm 2.3cm 0cm, clip=true,width=.16\linewidth]{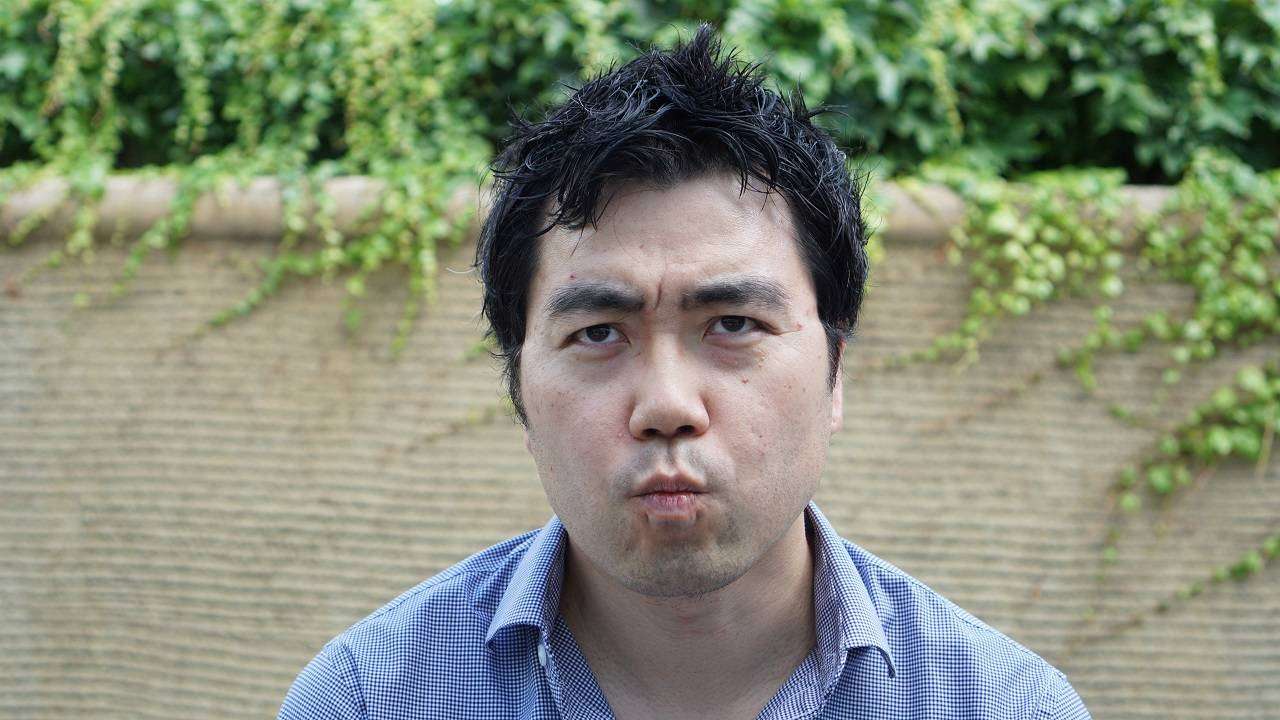}} 
		\hfill
		\subfloat{\includegraphics[trim=2.5cm 0.6cm 2.3cm 0cm, clip=true,width=.16\linewidth]{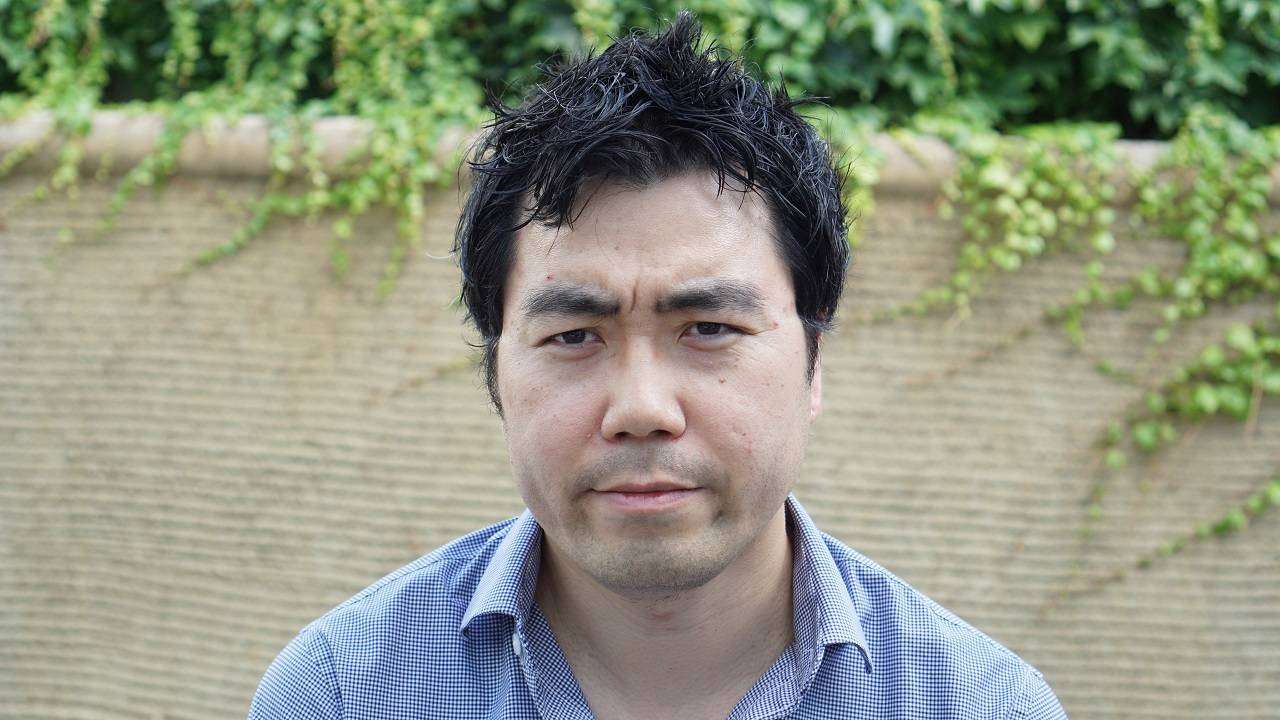}} 
		\\
		\vspaceSame
		\subfloat{\parbox[t]{.02\linewidth}{\begin{sideways}\centering \footnotesize \qquad \, 3D point cloud\end{sideways}}}
		\hfill
		\subfloat{\includegraphics[trim=14cm 6cm 13cm 4cm, clip=true,width=.16\linewidth]{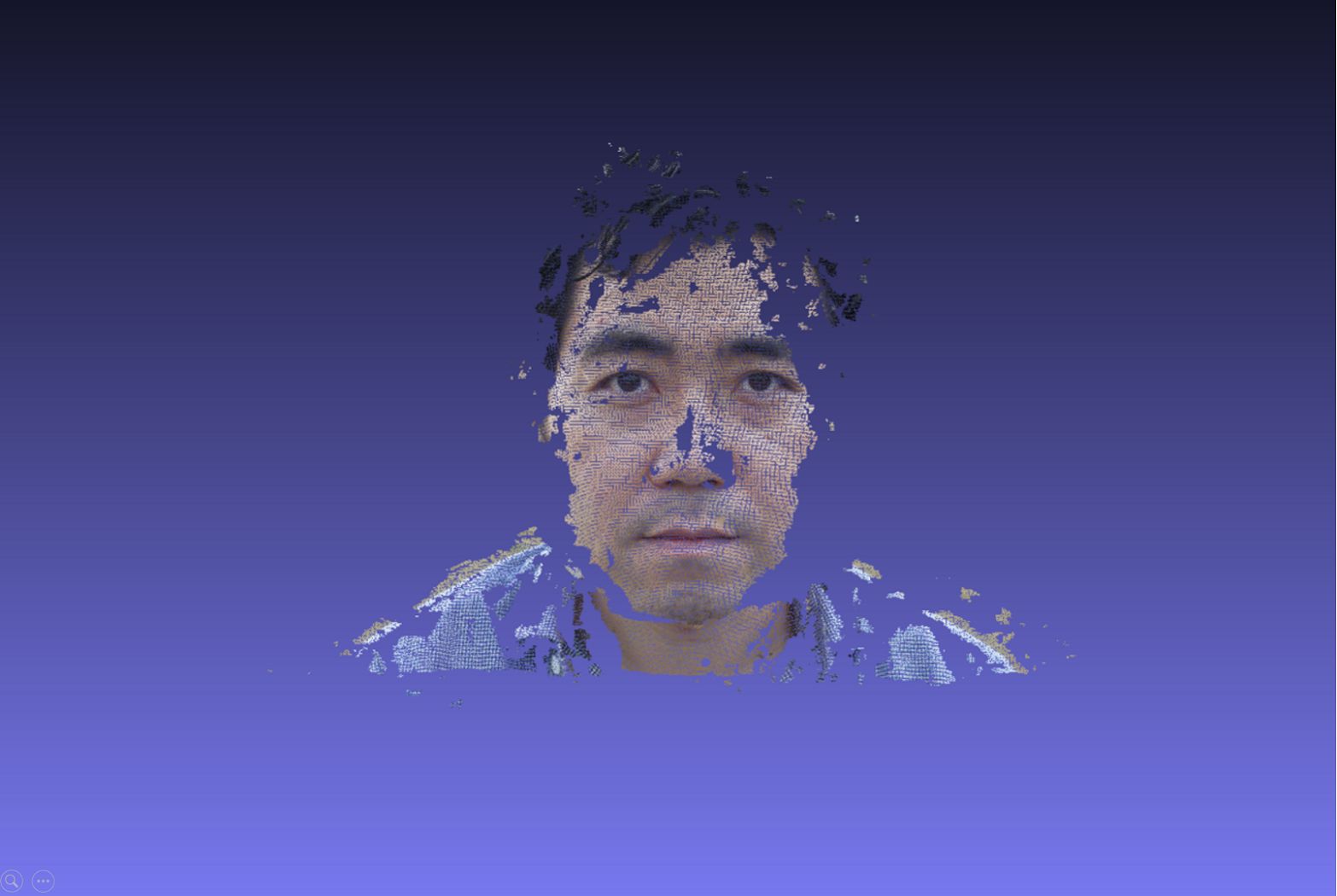}}
		\hfill
		\subfloat{\includegraphics[trim=14cm 6cm 13cm 4cm, clip=true,width=.16\linewidth]{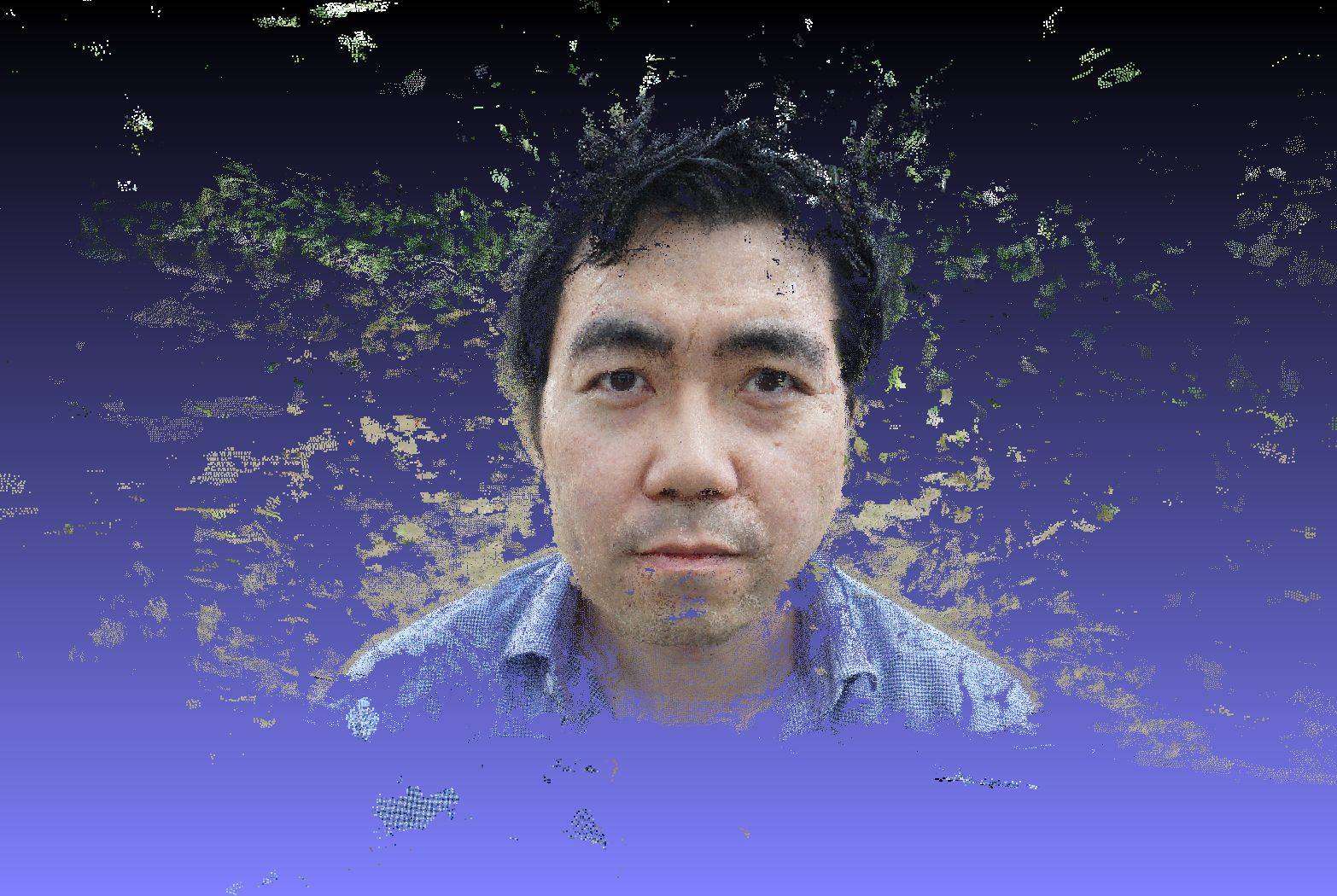}}
		\hfill
		\subfloat{\includegraphics[trim=14cm 6cm 13cm 4cm, clip=true,width=.16\linewidth]{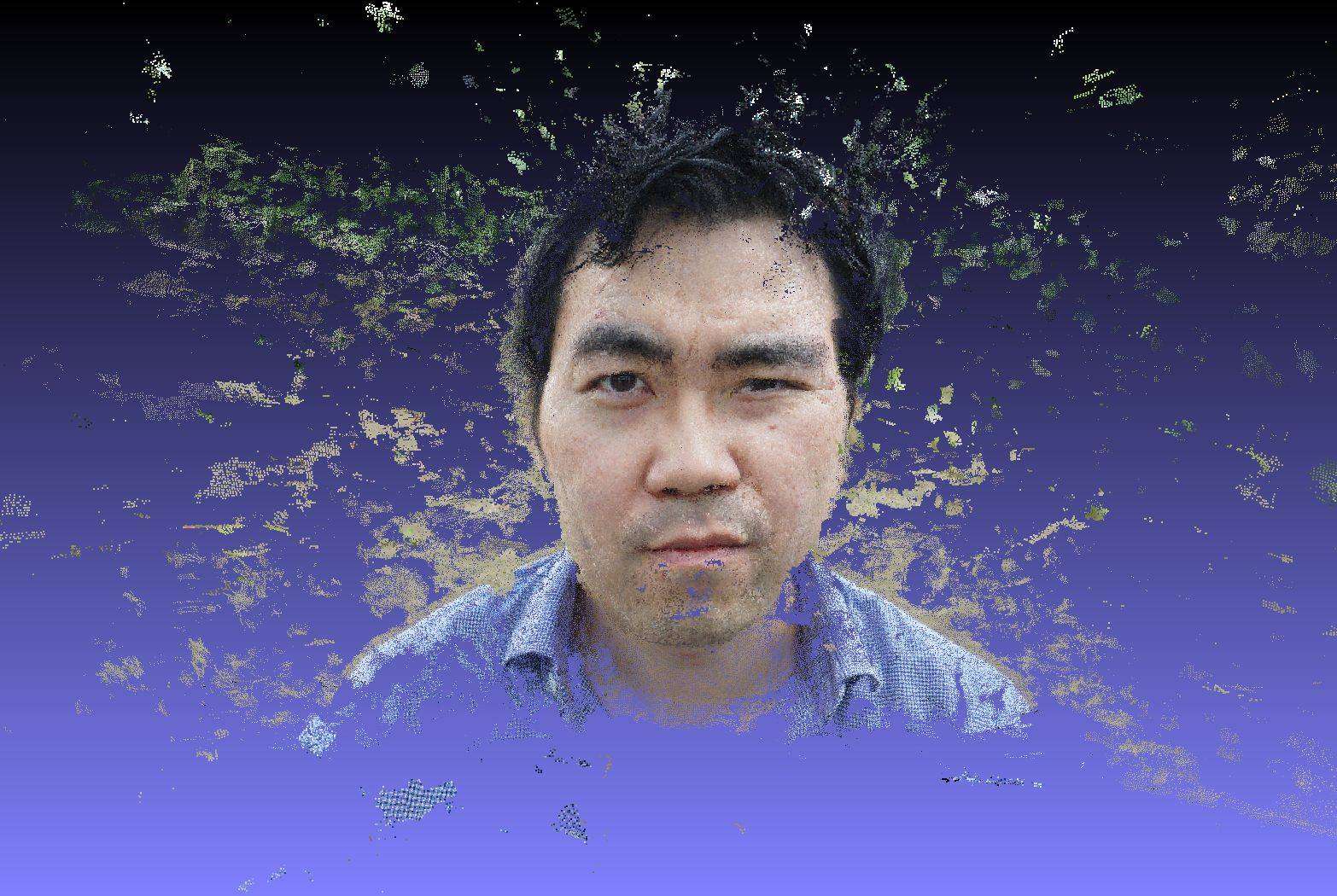}}
		\hfill
		\subfloat{\includegraphics[trim=14cm 6cm 13cm 4cm, clip=true,width=.16\linewidth]{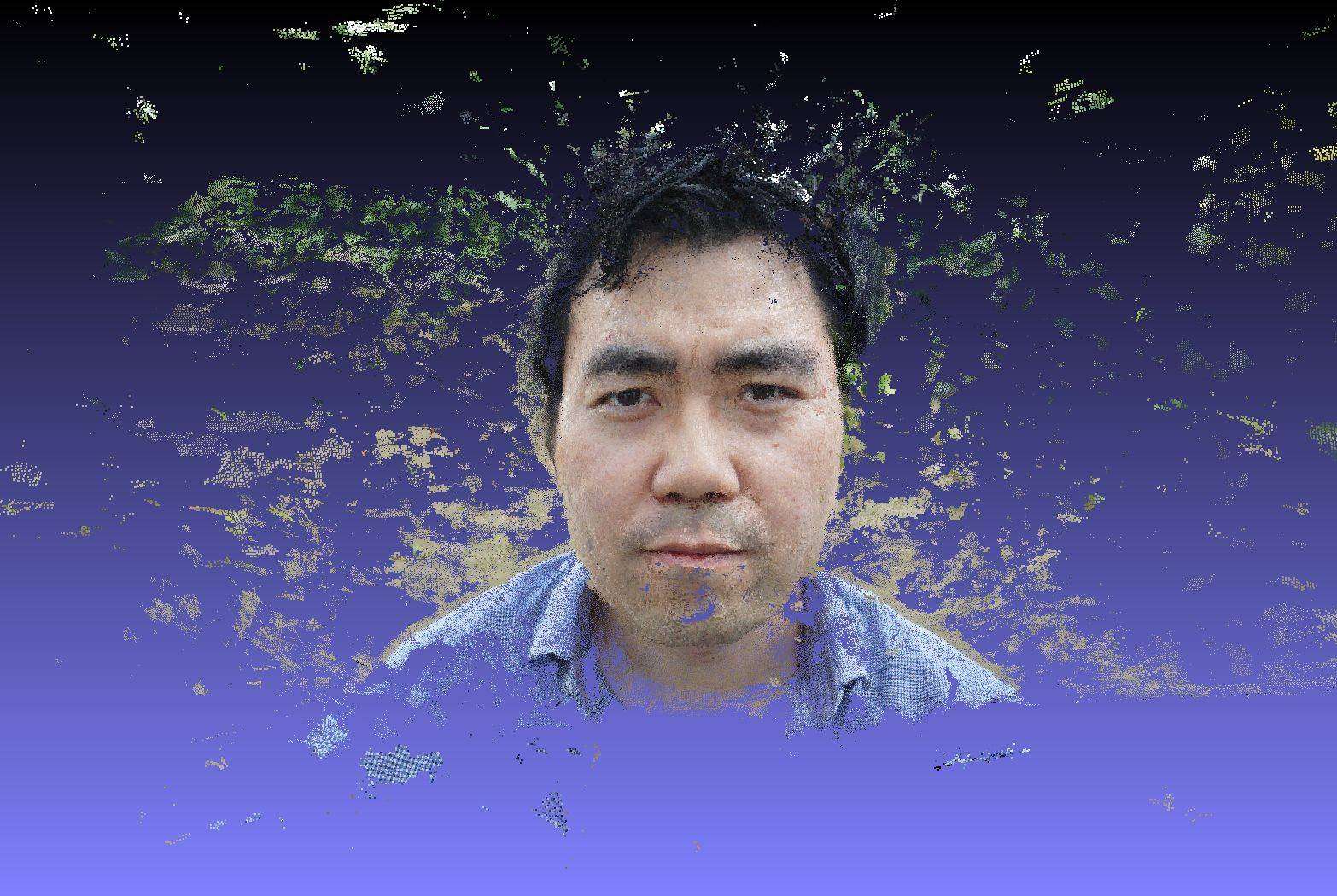}}
		\hfill
		\subfloat{\includegraphics[trim=14cm 6cm 13cm 4cm, clip=true,width=.16\linewidth]{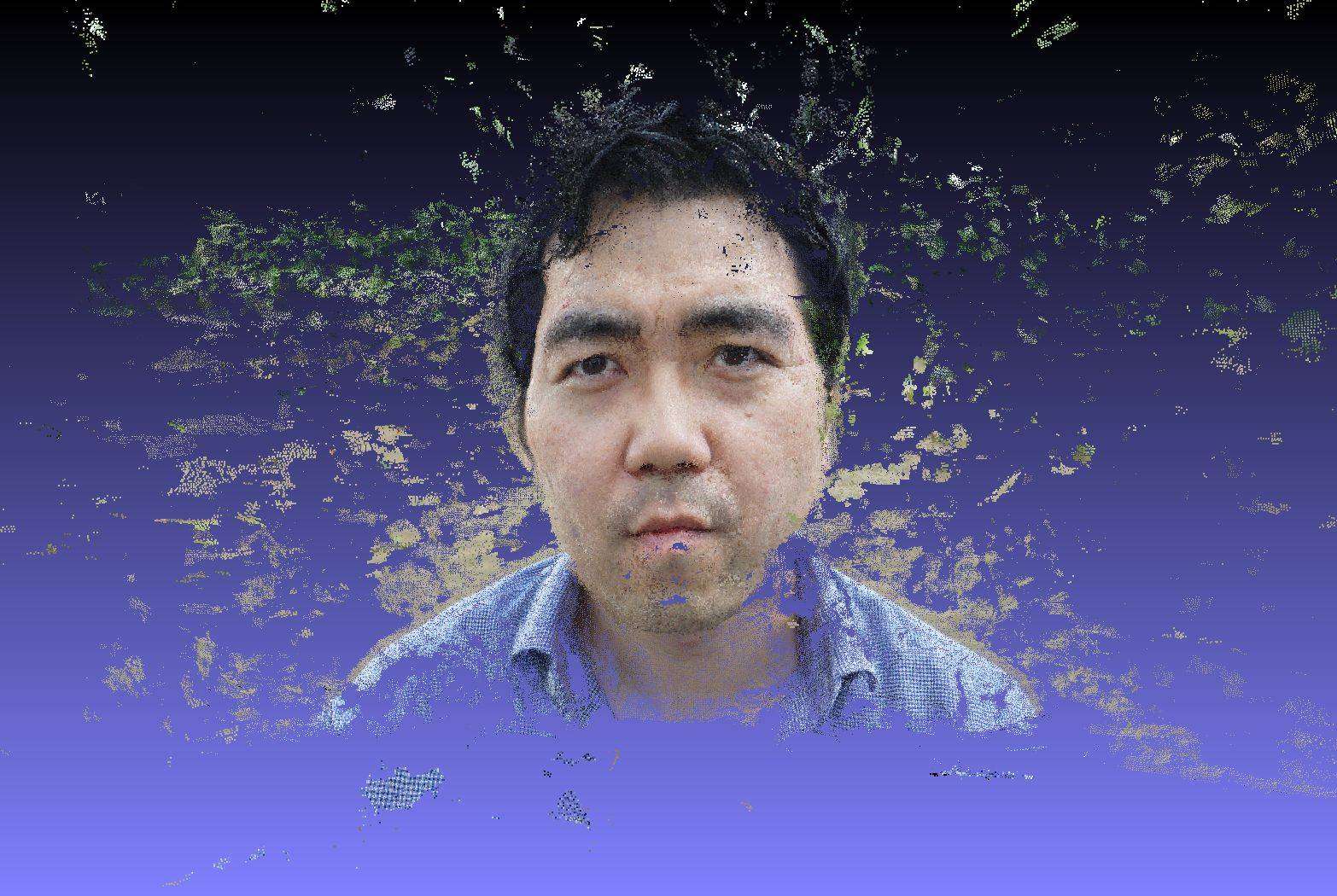}}
		\hfill
		\subfloat{\includegraphics[trim=14cm 6cm 13cm 4cm, clip=true,width=.16\linewidth]{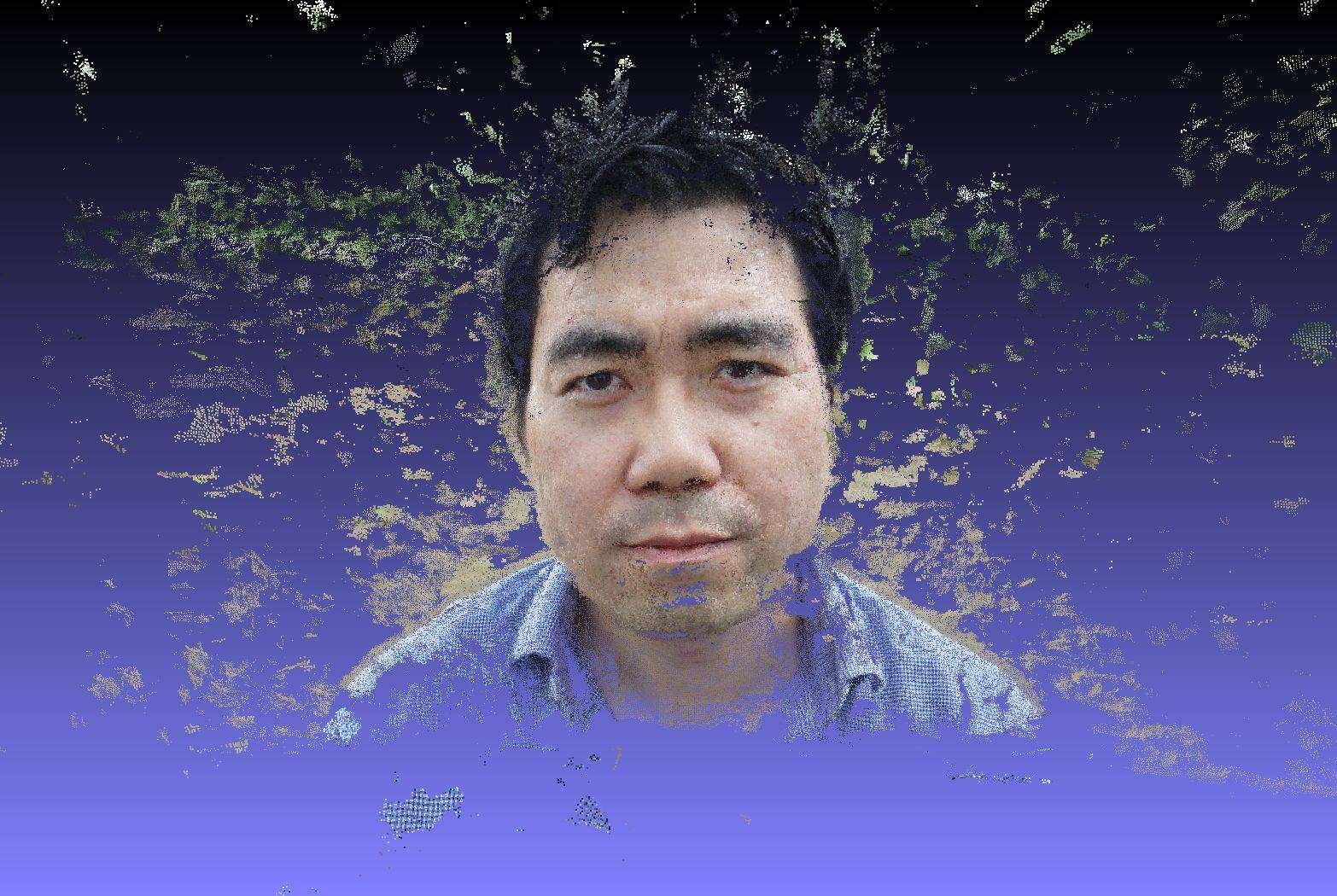}} 
		\\
		\vspaceScene
		\subfloat{\parbox[t]{.02\linewidth}{\begin{sideways}\centering \footnotesize \, \, Input images\end{sideways}}}
		\hfill
		\subfloat{\includegraphics[trim=1cm   0cm 1cm  0cm, clip=true,width=.16\linewidth]{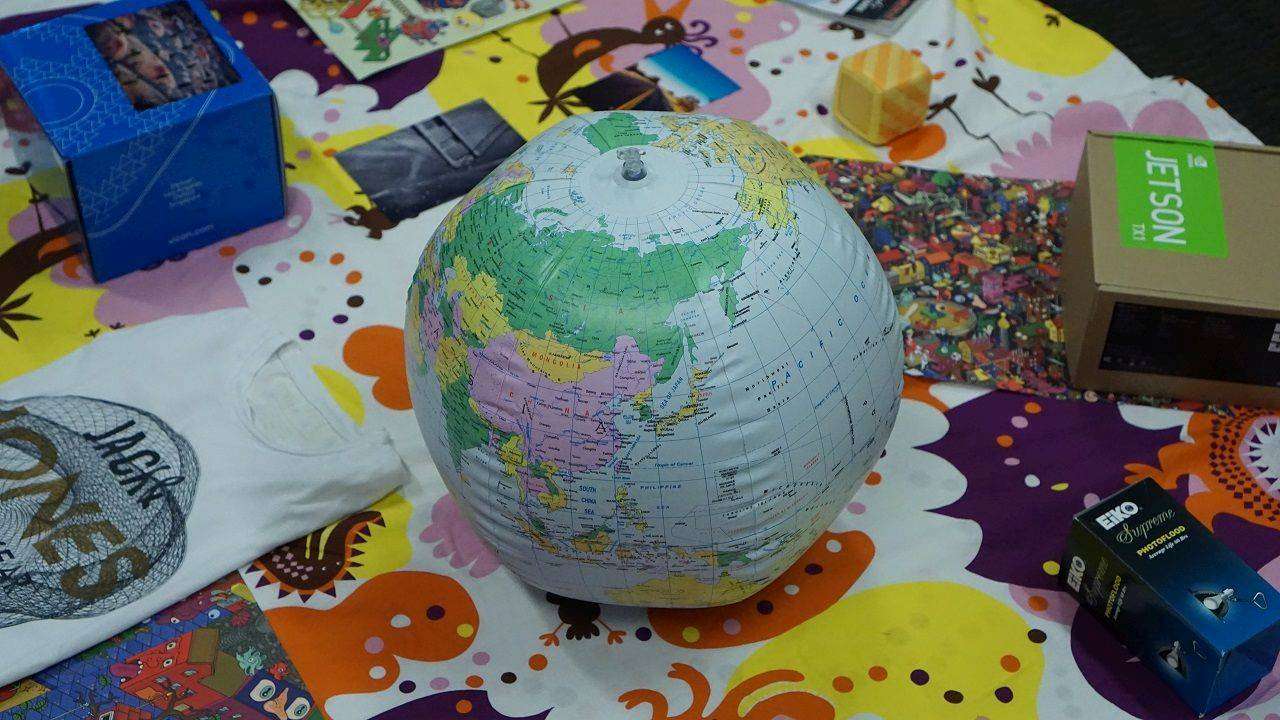}} %
		\hfill
		\subfloat{\includegraphics[trim=1cm   0cm 1cm  0cm, clip=true,width=.16\linewidth]{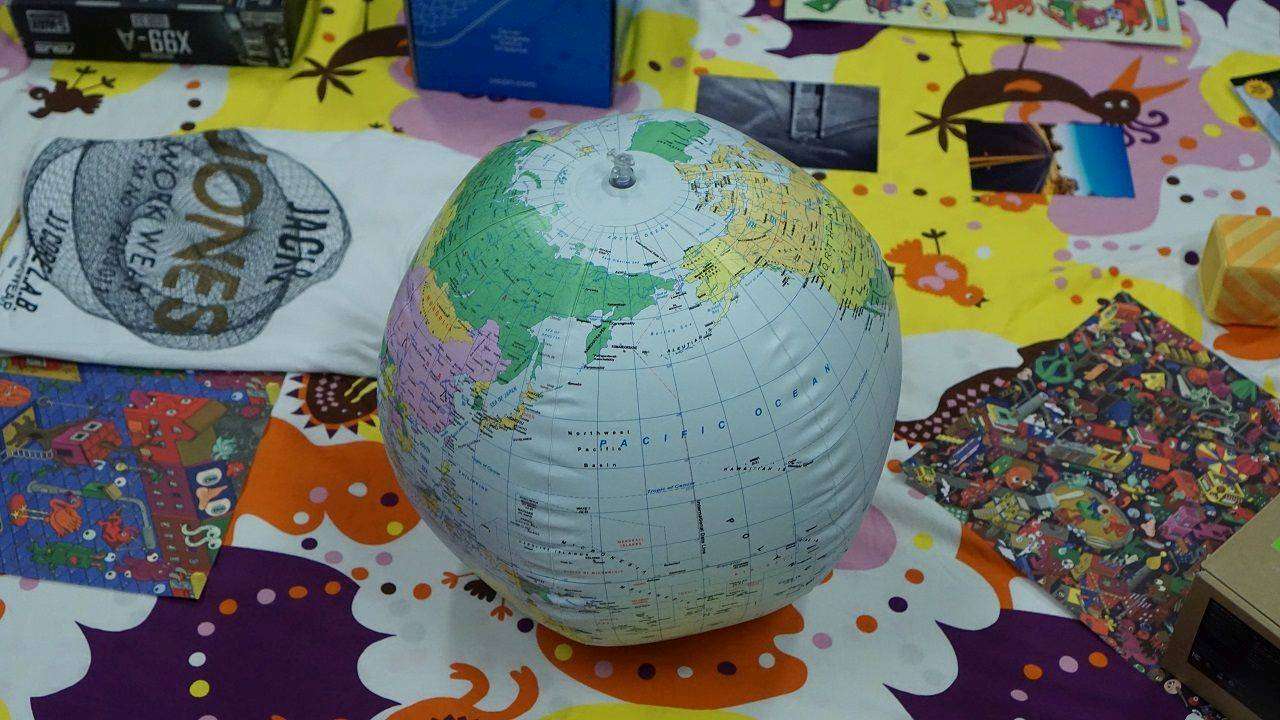}} 
		\hfill
		\subfloat{\includegraphics[trim=1cm   0cm 1cm  0cm, clip=true,width=.16\linewidth]{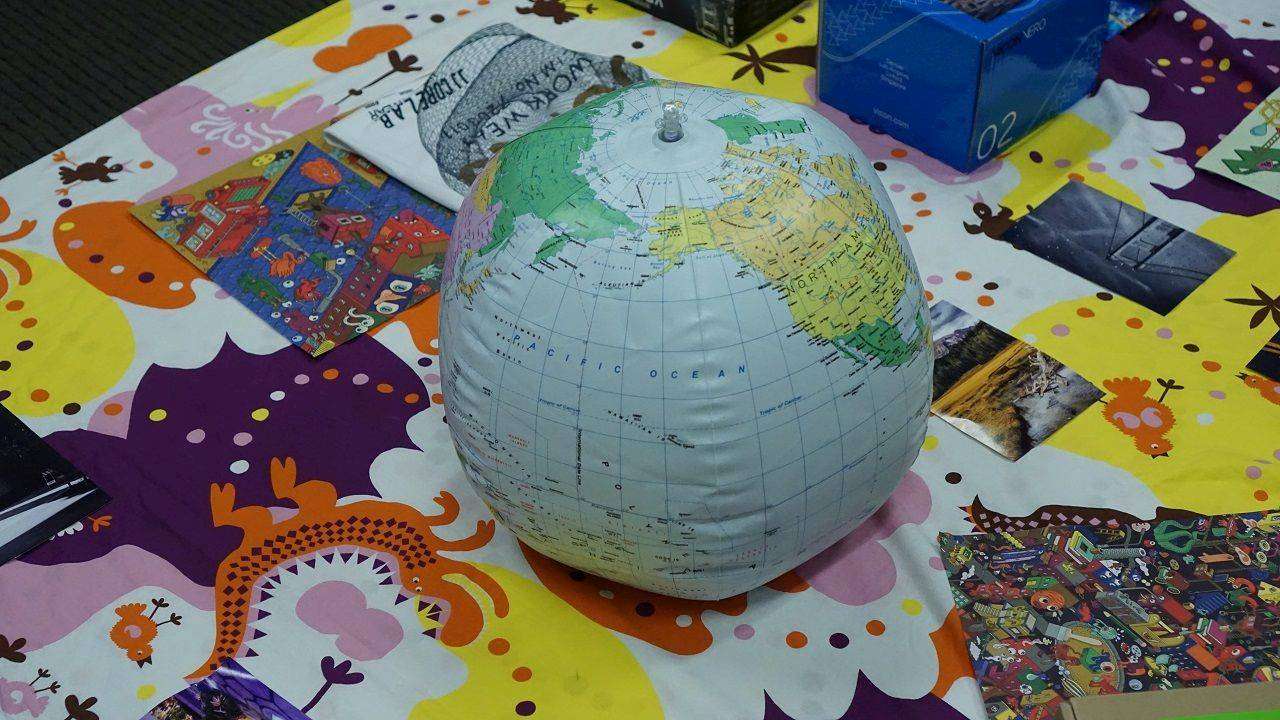}} %
		\hfill
		\subfloat{\includegraphics[trim=1cm   0cm 1cm  0cm, clip=true,width=.16\linewidth]{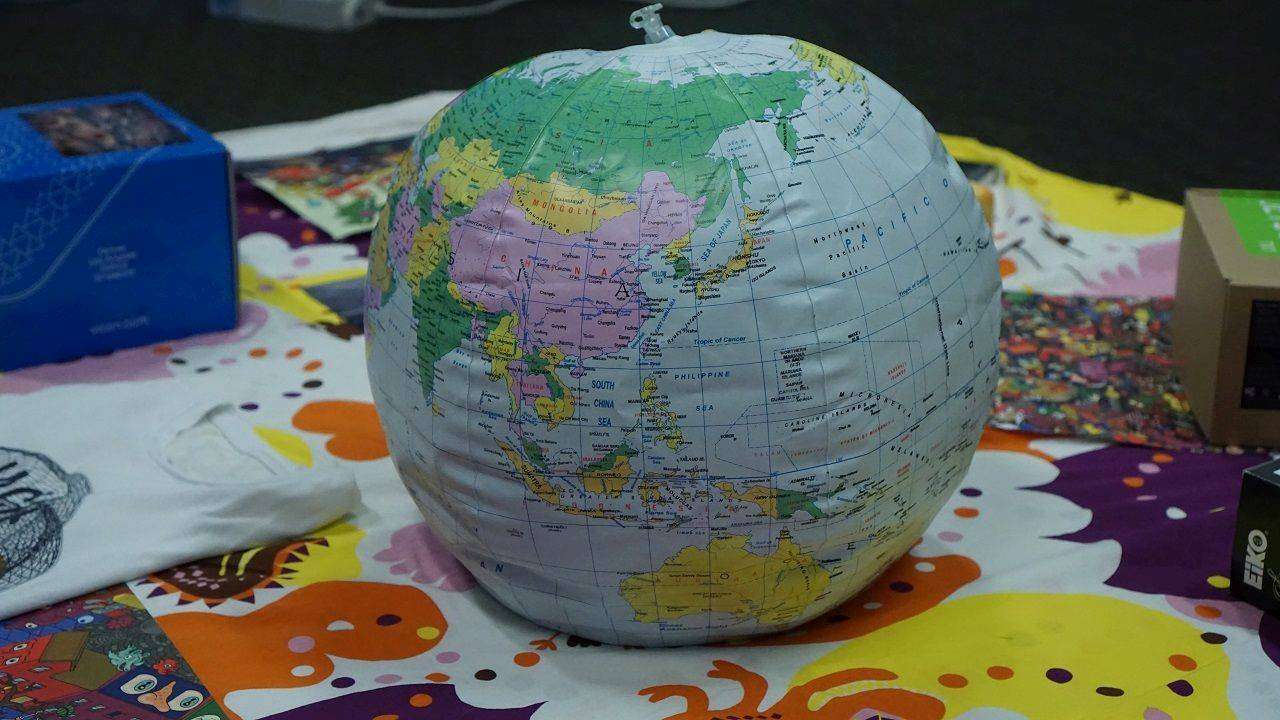}} %
		\hfill
		\subfloat{\includegraphics[trim=1cm   0cm 1cm  0cm, clip=true,width=.16\linewidth]{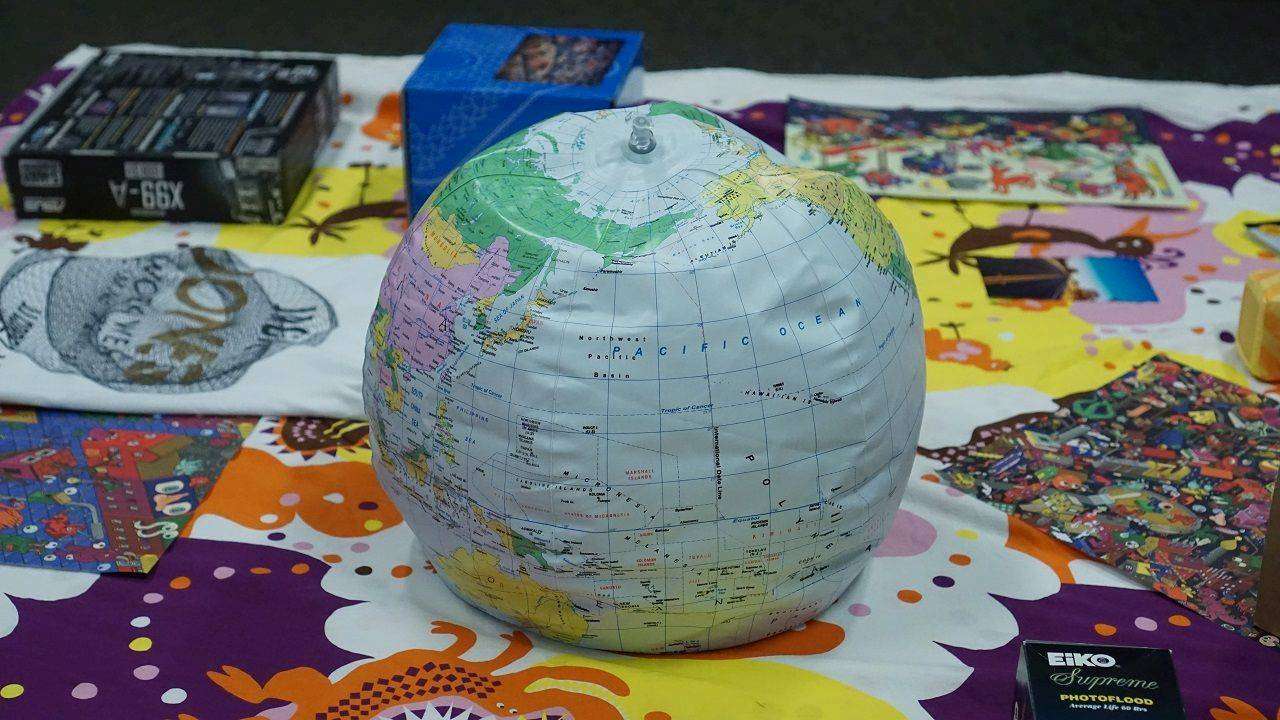}} %
		\hfill
		\subfloat{\includegraphics[trim=1cm   0cm 1cm  0cm, clip=true,width=.16\linewidth]{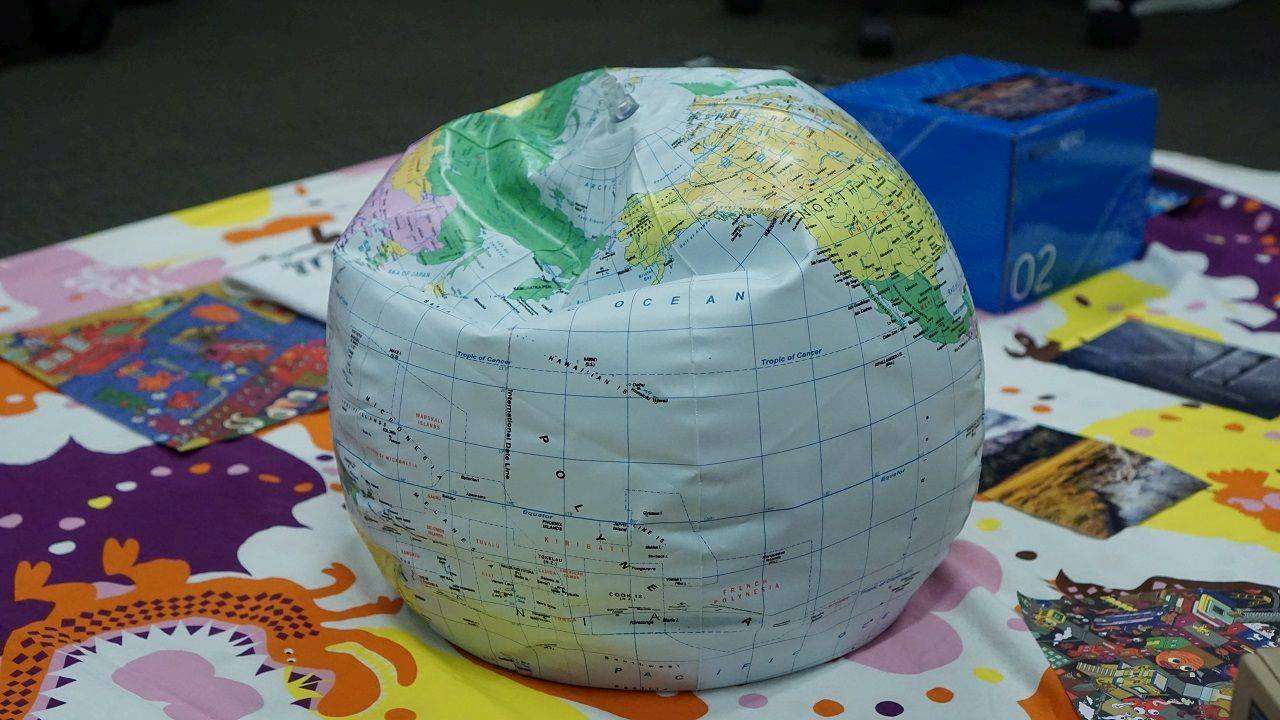}} 
		\\
		\vspaceSame
		\subfloat{\parbox[t]{.02\linewidth}{\begin{sideways}\centering \footnotesize \, 3D point cloud \end{sideways}}}
		\hfill
		\subfloat{\includegraphics[trim=12cm   8cm 12cm  8cm, clip=true,width=.16\linewidth]{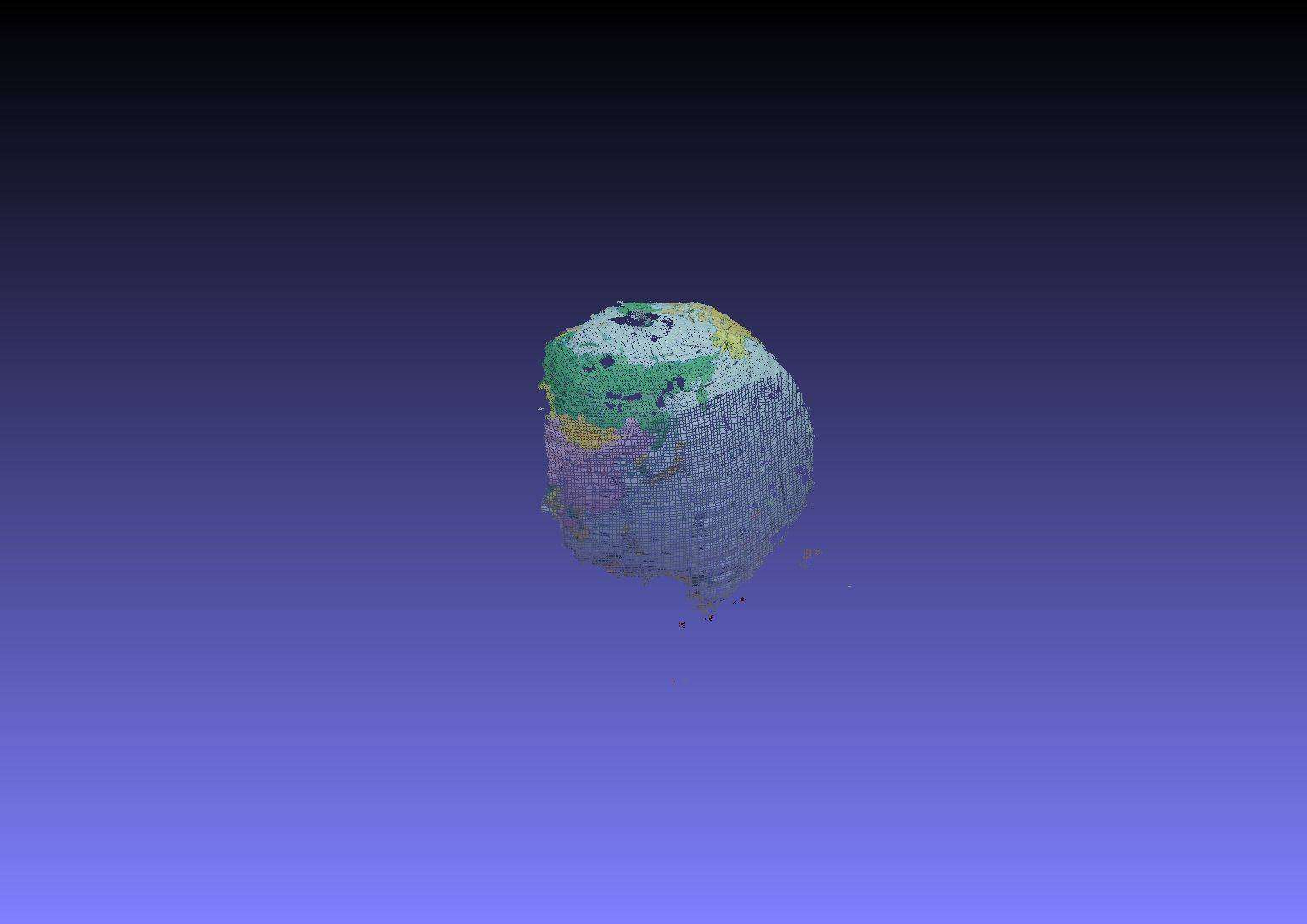}}
		\hfill
		\subfloat{\includegraphics[trim=12cm   8cm 12cm  8cm, clip=true,width=.16\linewidth]{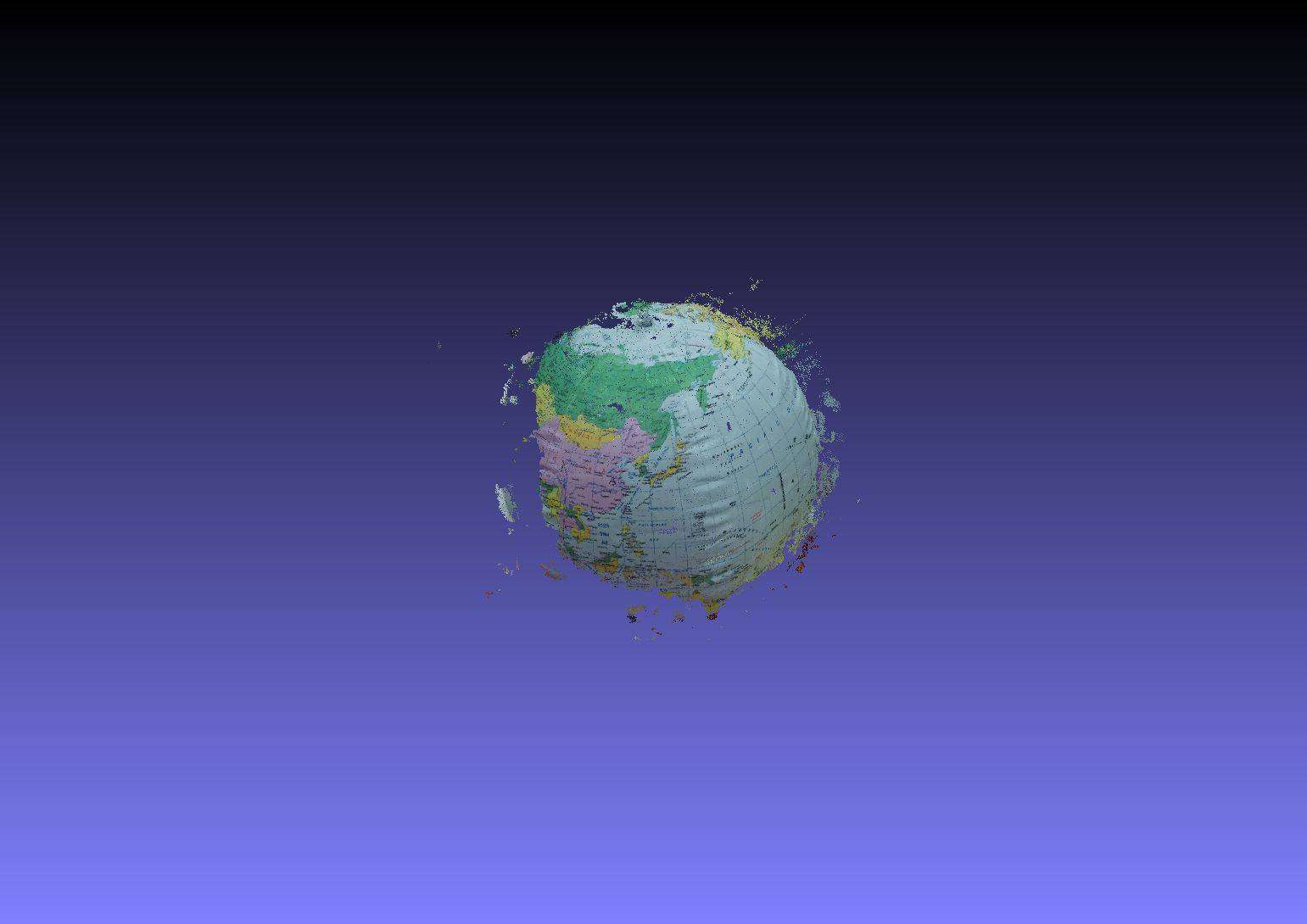}}
		\hfill
		\subfloat{\includegraphics[trim=12cm   8cm 12cm  8cm, clip=true,width=.16\linewidth]{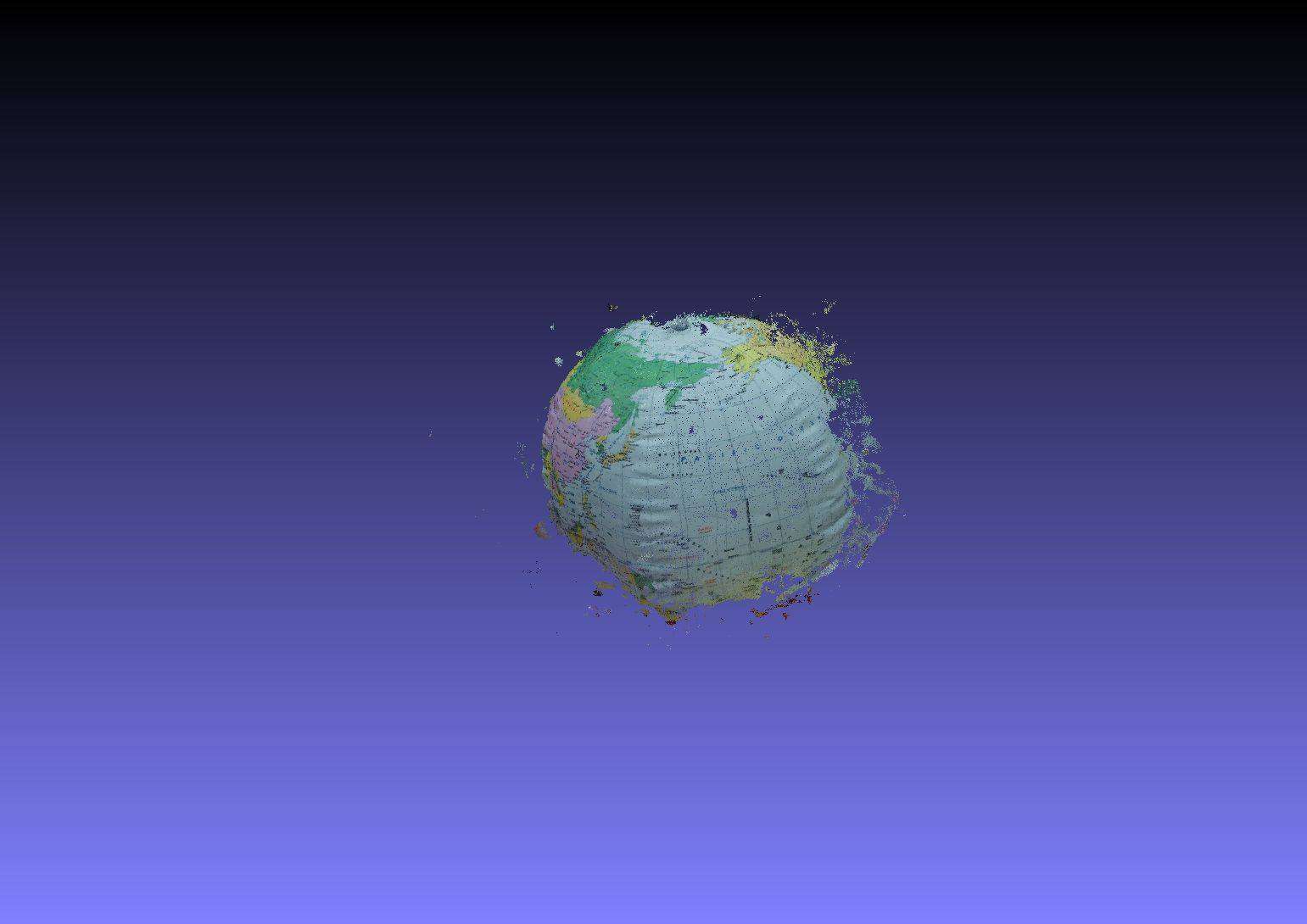}}
		\hfill
		\subfloat{\includegraphics[trim=12cm   8cm 12cm  8cm, clip=true,width=.16\linewidth]{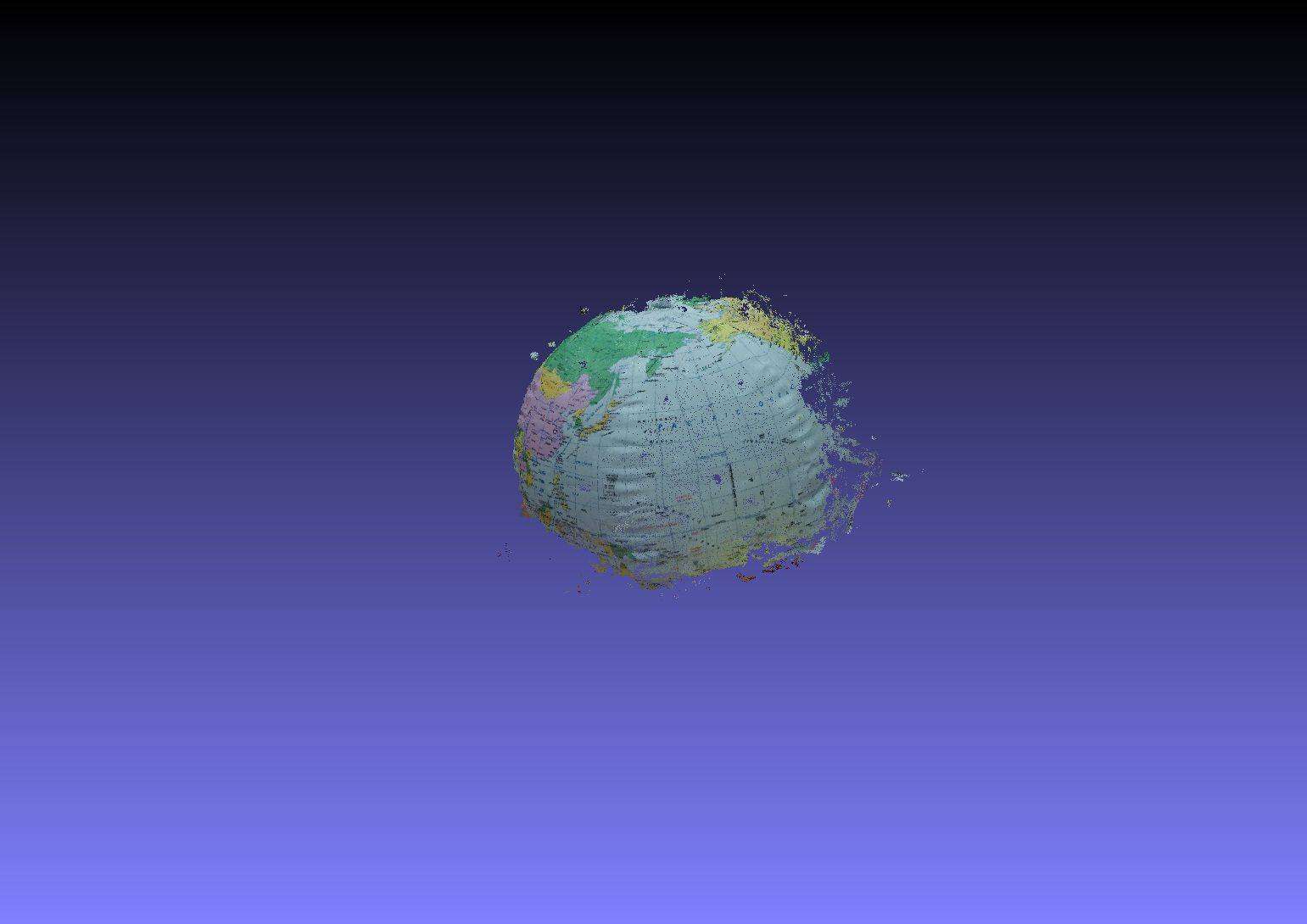}}
		\hfill
		\subfloat{\includegraphics[trim=12cm   8cm 12cm  8cm, clip=true,width=.16\linewidth]{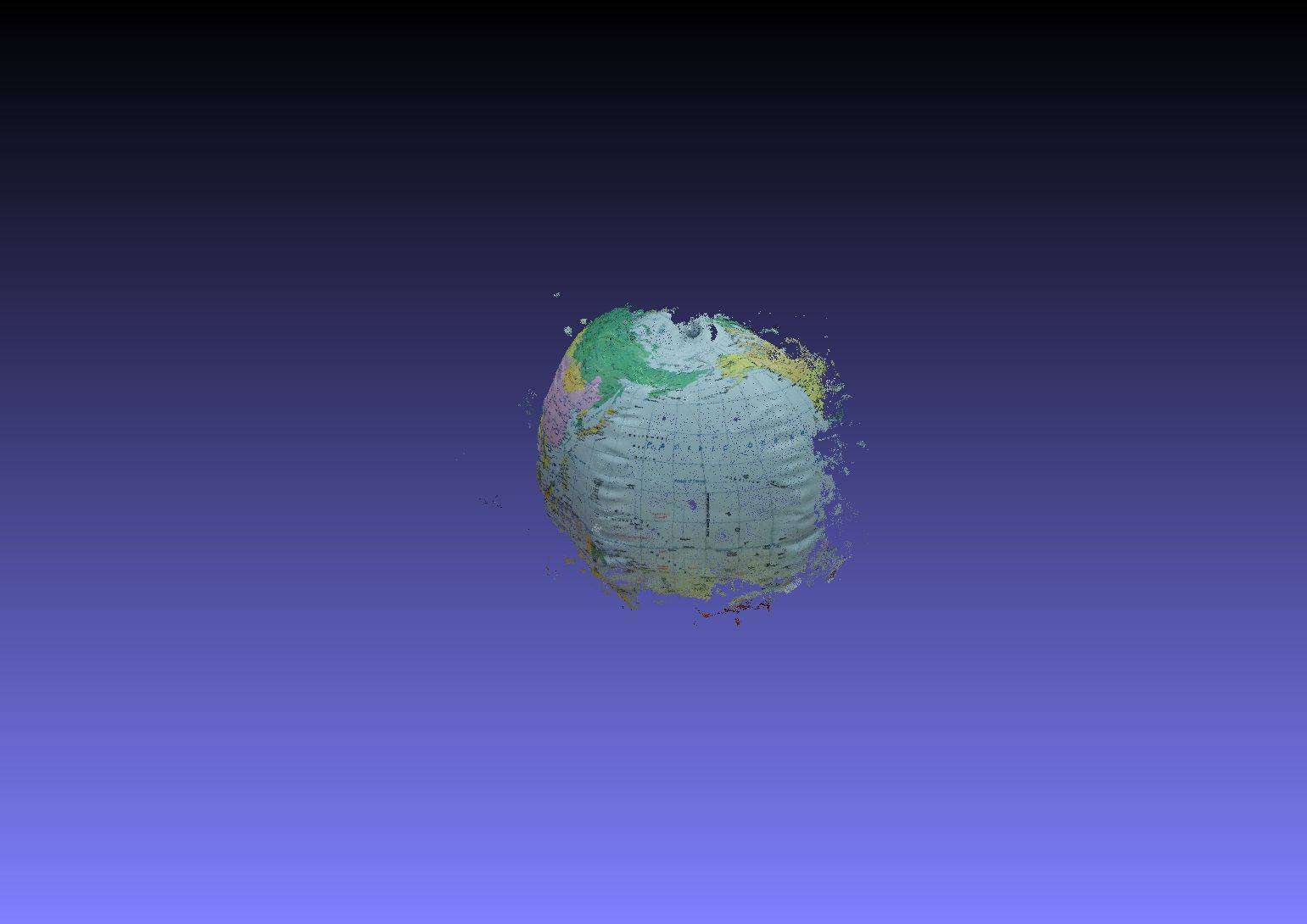}}
		\hfill
		\subfloat{\includegraphics[trim=12cm   8cm 12cm  8cm, clip=true,width=.16\linewidth]{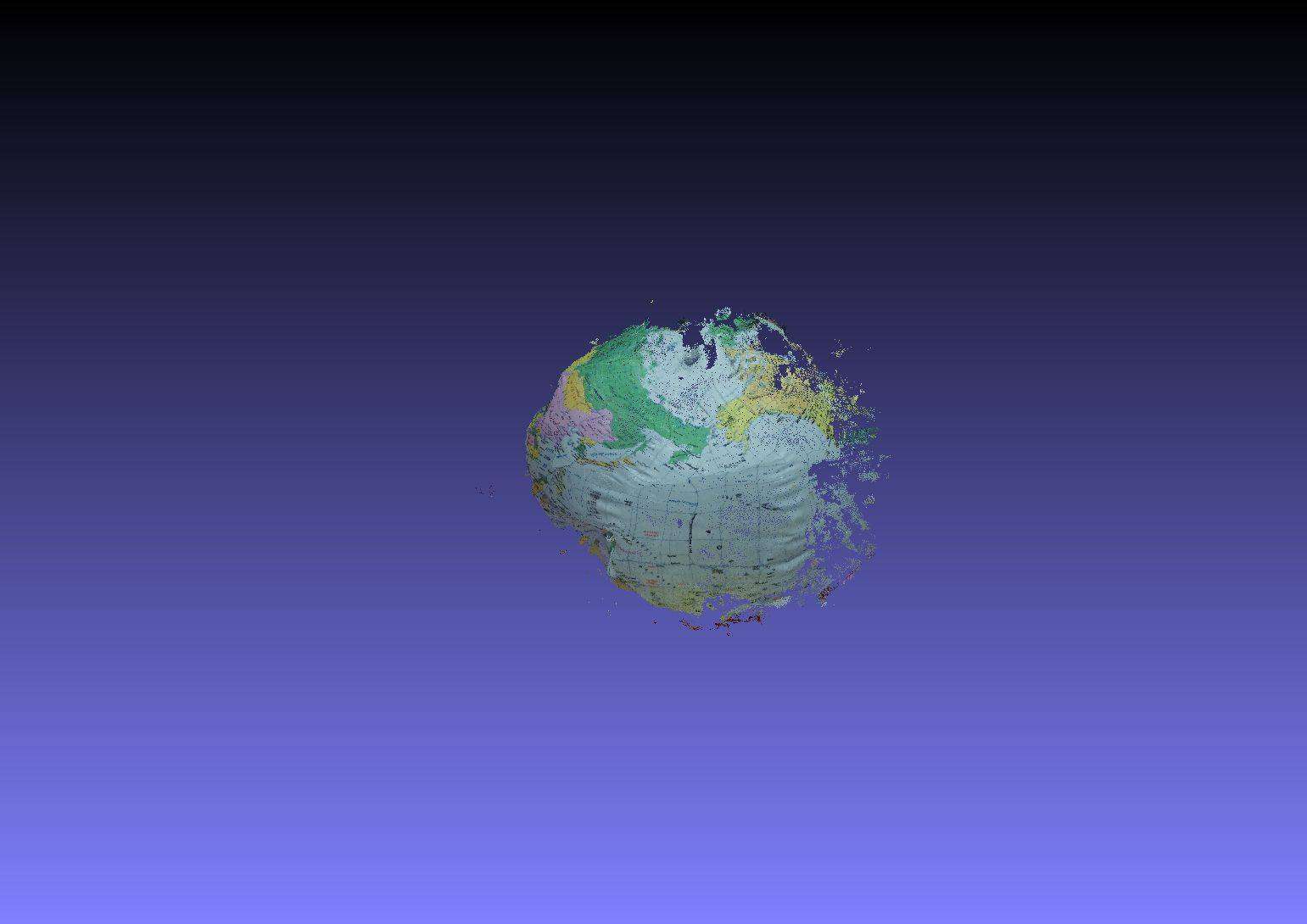}}
		\\
		\vspaceScene
		\subfloat{\parbox[t]{.02\linewidth}{\begin{sideways}\centering \footnotesize \quad\ Input images\end{sideways}}}
		\hfill
		\subfloat{\includegraphics[trim=2.4cm 0.5cm 1.6cm 0.5cm, clip=true,width=.16\linewidth]{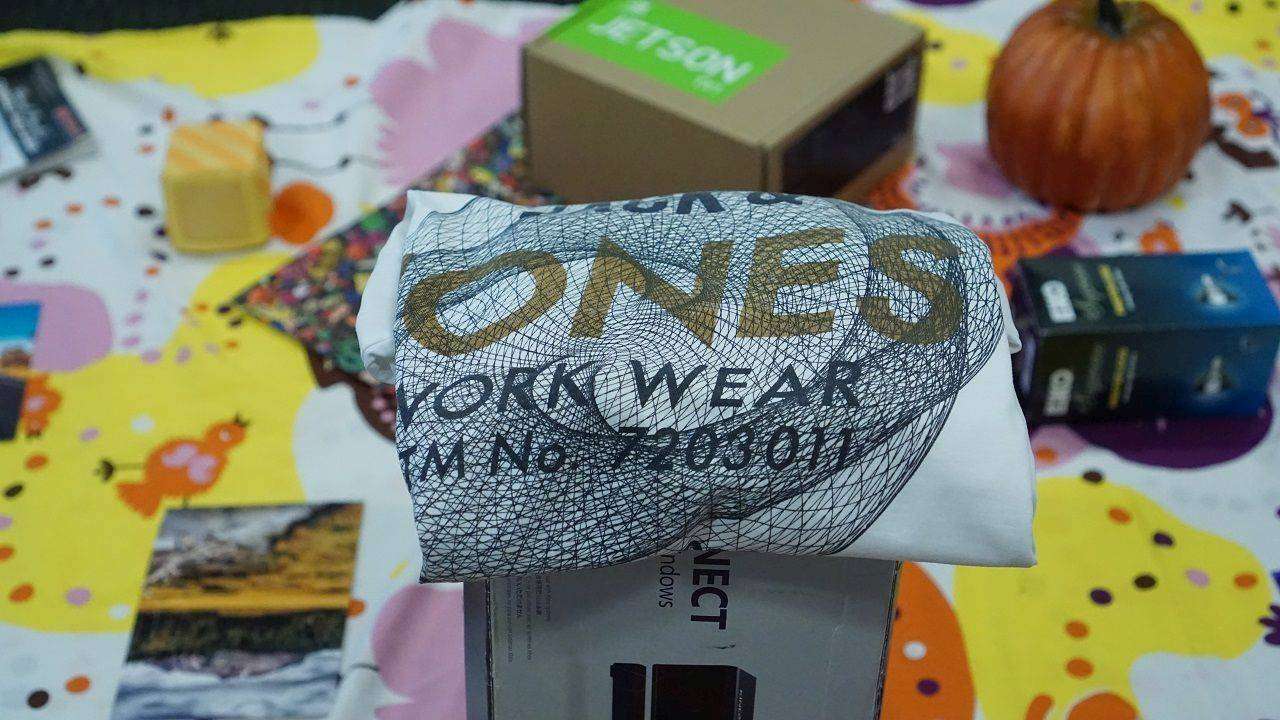}} %
		\hfill
		\subfloat{\includegraphics[trim=2.3cm 0.5cm 1.7cm 0.5cm, clip=true,width=.16\linewidth]{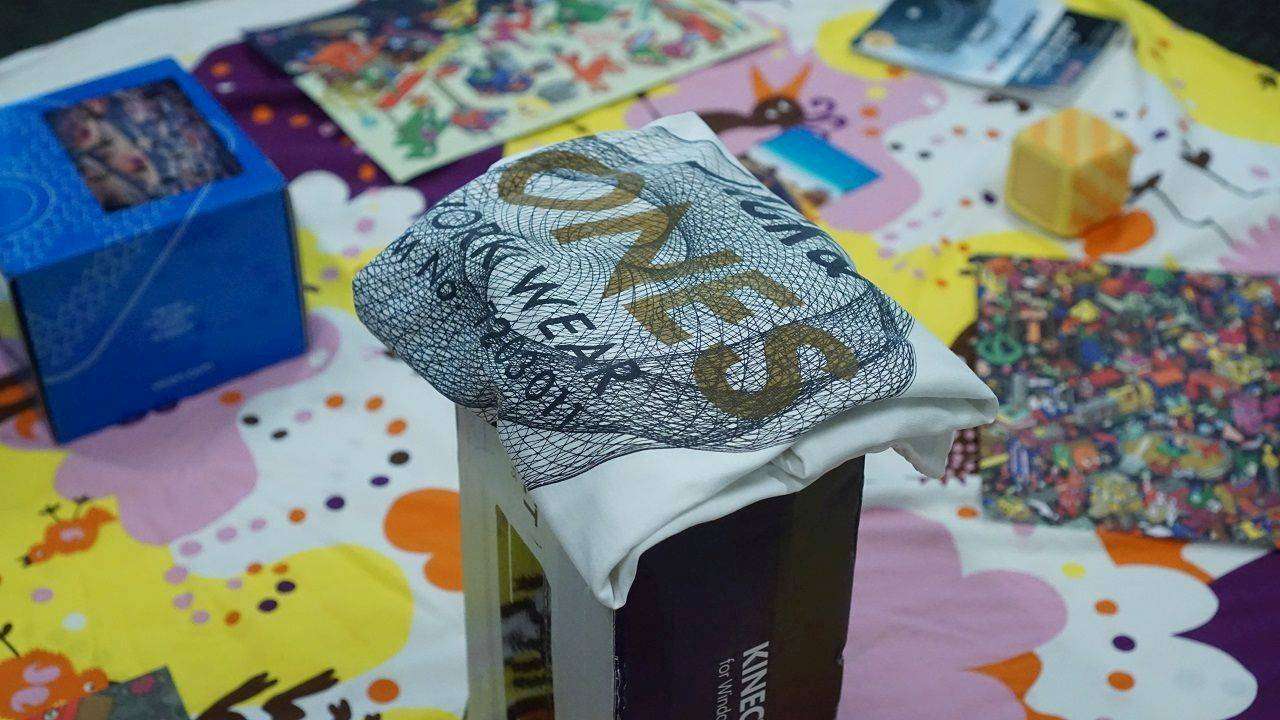}} 
		\hfill
		\subfloat{\includegraphics[trim=2.2cm 0.5cm 1.8cm 0.5cm, clip=true,width=.16\linewidth]{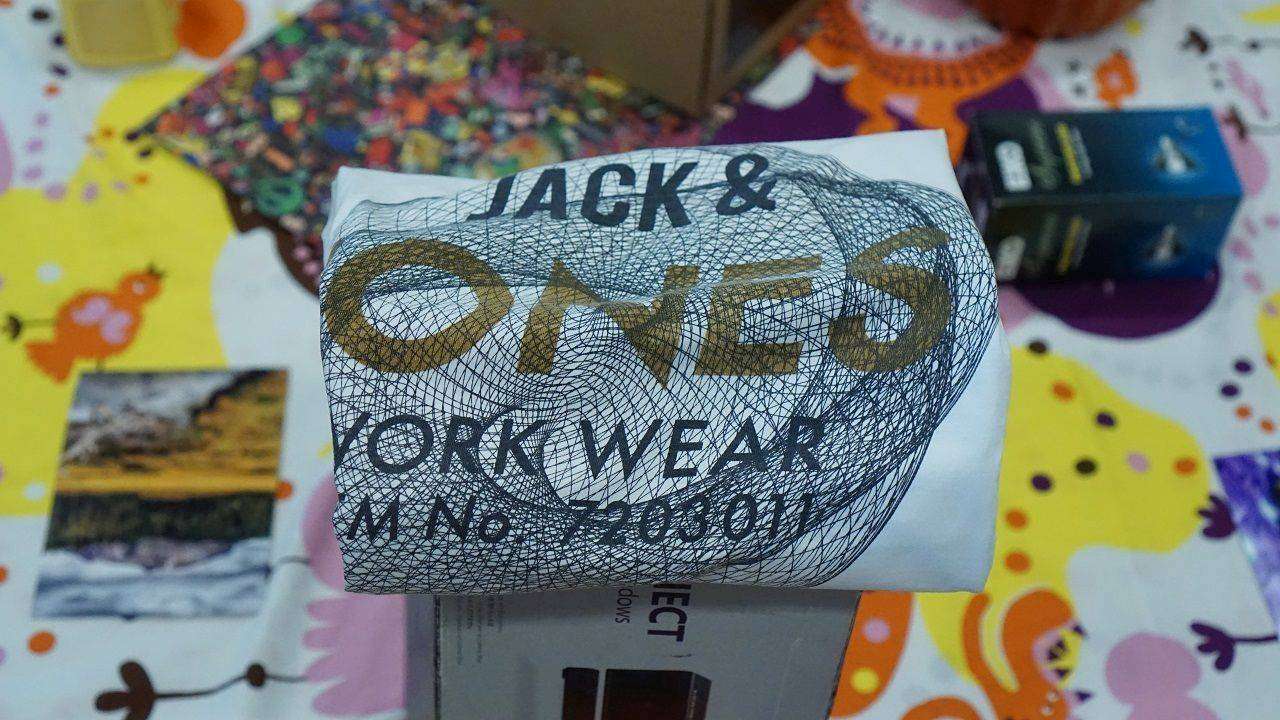}} %
		\hfill
		\subfloat{\includegraphics[trim=2cm 0.5cm 2cm 0.5cm, clip=true,width=.16\linewidth]{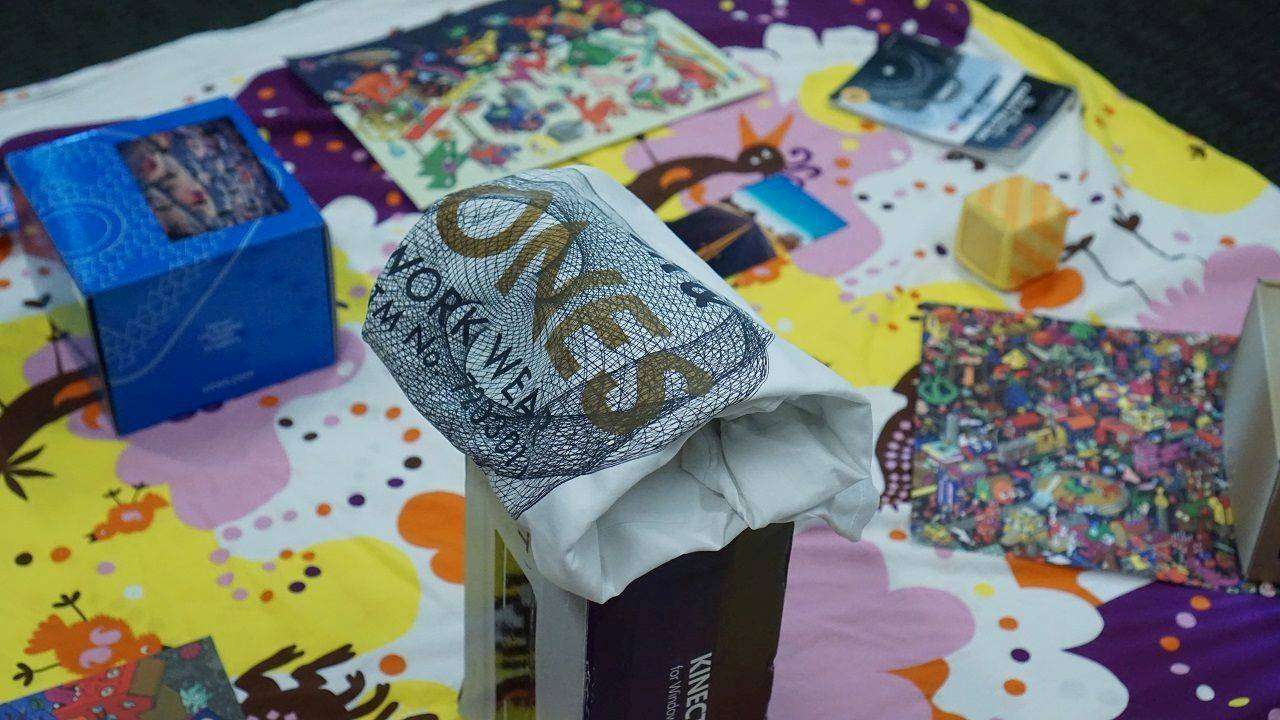}} %
		\hfill
		\subfloat{\includegraphics[trim=2.5cm 0.5cm 1.5cm 0.5cm, clip=true,width=.16\linewidth]{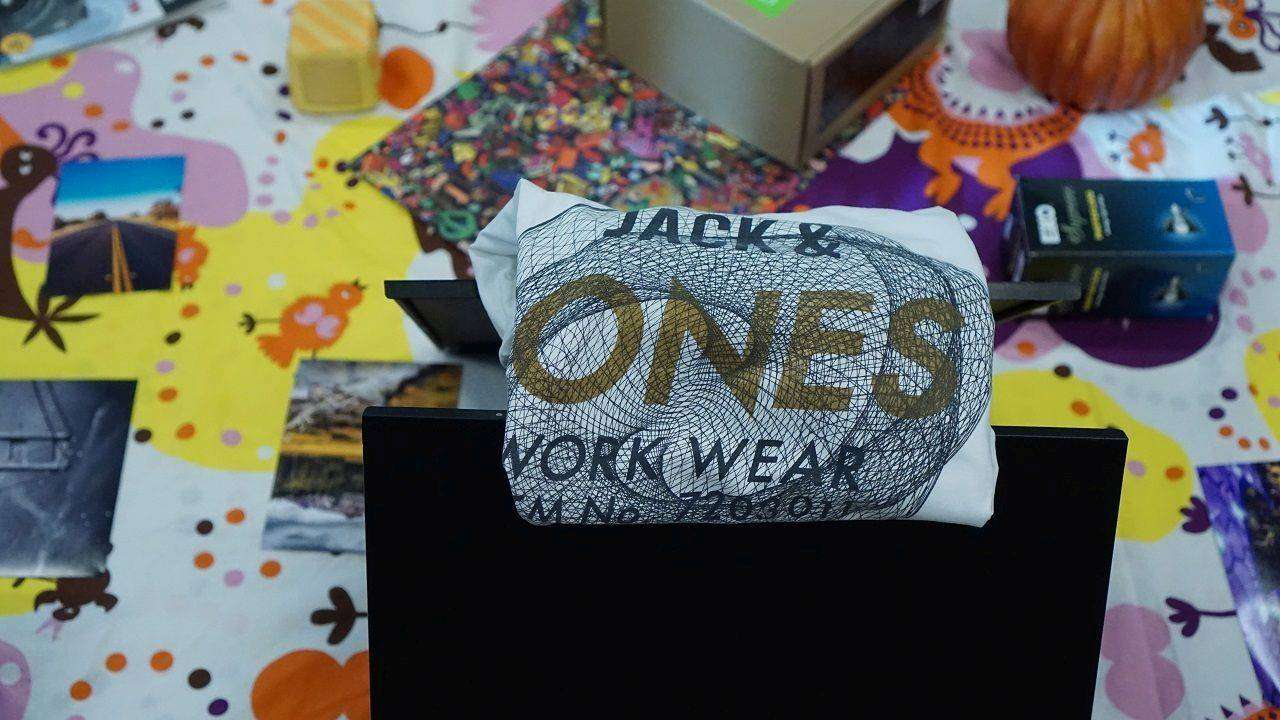}} %
		\hfill
		\subfloat{\includegraphics[trim=2.4cm 1cm 1.6cm 0cm, clip=true,width=.16\linewidth]{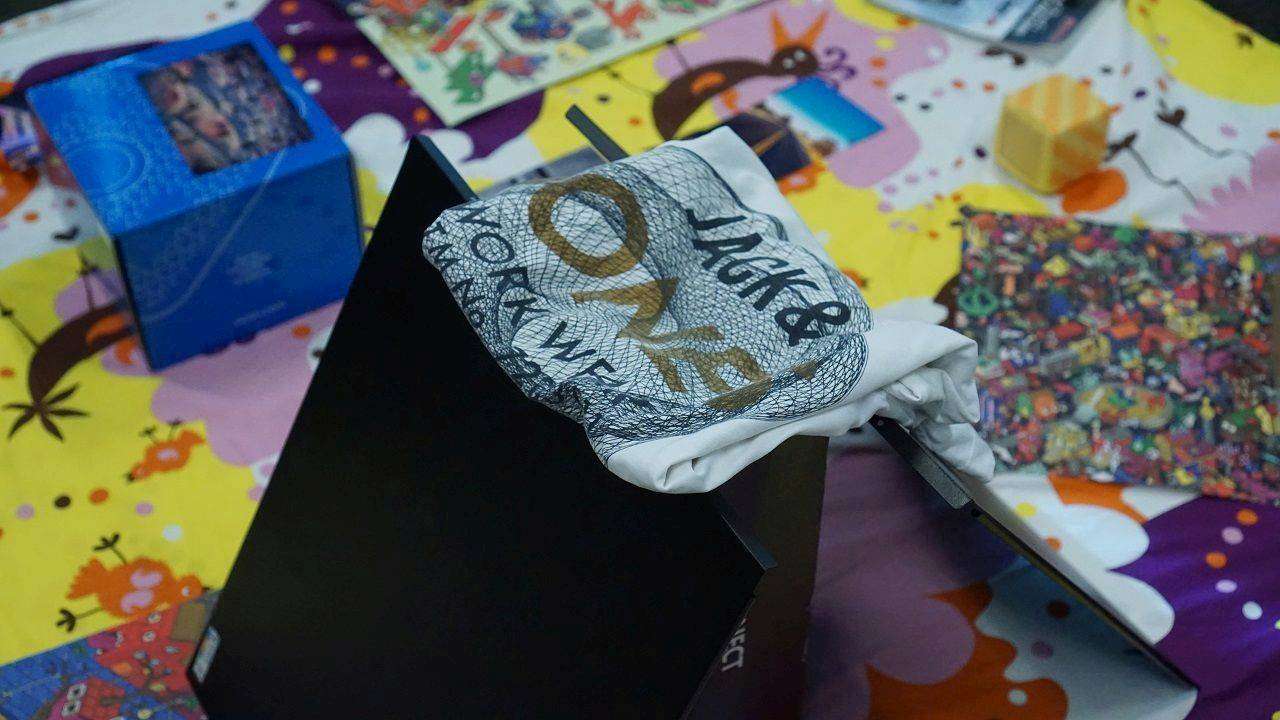}} %
		\\
		\vspaceSame
		\subfloat{\parbox[t]{.02\linewidth}{\begin{sideways}\centering \footnotesize 3D point cloud \end{sideways}}}
		\hfill
		\subfloat{\includegraphics[trim=11cm 5cm 8cm 5cm, clip=true,width=.16\linewidth]{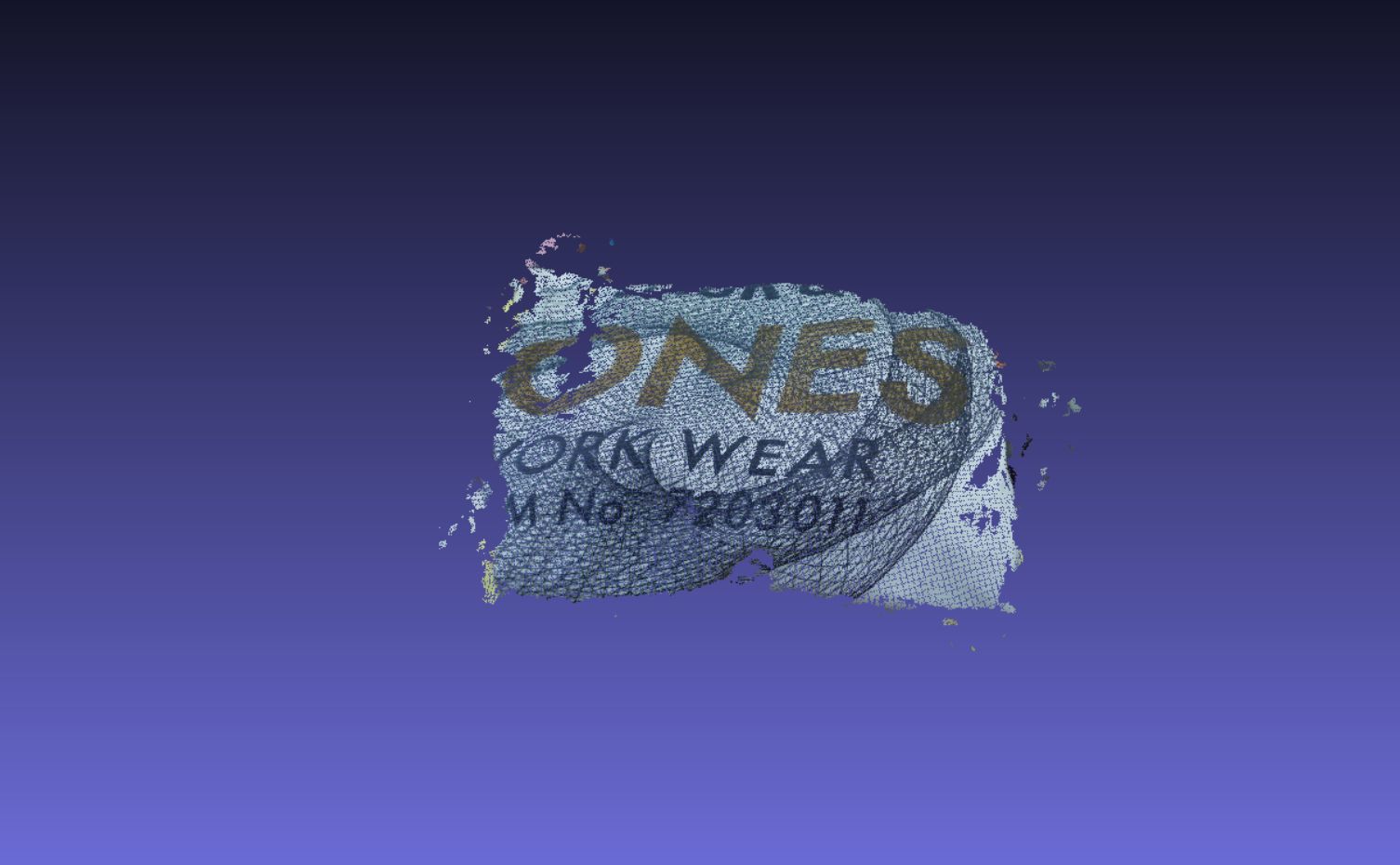}}
		\hfill
		\subfloat{\includegraphics[trim=10cm 5cm 9cm 5cm, clip=true,width=.16\linewidth]{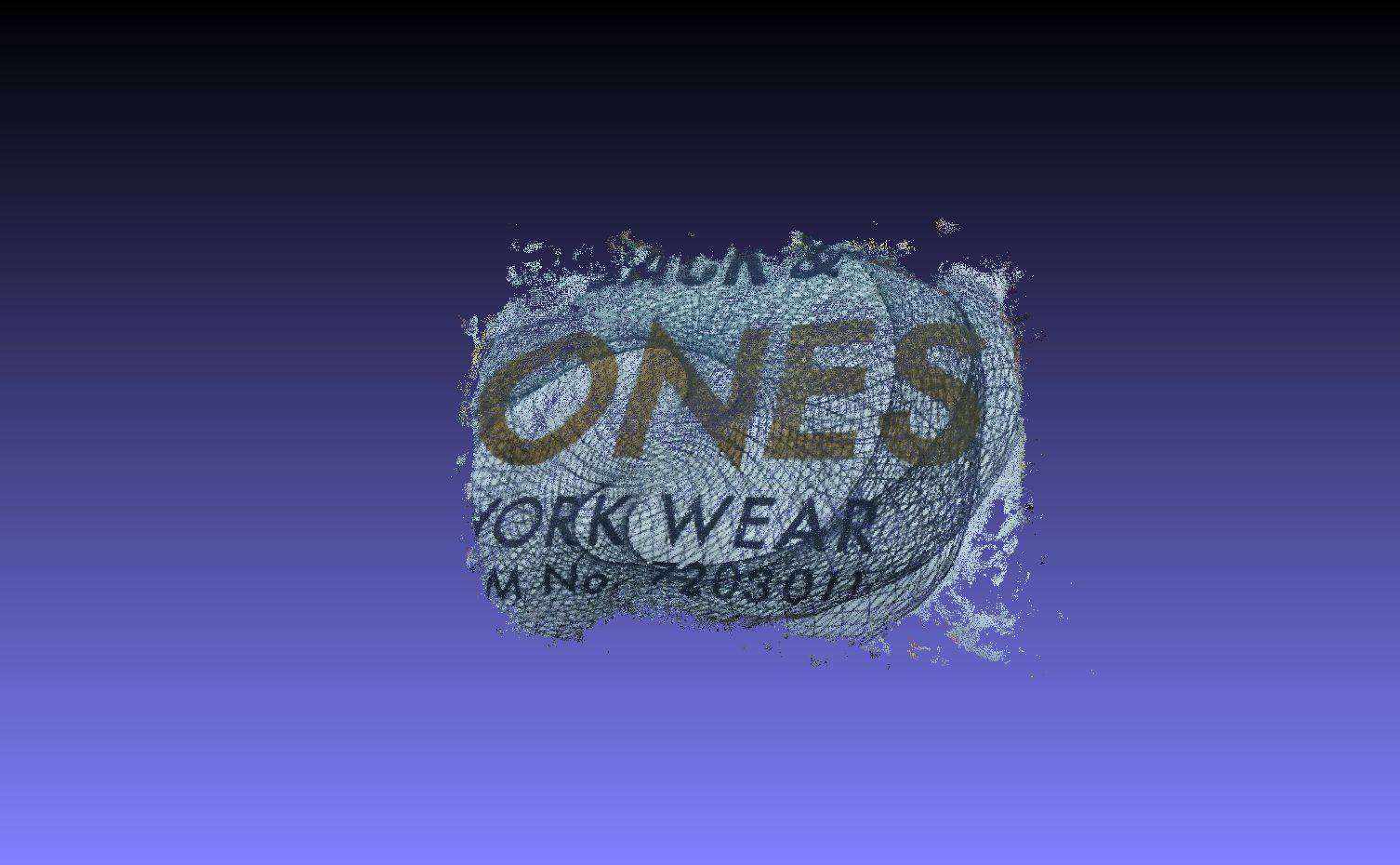}} 
		\hfill
		\subfloat{\includegraphics[trim=10cm 5cm 9cm 5cm, clip=true,width=.16\linewidth]{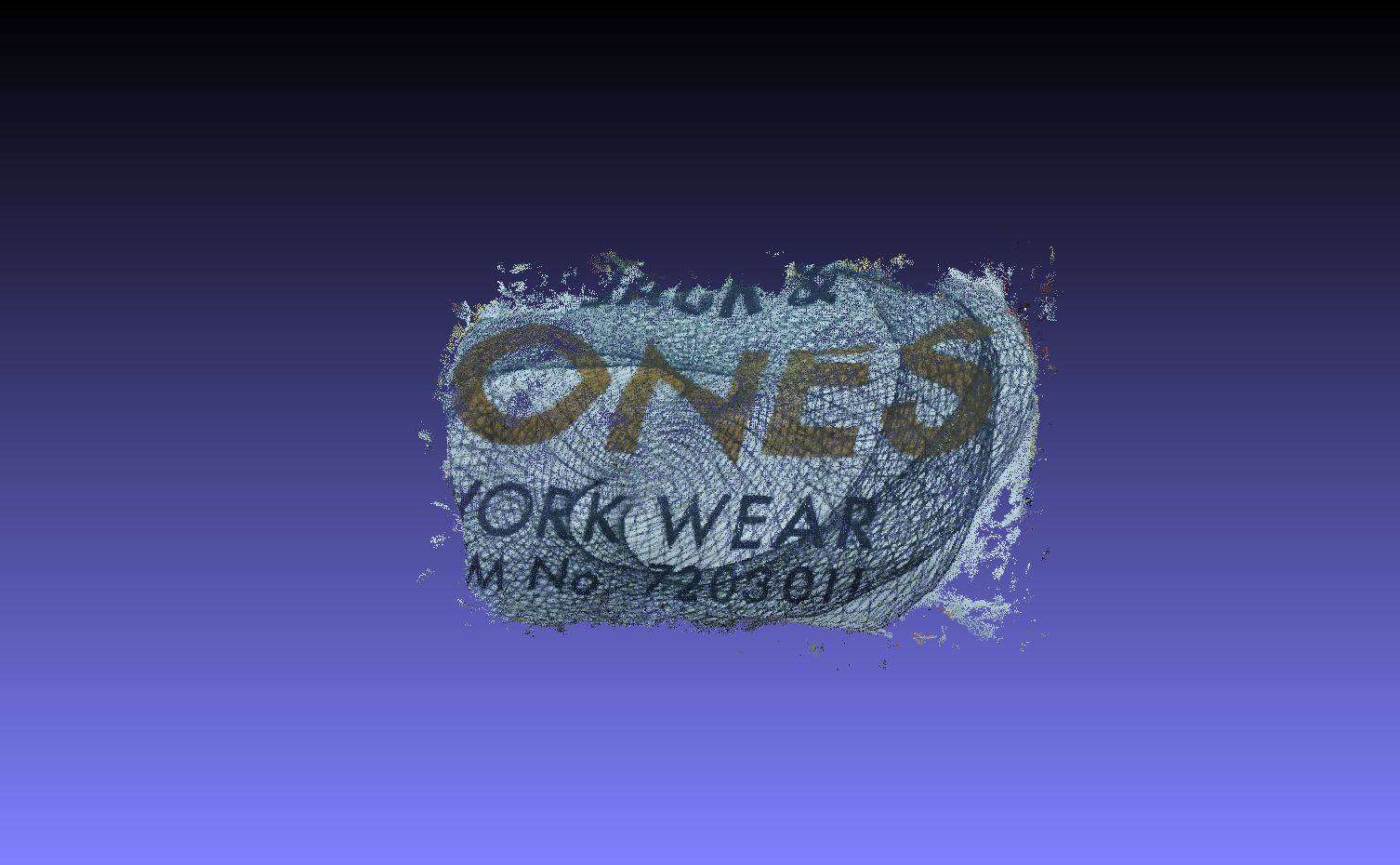}} %
		\hfill
		\subfloat{\includegraphics[trim=11cm 5cm 8cm 5cm, clip=true,width=.16\linewidth]{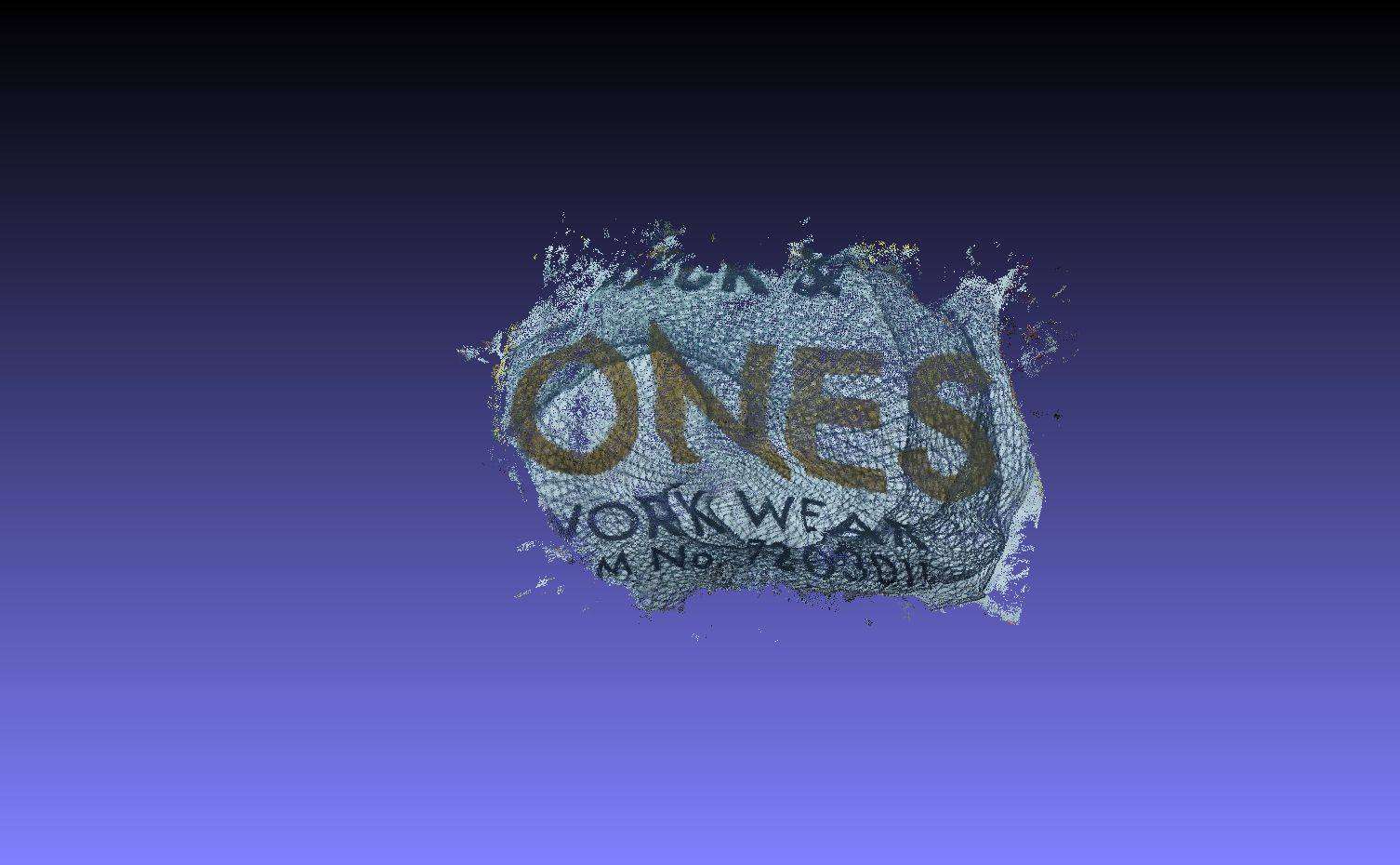}} %
		\hfill
		\subfloat{\includegraphics[trim=10cm 5cm 9cm 5cm, clip=true,width=.16\linewidth]{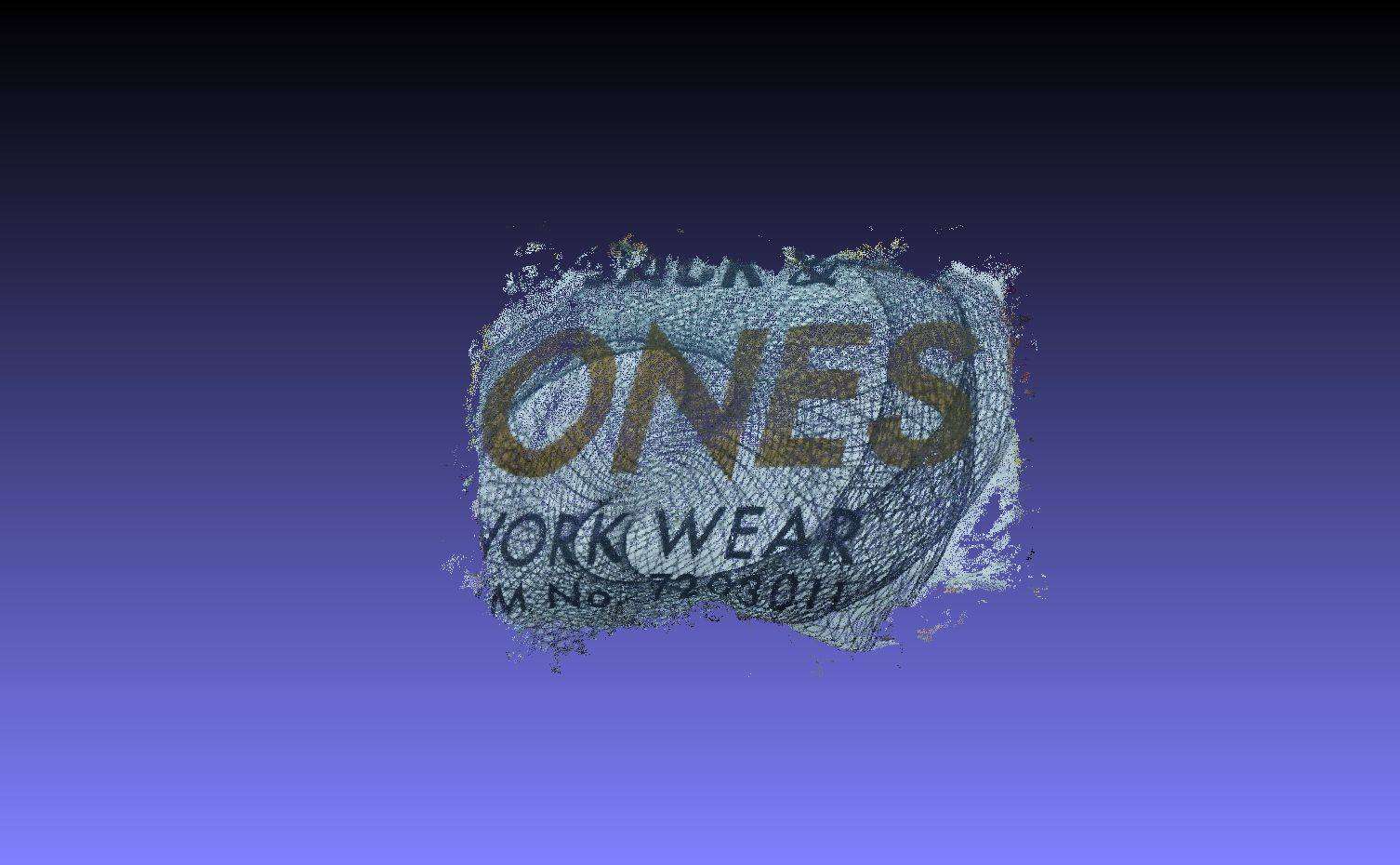}} %
		\hfill
		\subfloat{\includegraphics[trim=10cm 5cm 9cm 5cm, clip=true,width=.16\linewidth]{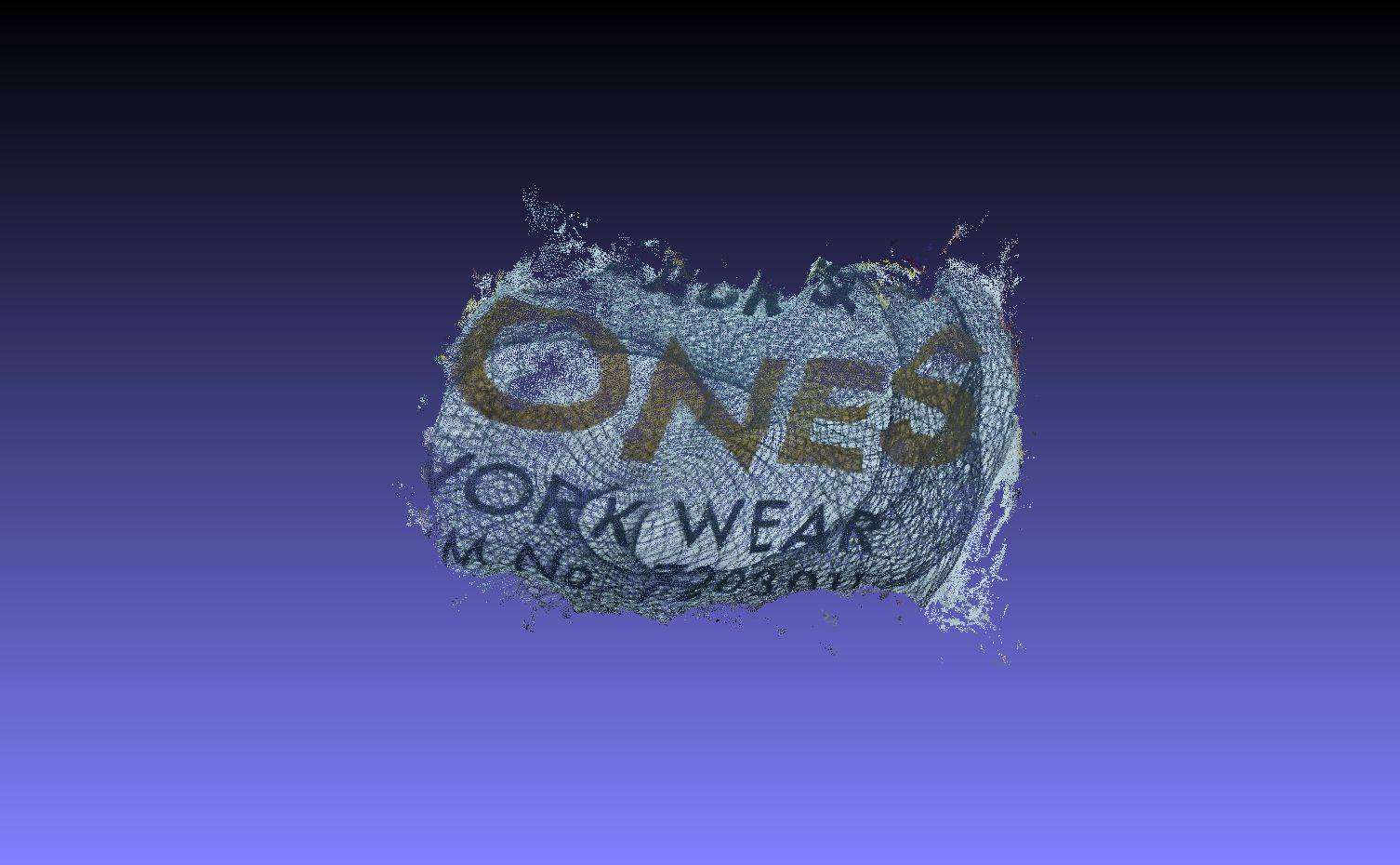}} %
		\\
		\vspaceScene
		\subfloat{\parbox[t]{.02\linewidth}{\begin{sideways}\centering \footnotesize \quad Input images\end{sideways}}}
		\hfill
		\subfloat{\includegraphics[trim=2.5cm 0.5cm 1.5cm 0.5cm, clip=true,width=.16\linewidth]{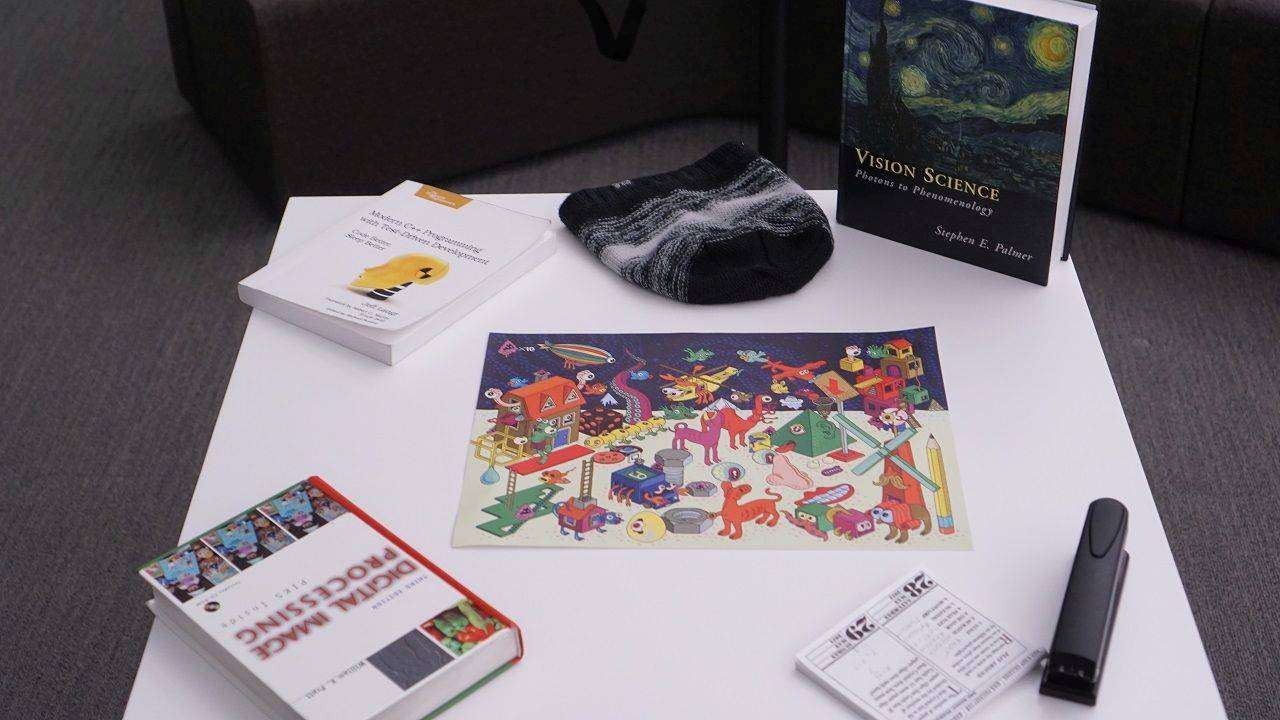}} %
		\hfill
		\subfloat{\includegraphics[trim=2.5cm 0.5cm 1.5cm 0.5cm, clip=true,width=.16\linewidth]{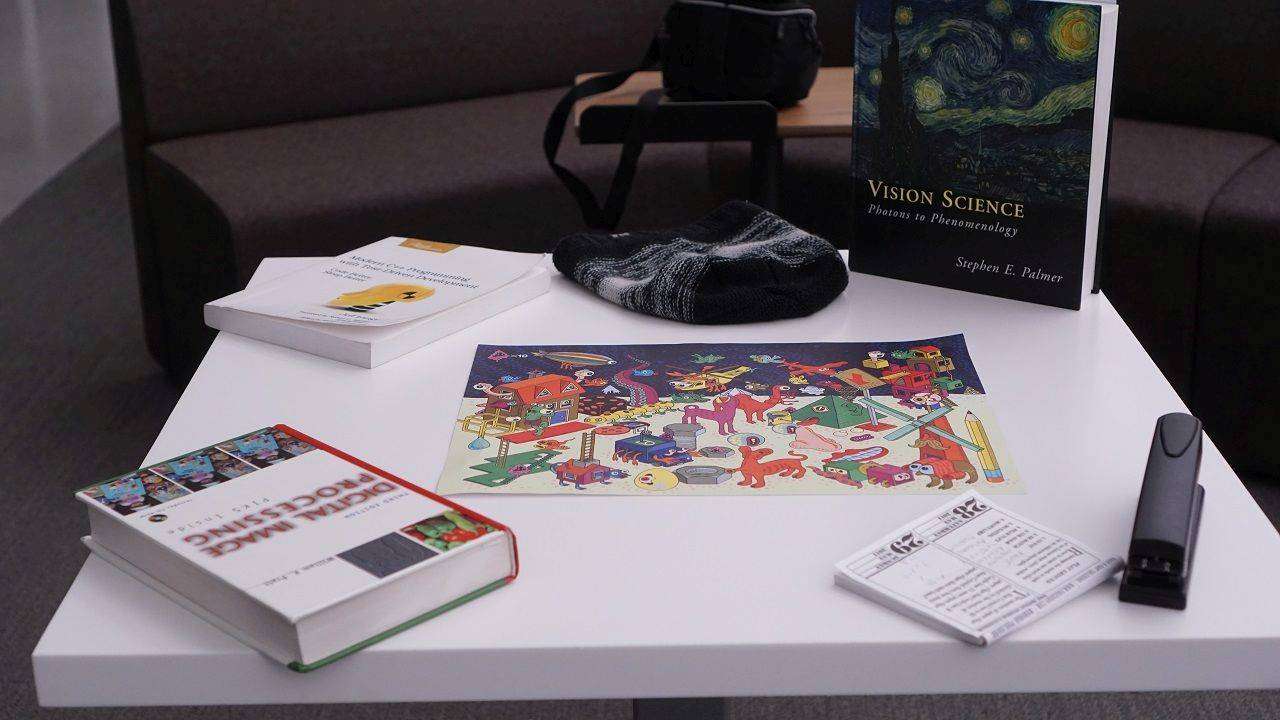}} 
		\hfill
		\subfloat{\includegraphics[trim=2cm 0cm 2cm 1cm, clip=true,width=.16\linewidth]{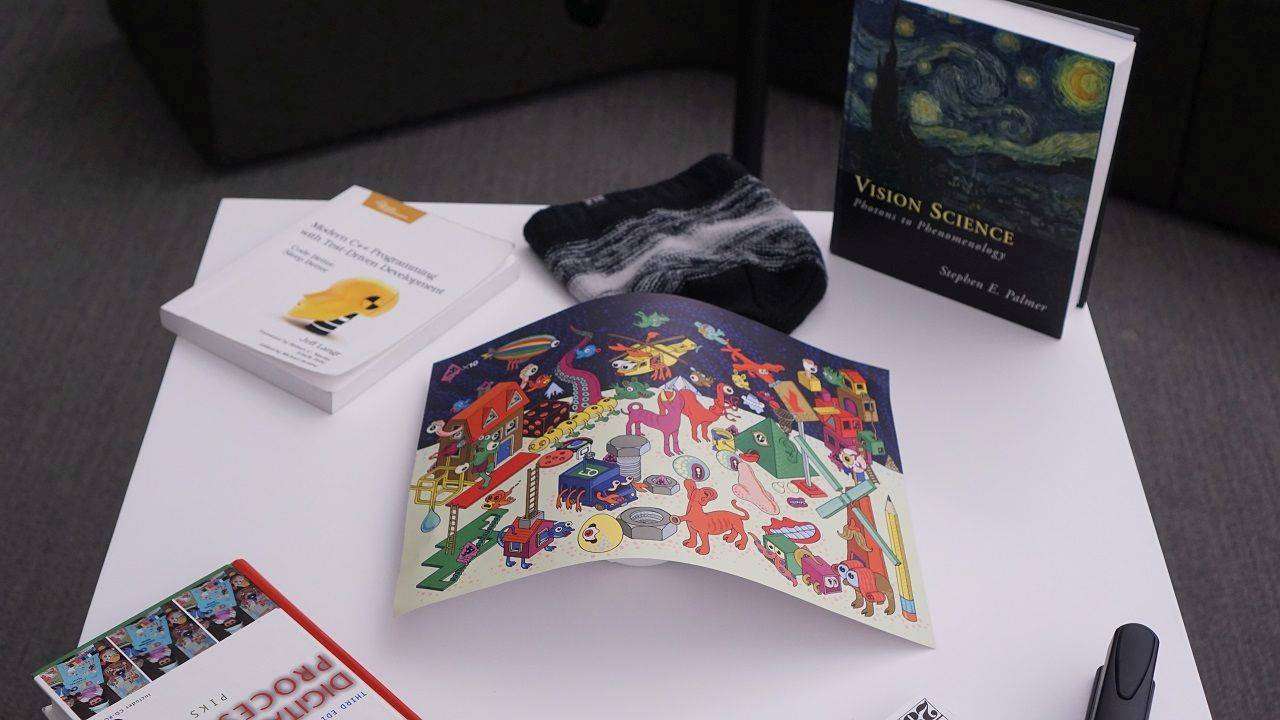}} %
		\hfill
		\subfloat{\includegraphics[trim=2.3cm 0.8cm 1.7cm 0.2cm, clip=true,width=.16\linewidth]{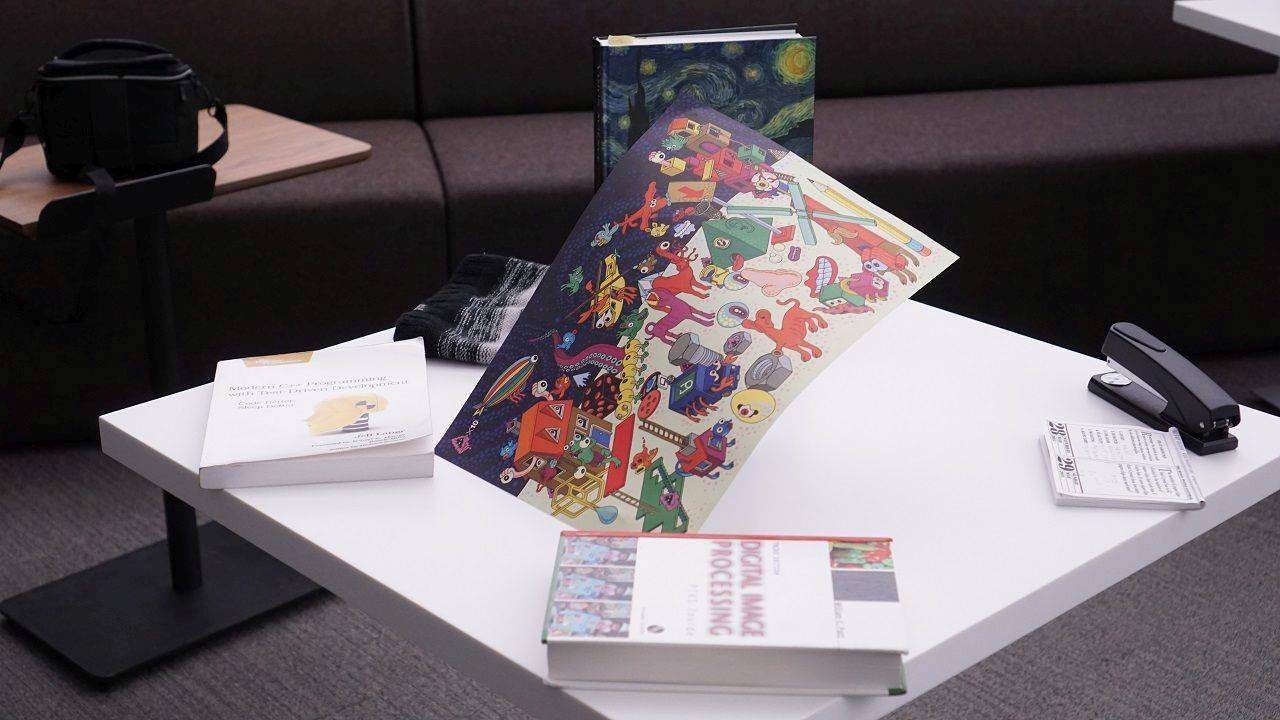}} %
		\hfill
		\subfloat{\includegraphics[trim=2cm 0cm 2cm 1cm, clip=true,width=.16\linewidth]{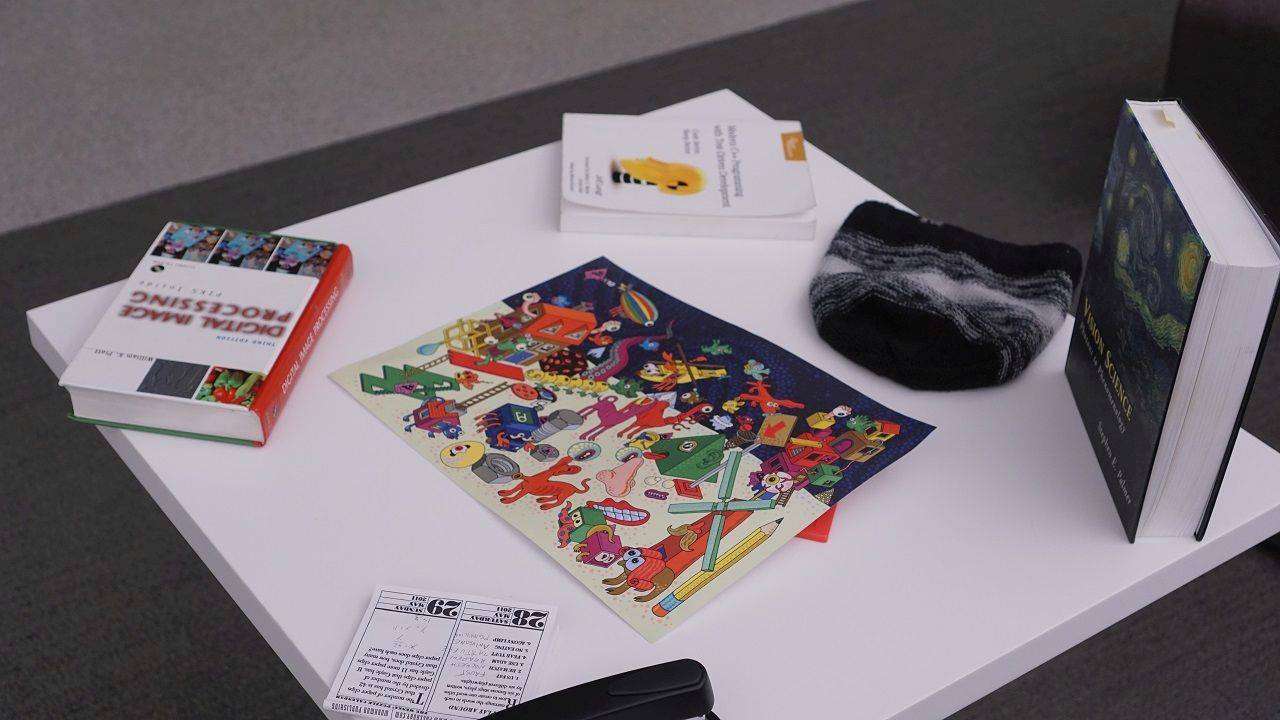}} %
		\hfill
		\subfloat{\includegraphics[trim=2cm 0.8cm 2cm 0.2cm, clip=true,width=.16\linewidth]{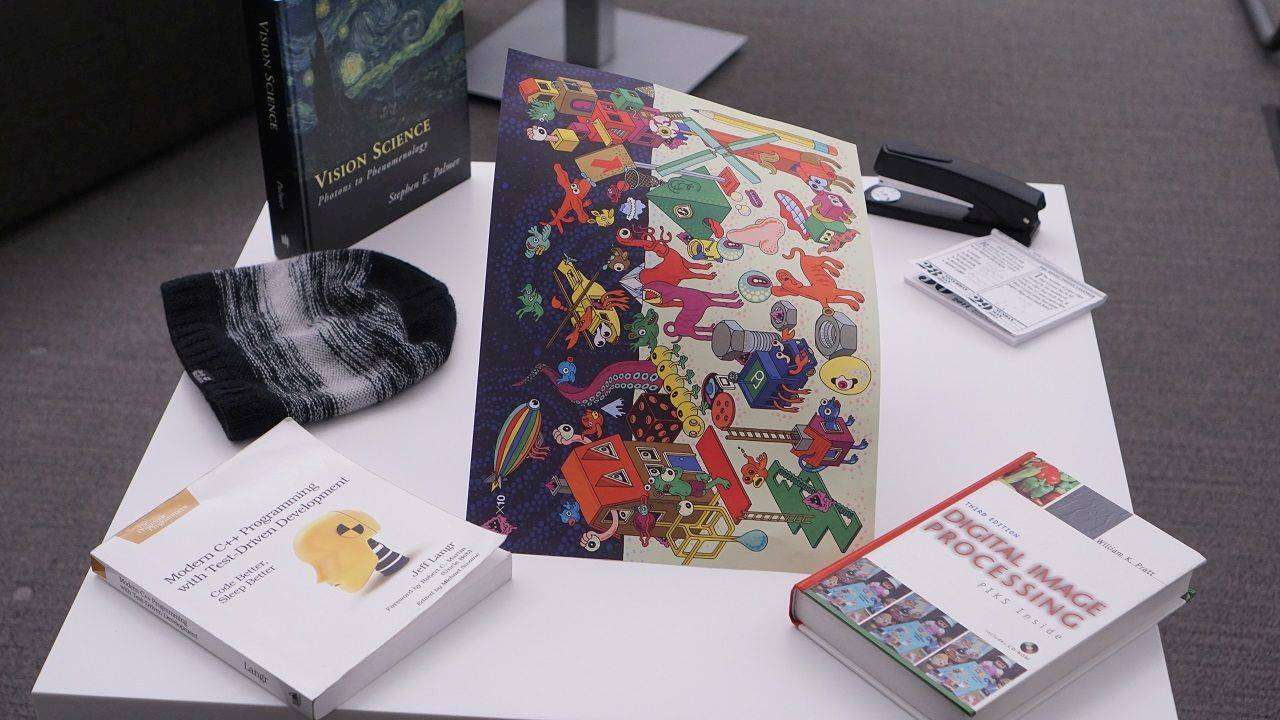}} %
		\\
		\vspaceSame
		\subfloat{\parbox[t]{.02\linewidth}{\begin{sideways}\centering \footnotesize 3D point cloud \end{sideways}}}
		\hfill
		\subfloat{\includegraphics[trim=10cm 5cm 9cm 9cm, clip=true,width=.16\linewidth]{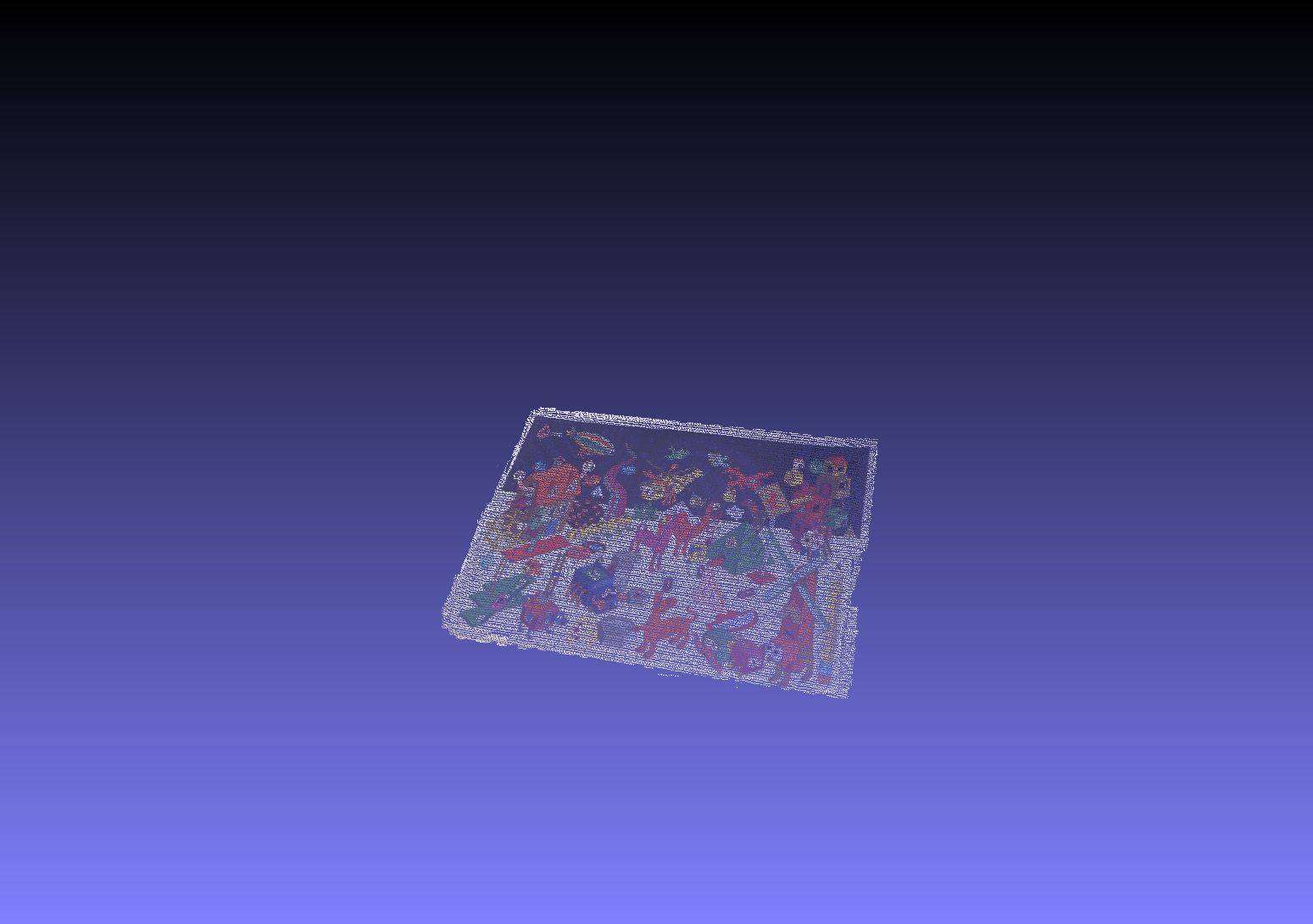}}
		\hfill
		\subfloat{\includegraphics[trim=10cm 5cm 9cm 9cm, clip=true,width=.16\linewidth]{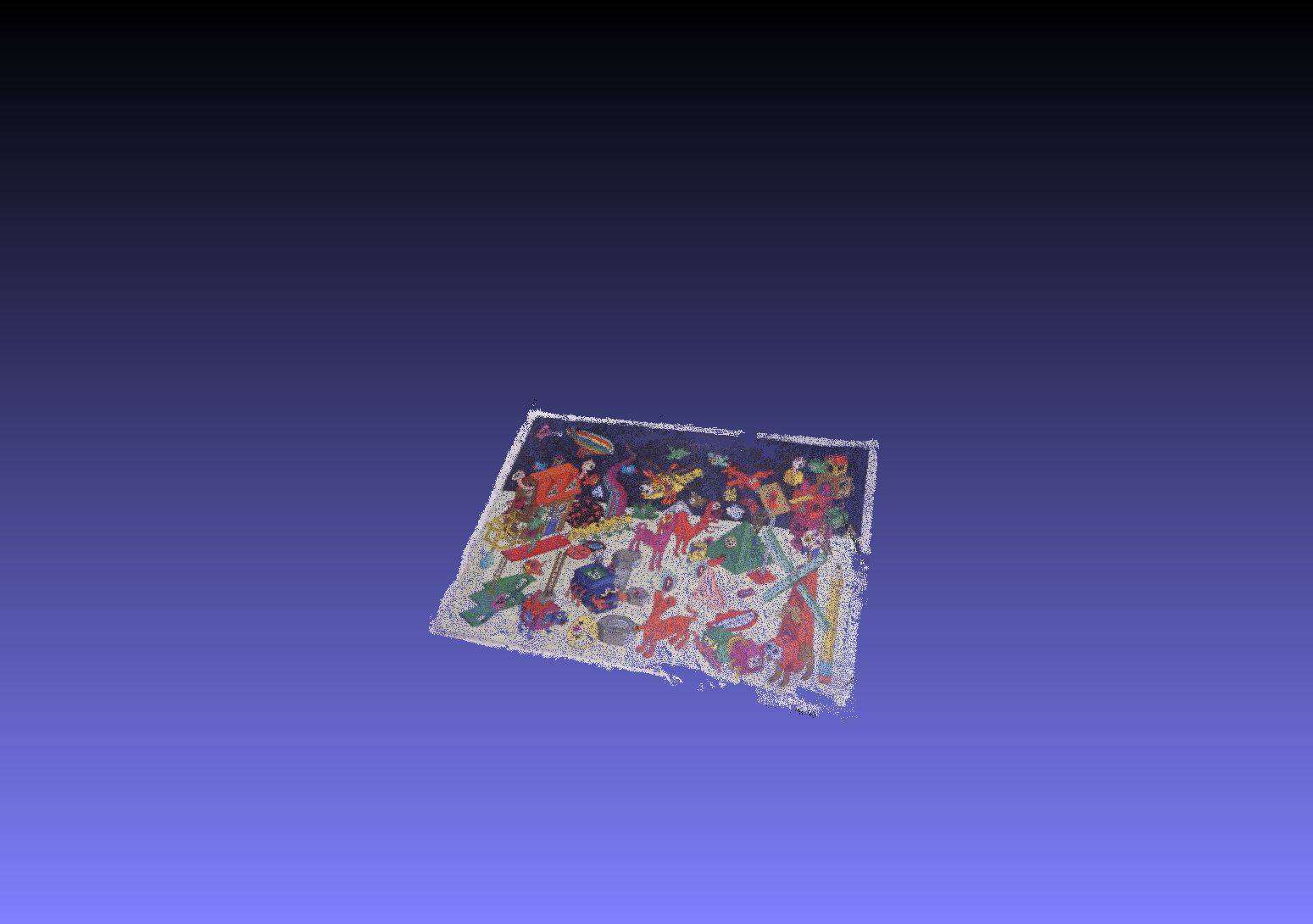}}
		\hfill
		\subfloat{\includegraphics[trim=10cm 8cm 9cm 6cm, clip=true,width=.16\linewidth]{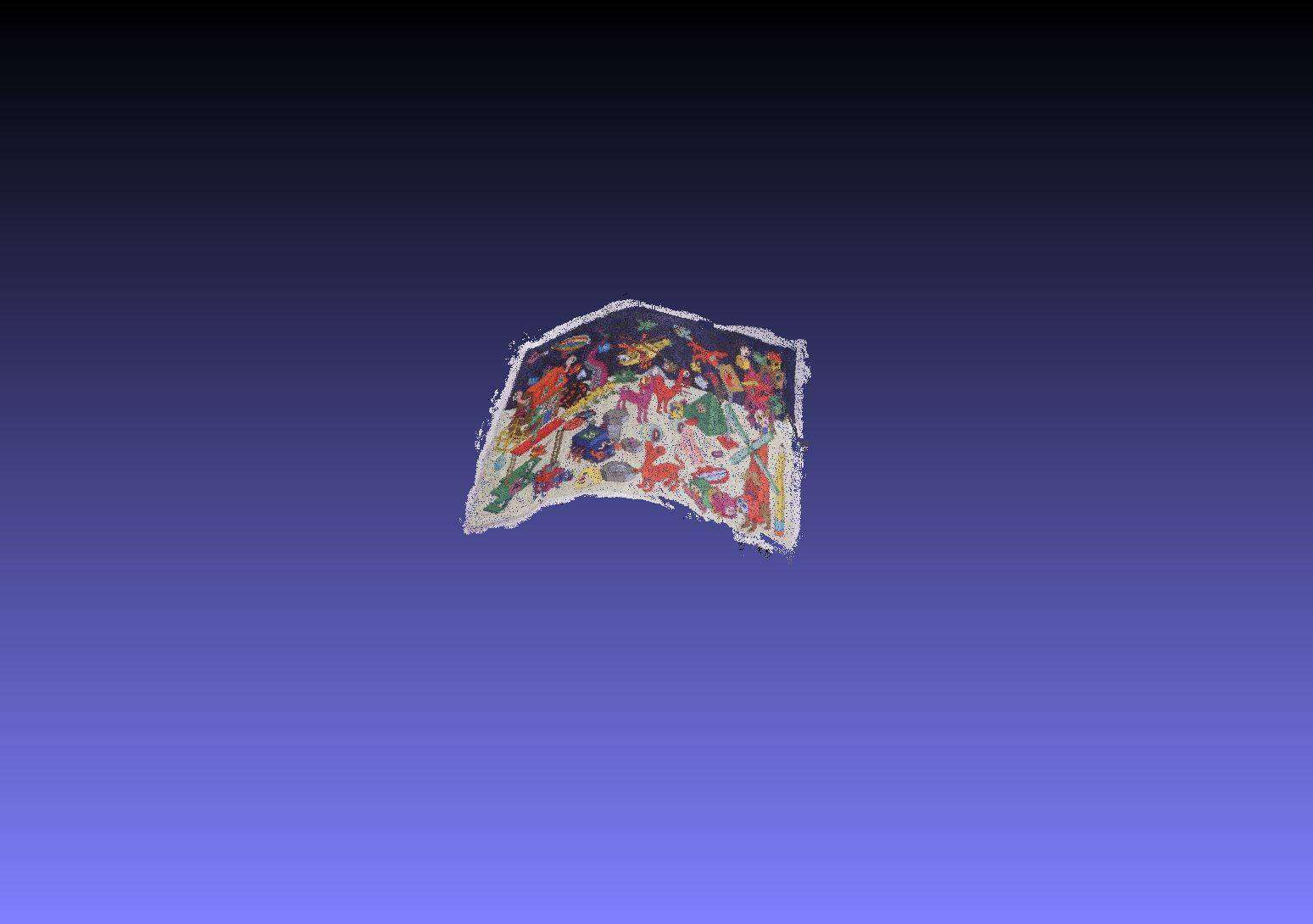}} %
		\hfill
		\subfloat{\includegraphics[trim=10cm 9cm 9cm 5cm, clip=true,width=.16\linewidth]{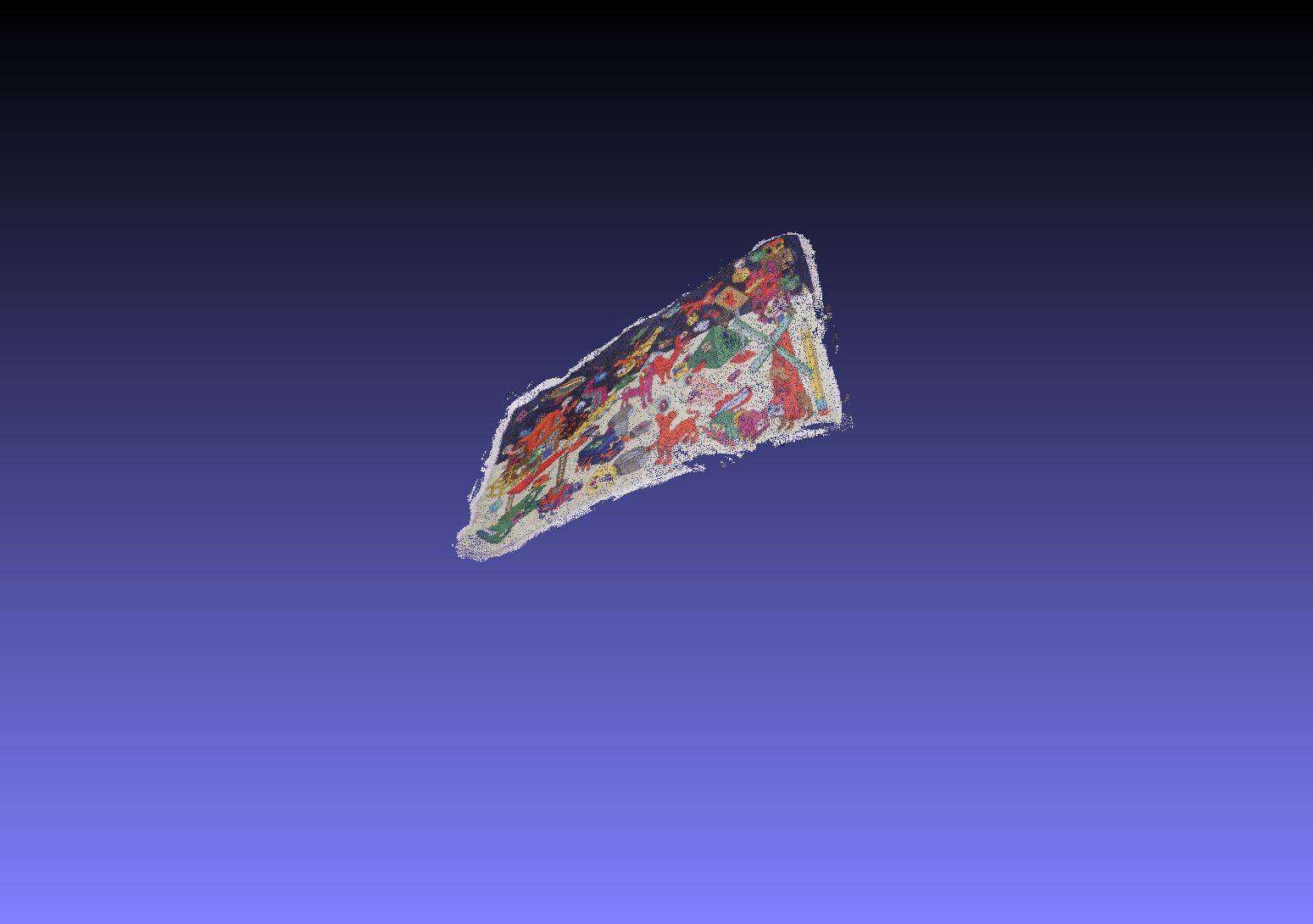}} %
		\hfill
		\subfloat{\includegraphics[trim=10cm 6cm 9cm 8cm, clip=true,width=.16\linewidth]{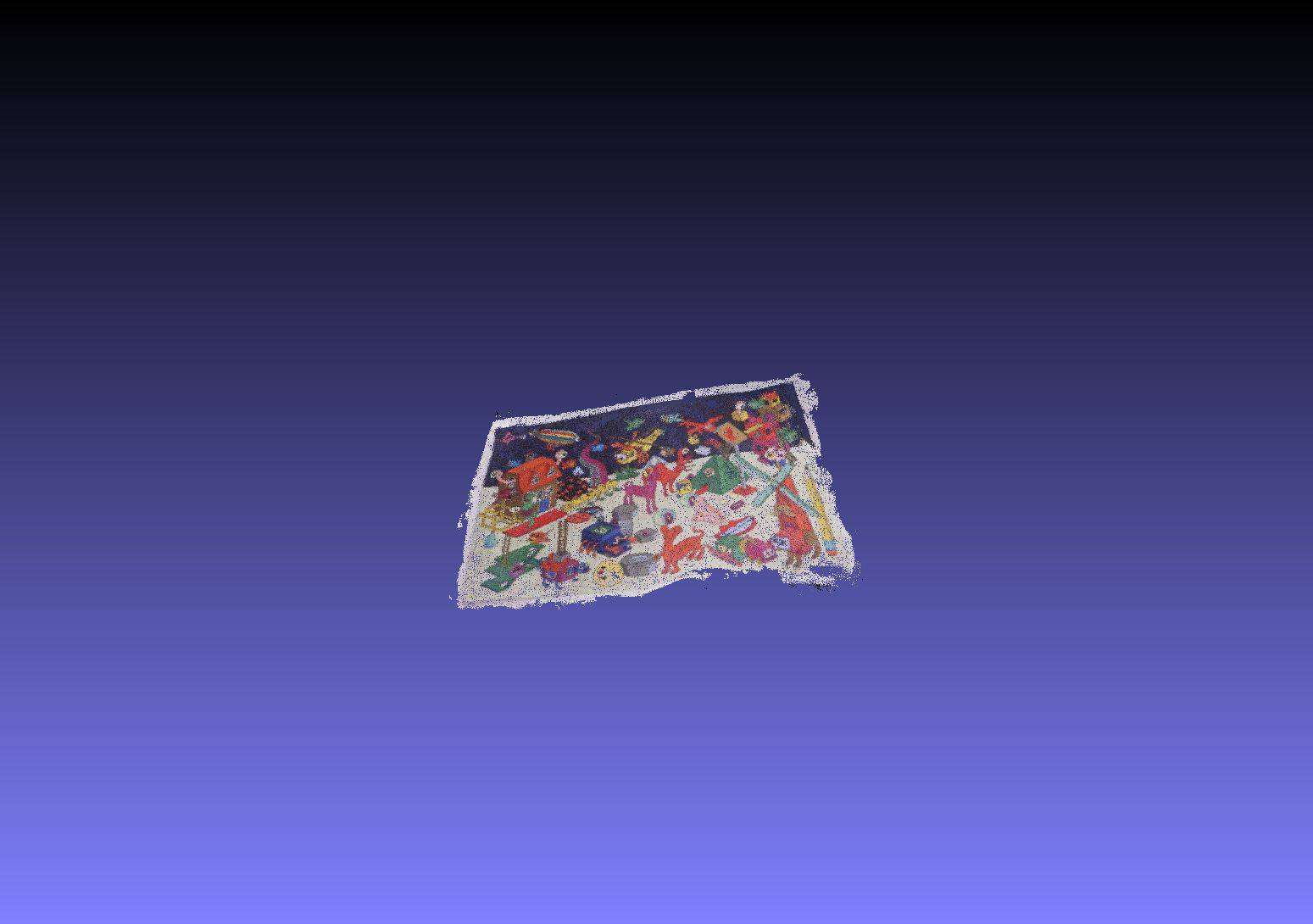}} %
		\hfill
		\subfloat{\includegraphics[trim=10cm 8.5cm 9cm 5.5cm, clip=true,width=.16\linewidth]{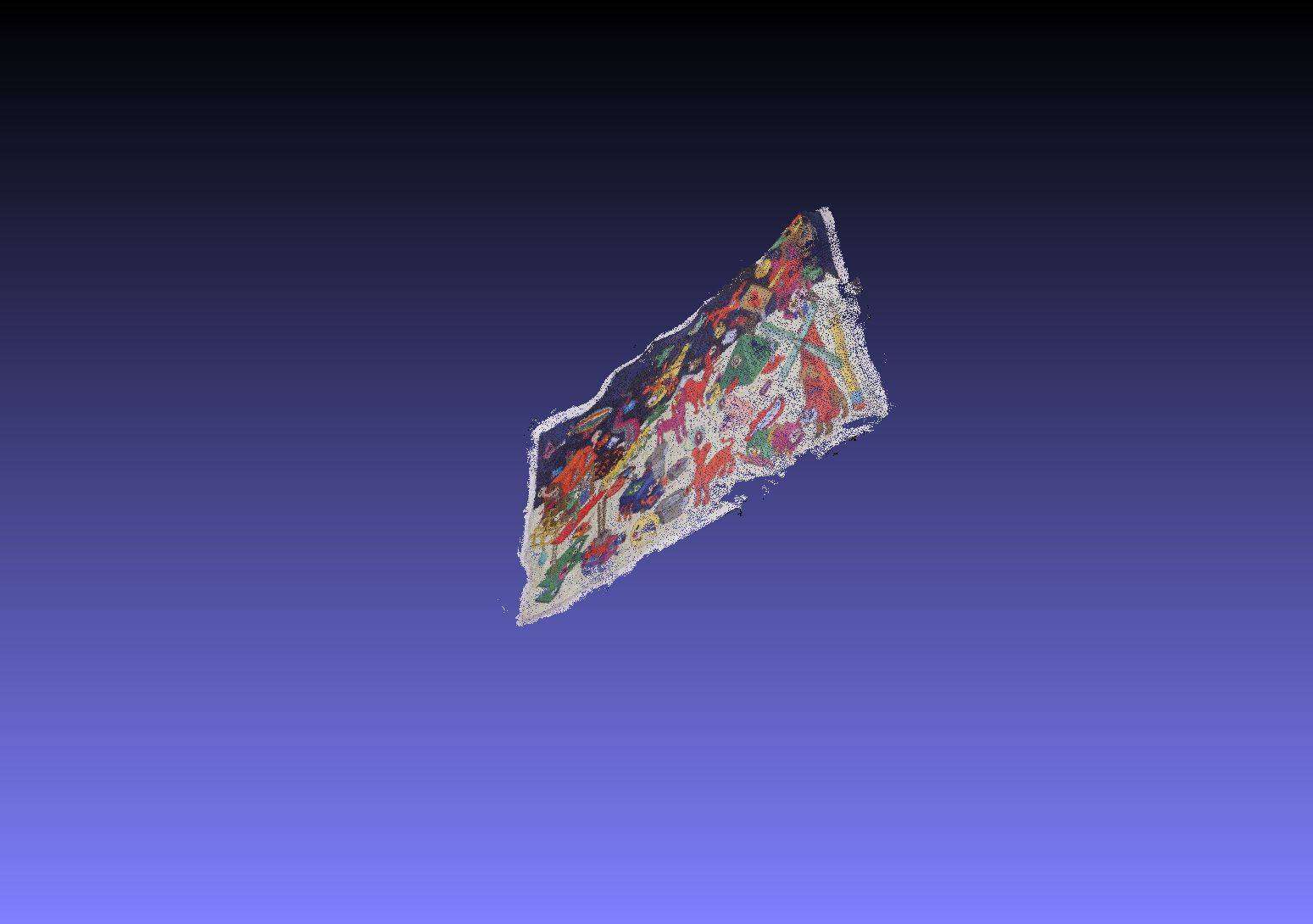}} %
		
		\caption{\textbf{Qualitative evaluation with real data}: In each row, the first two columns show the views used to create each canonical surface. The first column of each result row (even row) shows the original canonical surface. The remaining views from the second column of each result row shows the propagated version of reconstructed surfaces for each view. }
		\label{fig:results}
		\vspace{-.2cm}
	\end{figure*}
	
	\begin{figure}
		\centering
		\includegraphics[clip=true,width=.9\linewidth]{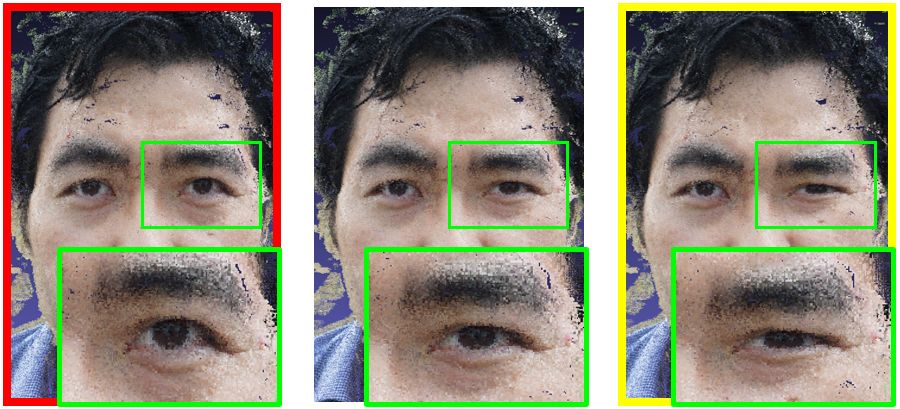}
		\includegraphics[clip=true,width=.9\linewidth]{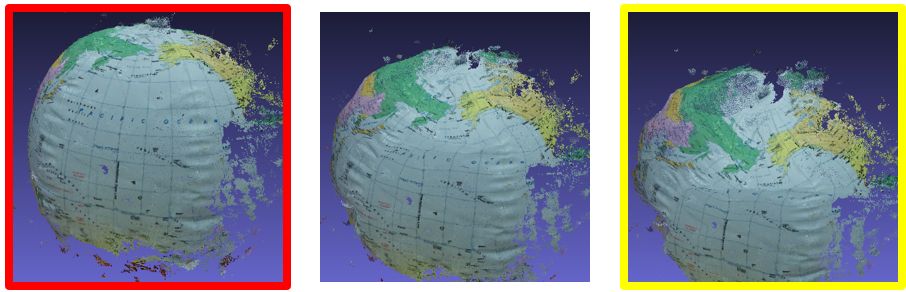}
		\parbox[h]{.29\linewidth}{\centering \scriptsize Source}
		\parbox[h]{.29\linewidth}{\centering \scriptsize Intermediate scene}
		\parbox[h]{.29\linewidth}{\centering \scriptsize Target}
		\caption{\textbf{Dynamic 3D scene interpolation with in-between deformations:} We interpolate a point cloud between two reconstructed views from their depth and deformation. We show two key-frames, source and target, denoted as red and yellow frames respectively, and then demonstrate the interpolated \emph{intermediate point cloud} in the middle column. For the top row, zoomed in-set images of the eye region show how the deformation is applied to the intermediate point cloud. More interpolated frames and 4D animations created from our deformation estimation are shown in the supplementary video with various views\textsuperscript{\ref{ftn:video}}.}
		\label{fig:results-interpolation}
		\vspace{-.2cm}
	\end{figure}
	
	\section{Evaluation}
	For existing non-rigid structure from motion methods, different types of datasets are used to evaluate sparse points~\cite{Jensen18, Dai17}, and dense video frames with small baseline (including actual camera view variation)~\cite{Ansari17}. 
	Since our problem formulation is intended for dense reconstruction of scenes with sufficient variation in both camera view and deformation, there are only few examples applicable to our scenario~\cite{Li13, Wang15, Innmann16eccv}. Unfortunately, these datasets are either commercial and not available~\cite{Li13}, or only exhibit rigid changes~\cite{Wang15}. Few depth-based approaches share the input RGB as well~\cite{Innmann16eccv}, but the quality of the images is not sufficient for our method (i.e., severe motion blur, low resolution (VGA) that does not provide sufficient detail for capturing non-rigid changes). 
	Thus, we created both synthetic data and captured real-world examples for the evaluation.
	To quantitatively evaluate how our method can accurately capture a plausible deformation and reconstruct each scene undergoing non-rigid changes, we rendered several synthetic scenes with non-rigid deformations as shown in the first row of Fig.~\ref{fig:results-synth}. 
	We also captured several real-world scenes containing deforming surfaces from different views at different times. Some examples (face, rubber globe, cloth and paper) appear in Fig.~\ref{fig:results}, and several more viewpoints are contained in the supplementary video\footnote{\url{https://youtu.be/et_DFEWeZ-4}\label{ftn:video}}.
	
	\subsection{Quantitative Evaluation with Synthetic Data}
	First, we evaluate the actual depth errors of the reconstructed depth of each time frame (i.e., propagated/refined to a specific frame), and of the final refined depth of the \emph{canonical view}. 
	Because we propose the challenging new problem of reconstructing non-rigid dynamic scenes from a small set of images, it is not easy to find other baseline methods. 
	Thus, we conduct the evaluation with an existing MVS method, COLMAP~\cite{schoenberger2016mvs}, as a lower bound, and use as an upper bound a non-rigid ICP method similar to Li~et~al.~\cite{Li09} based on the ground truth depth. 
	The non-rigid ICP using the point-to-plane error metric serves as a geometric initialization.
	We refine the deformation using our dense photometric alignment (see Sec.~\ref{sec:photometric-conssitency}).
	
	To compare the influence of our proposed objectives for deformation estimation, i.e. sparse 3D-3D correspondences and dense non-rigid photometric consistency, we evaluate our algorithm with different settings. 
	The relative performance of these variants can be viewed as an ablation study. 
	We perform evaluation on the following variants: 1) considering only the sparse correspondence association using different numbers of iterations (see Sec.~\ref{sec:photometric-conssitency}), 2) considering only the dense photometric alignment, and 3) the combination of sparse and dense.
	The results of the quantitative evaluation can be found in Table~\ref{tab:eval}.
	As can be seen, all methods/variants obtain a mean relative error $< 2.4 \%$, overall resulting in faithfully reconstructed geometry.
	Our joint optimization algorithm considerably improves the reconstruction result both in terms of accuracy (by a factor of $1.9$) and completeness (by $30$ pp, a factor of $1.4$).
	Additionally, we compute the mean relative depth error (MRE) without rejecting outliers; i.e., resulting in depth images with a completeness of 100\%.
	
	\subsection{Qualitative Evaluation with Real Data}
	Fig.~\ref{fig:results} shows results of our non-rigid 3D reconstruction. 
	For each pair of rows, we show six input images and the corresponding deformed 3D point clouds. 
	Note that the \emph{deformed} surfaces belong to the collection of 3D reconstructed points propagated by the computed deformations using the other views as described in Sec.~\ref{sec:photometric-conssitency}. 
	The point cloud of each first column of Fig.~\ref{fig:results} shows the first canonical surface (triangulated points from two views with minimal deformation).
	For evaluation purposes, we visualize each reconstructed scene from a similar viewpoint as one of the canonical views. 
	More viewpoints of the reconstructed 3D results can be found in the supplementary video\textsuperscript{\ref{ftn:video}}.
	
	\subsection{Dynamic Scene Interpolation} 
	Since we estimate deformations between each view and the canonical surface, once all deformation pairs have been created, we can easily interpolate the non-rigid structure.
	To blend between the deformations, we compute interpolated deformation graphs by blending the rigid body transform at each node using dual-quaternions \cite{kavan2007skinning}.
	
	In Fig.~\ref{fig:results-interpolation}, we show interpolated results from reconstructed scenes of the face example and the globe example shown in Fig.~\ref{fig:results}. 
	The scene deformations used for this interpolation (like key-frames) are framed in as red and yellow. 
	Note that even though the estimated deformation is defined between each view and the canonical pose, any combination of deformation interpolation is possible. 
	More examples and interpolated structures from various viewpoints can be found in the supplementary video\textsuperscript{\ref{ftn:video}}.
	
	\begin{figure}[t]
		\centering
		\def\size{2.1cm}
		\vspace{-0.3cm}
		\subfloat[Bad canonical views selection]{\includegraphics[height=\size]{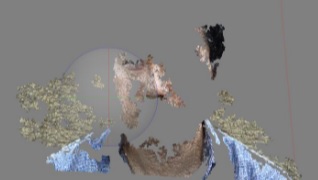}}
		\hfill
		\subfloat[Ambiguity along view direction]{\includegraphics[height=\size]{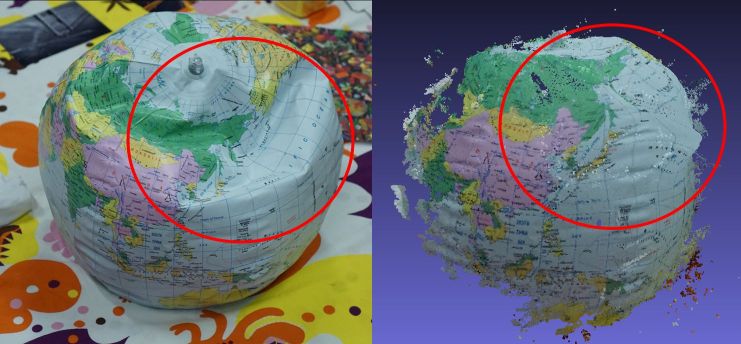}}
		\caption{\textbf{Failure cases:} (a) shows the result of canonical surface reconstruction from two views that are incorrectly selected (large deformation between two views: images in first and third column in top row of Fig.~\ref{fig:results}). While the camera pose is successfully computed, since there are large portions of non-rigid changes happening in the upper part of face and near the mouth, there are many holes on the face, which is not the best case if we choose this pair.  (b) shows a failure case when deformation (red circles) occurs along the view direction, which causes the ambiguity.}
		\label{fig:failure}
		\vspace{-.2cm}
	\end{figure}

	\section{Conclusion and Discussion}
	We propose a challenging new research problem for dense 3D reconstruction of scenes containing deforming surfaces from sparse, wide-baseline RGB images. 
	As a solution, we present a joint optimization technique that optimizes over depth, appearance, and the deformation field in order to model these non-rigid scene changes. 
	We show that an MVS solution for non-rigid change is possible, and that the estimated deformation field can be used to interpolate motion in-between views.
	
	It is also important to point out the limitations of our approach (Fig.~\ref{fig:failure}). We first assume that there is at least one pair of images that has minimal deformation for the initial canonical model. 
	This can be interpreted as the first step used by many SLAM or 3D reconstruction algorithms for the initial triangulation. 
	Fig.~\ref{fig:failure}(a) shows an example of a canonical surface created from two views that contain too much deformation, only leading to a partial triangulation.  
	Fig.~\ref{fig:failure}(b) shows an example where the deformation occurs mostly along the view direction. 
	While we successfully estimate the deformation and reconstruct a similar example shown in Fig.~\ref{fig:results}, depending on the view this can cause an erroneous estimation of the deformation.
	On the other hand, we believe that recent advances in deep learning-based approaches to estimate depth from single RGB input~\cite{Fu2018DeepOR} or learning local rigidity~\cite{Lv18eccv} for rigid/non-rigid classification can play a key role for both the initialization and further mitigation of these ambiguities. 
	
	{\small
		\bibliographystyle{ieee}
		\bibliography{nrmvs}
	}

	\vfill
	\newpage

	\begin{appendices}
		We give an overview of the mathematical symbols used in our algorithm (Sec.~\ref{sec:symbols}). In Sec~\ref{sec:param}, we enumerate the parameters that we used for the optimization and Patchmatch steps shown in Sec~\ref{sec:photometric-conssitency}.
		In Sec.~\ref{sec:approach}, we show additional implementation details as well as extra experiments for our pre-processing steps (Sec.~\ref{sec:pose}, Sec.~\ref{sec:canonical}), and show how our optimization effects the photometric consistency. 
		In Sec.~\ref{sec:performance}, we discuss the runtime of our approach.		
		Finally, in Sec.~\ref{sec:results}, we provide extra experimental results in both qualitative and quantitative manners.
		
		We demonstrate various 4D animation examples created from few images by using our NRMVS framework in the attached video\footnote{\url{https://youtu.be/et_DFEWeZ-4}}.
		
		\section{List of Mathematical Symbols}
		\label{sec:symbols}
		\begin{table}[h]
			\centering
			\begin{tabular}{|c|c|}
				\hline
				\textbf{Symbol} & \textbf{Description}  \\
				\hline
				\hline
				$\mathbf{D}_i$ & deformation from canonical pose to image $i$ \\
				\hline
				$\mathbf{v}, \mathbf{x}$ & point in $\mathbb{R}^3$ \\
				\hline
				$\mathbf{n}$ & normal vector in $\mathbb{R}^3$ \\
				\hline
				$\mathbf{u}_i$ & SIFT keypoint in image $i$ \\
				\hline
				$C_i$ & consistency mask $\in \{ 0, 1 \}$ for point $\mathbf{x}_i$ \\
				\hline
				$\rho_{r,s}$ & weighted NCC value for images $r$ and $s$ \cite{schoenberger2016mvs} \\
				\hline
				$\mathbf{I}_i$ & greyscale image $i$ \\
				\hline
				$d_i$ & depthmap for image $i$ \\
				\hline
			\end{tabular}
			\label{tab:symbols}
		\end{table}
		
		\section{Parameter choices}
		\label{sec:param}
		\begin{table}[h]
			\centering
			\begin{tabular}{|c|c|}
				\hline
				\textbf{Parameter} & \textbf{Value}  \\
				\hline 
				\hline
				$w_\text{sparse}$ & $1000$ \\
				\hline
				$w_\text{dense}$ & $0.01$ \\
				\hline
				$w_\text{reg}$ & $10$ \\
				\hline
				$d_\text{max}$ & $0.1$ cm $\ldots 0.5$ cm \\
				\hline
				$\rho_\text{max}$ & $0.9$ \\
				\hline
				$\tau$ & $0.9$ \\
				\hline
			\end{tabular}
			\label{tab:params}
		\end{table}
		
		$d_\text{max}$ is chosen depending on the scale of the scene geometry; e.g., we choose $0.1$~cm for the face example and $0.5$~cm for the globe example.
		In case of the synthetic ground truth data, we use $d_\text{max} = 0.01$, with the rendered plane having a size of $6$.
		
		The parameter \texttt{filter\_min\_num\_consistent} in the implementation of COLMAP's PatchMatch~\cite{schoenberger2016mvs} as well as our non-rigid PatchMatch is set to $1$ (default $2$).
		Besides that, we use COLMAP's default parameters throughout our pipeline.
		
		\section{Approach}
		\label{sec:approach}
		\subsection{Deformation Estimation}
\begin{figure}[h]
			\centering
			\subfloat[Initial state]{\includegraphics[trim=9cm 0cm 9cm 3cm, clip=true, width=0.41\columnwidth]{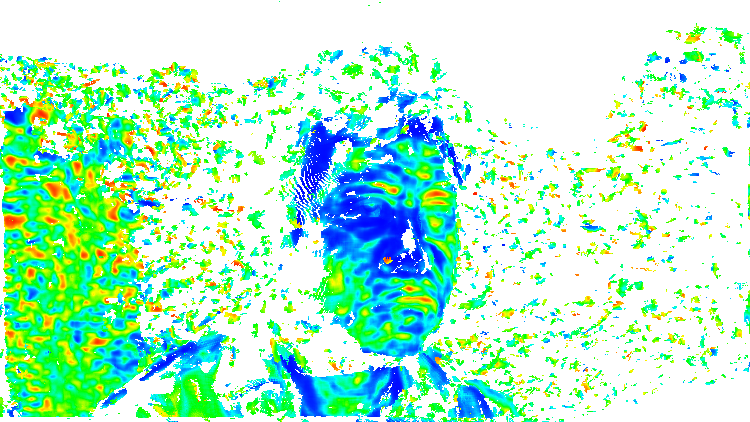}} 
			\hfill
			\subfloat[After optimization]{\includegraphics[trim=9cm 0cm 9cm 3cm, clip=true, width=0.41\columnwidth]{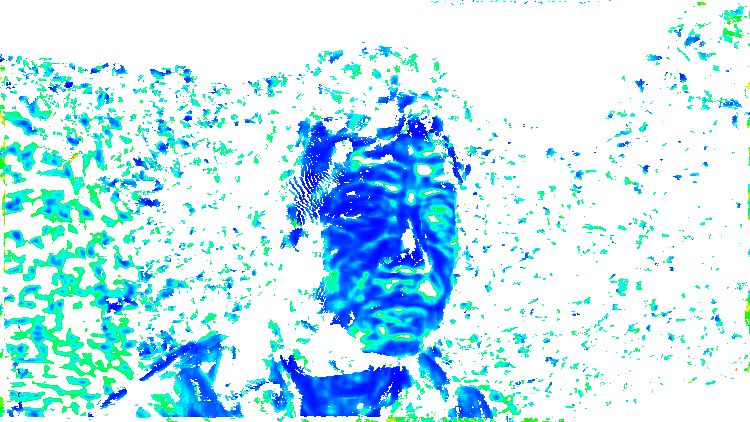}}
			\hfill
			\subfloat{\includegraphics[width=0.15\columnwidth]{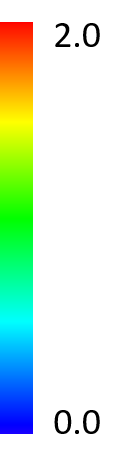}}
			\caption{Photoconsistency cost including consistency mask $C_i \cdot (1 - \text{NCC})$ between Fig.~\ref{fig:kihwan}(a) and (c). Masked out pixels are transparent in (b).}
			\label{fig:ncc_cost}
		\end{figure}

		In Fig.~\ref{fig:ncc_cost}, we show an example of the photometric consistency before and after the estimation of the non-rigid deformation. 
		As shown, the photometric error gets reduced and some inconsistent regions (not satisfying a user-specified threshold $\rho_\text{max}$) get masked out by the consistency map $C_i$.
		
		\subsection{Performance}
		\label{sec:performance}
\begin{figure}[h]
			\centering
			\subfloat[]{\includegraphics[width=.32\columnwidth]{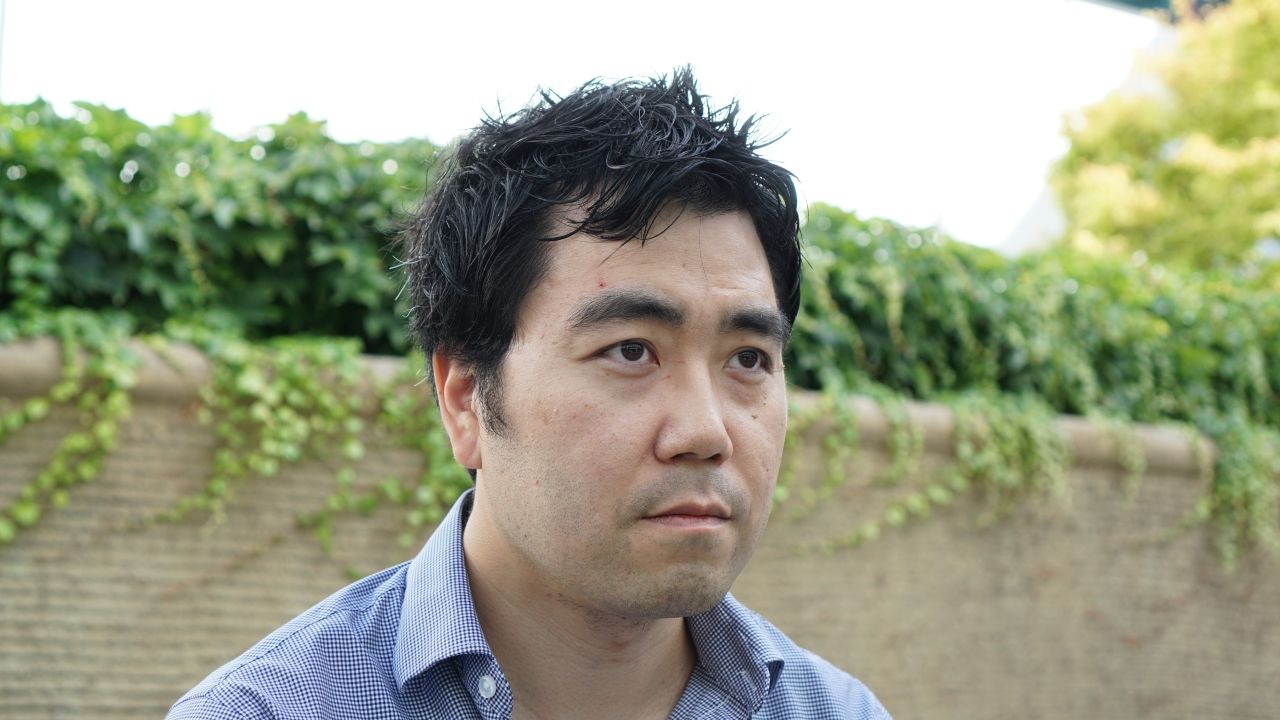}}
			\hfill
			\subfloat[]{\includegraphics[width=.32\columnwidth]{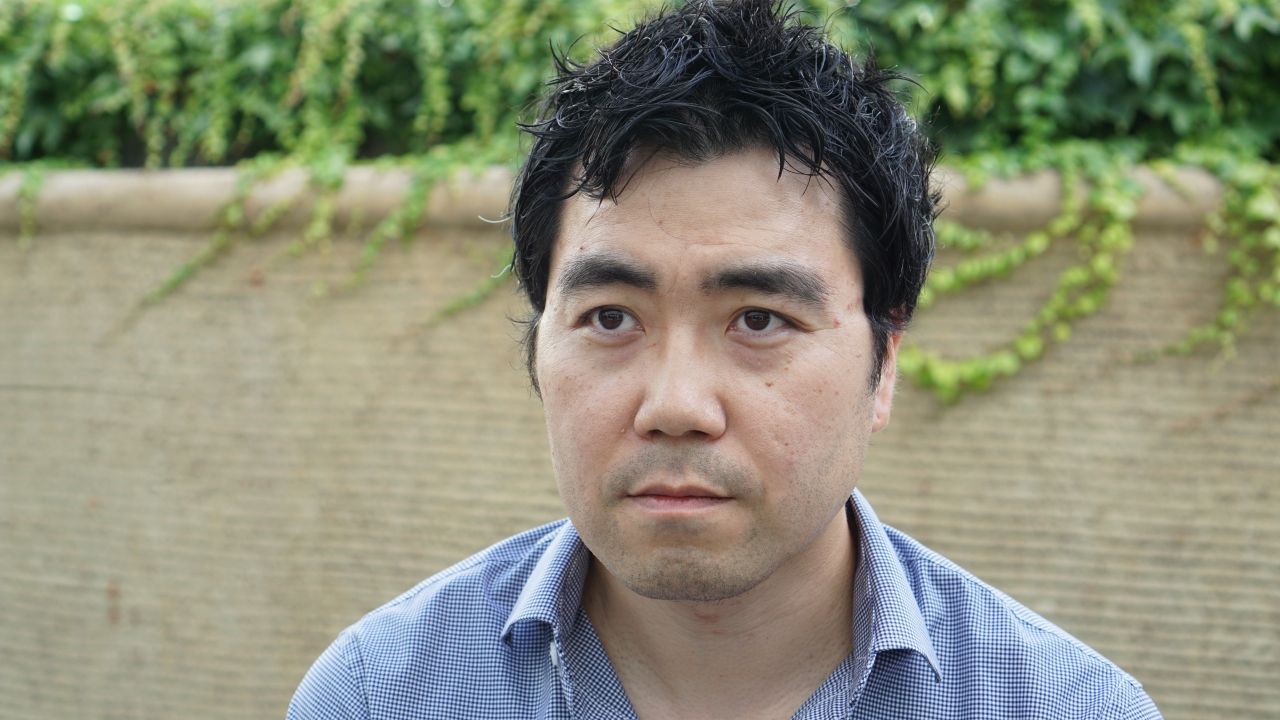}}
			\hfill
			\subfloat[]{\includegraphics[width=.32\columnwidth]{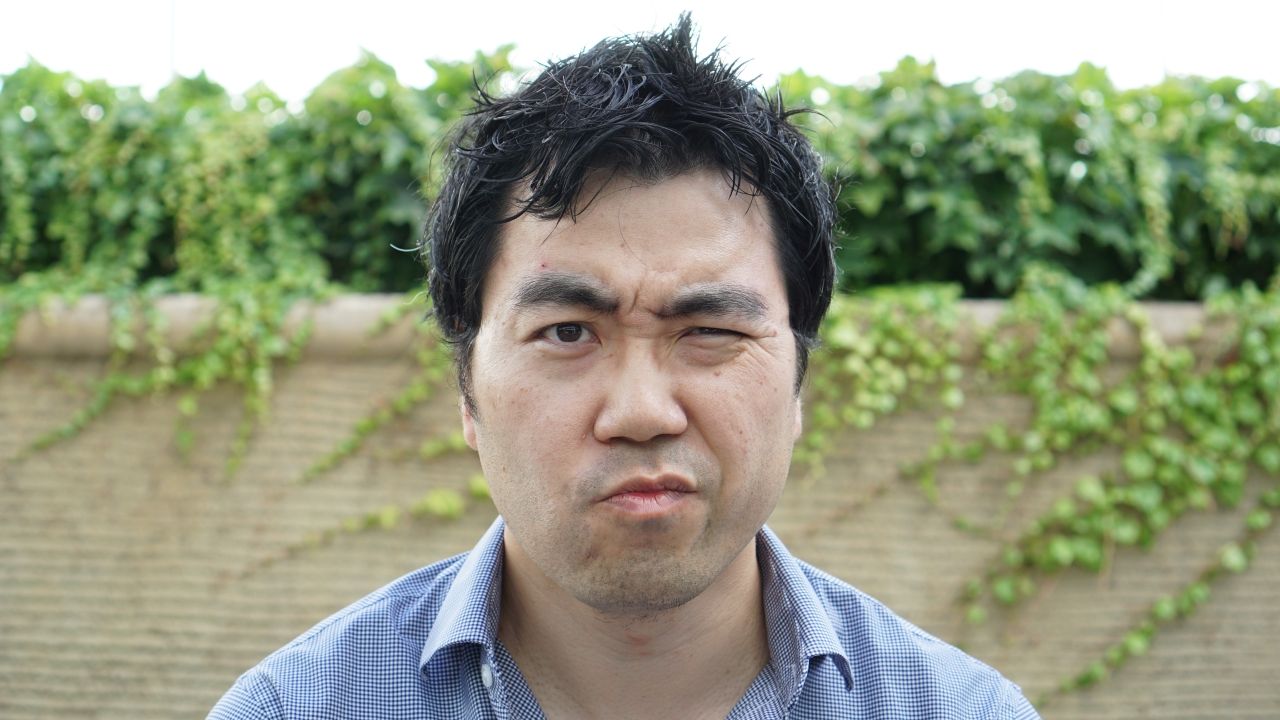}} \\
			\subfloat[]{\includegraphics[width=.32\columnwidth]{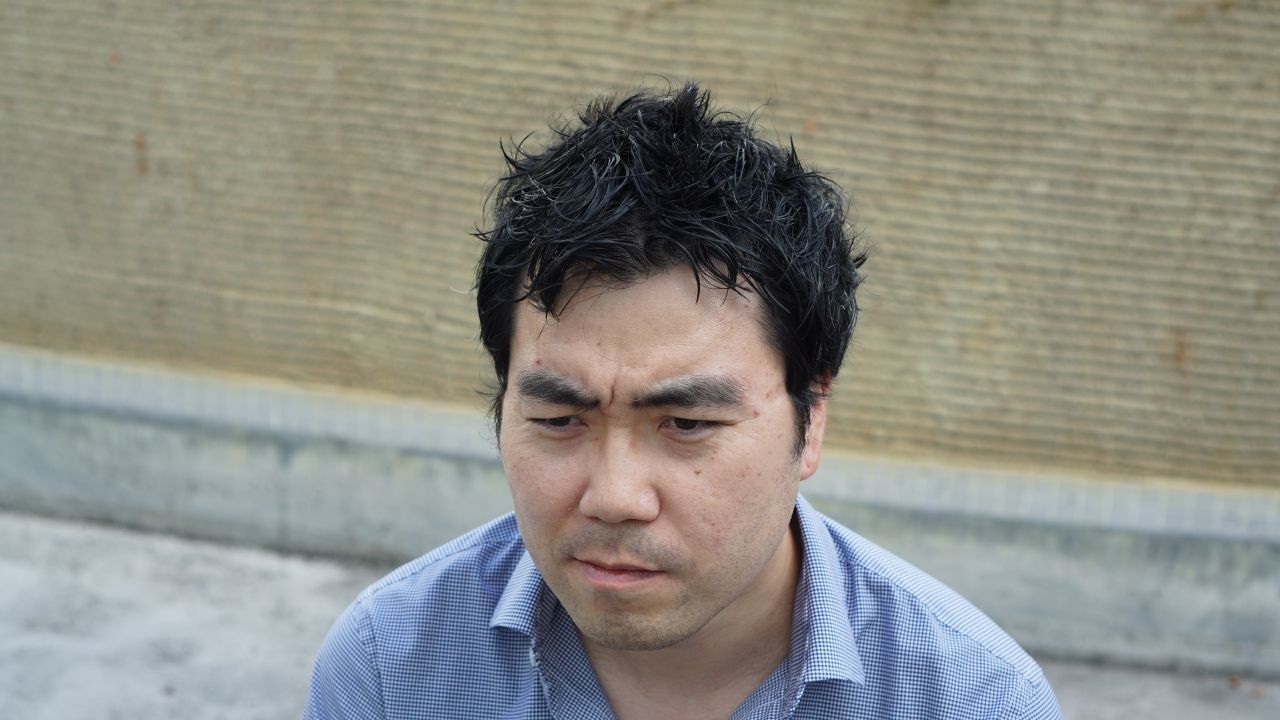}}
			\hfill
			\subfloat[]{\includegraphics[width=.32\columnwidth]{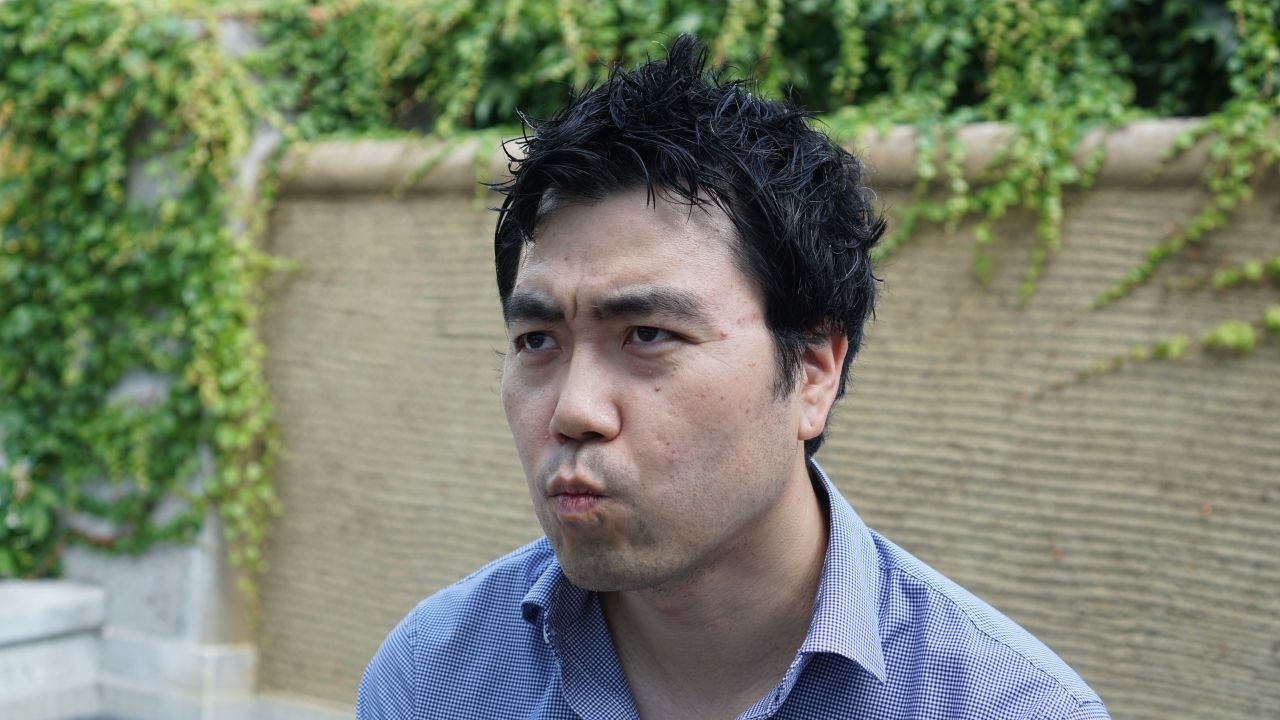}}
			\hfill
			\subfloat[]{\includegraphics[width=.32\columnwidth]{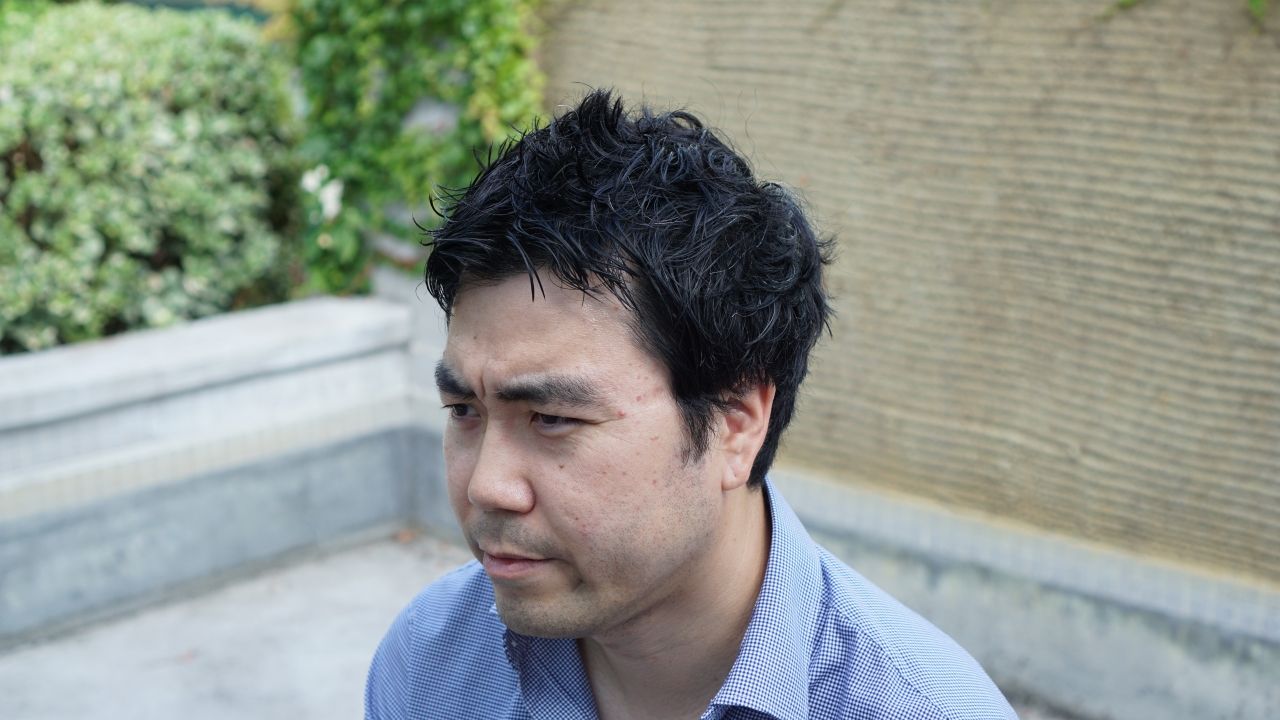}}
			\caption{Input images for the face example}
			\label{fig:kihwan}
		\end{figure}

		In Table~\ref{tab:performance}, we report the run-time for the face example (Fig.~\ref{fig:kihwan}) with 100 deformation nodes, depending on the number of iterations $N$ used for the sparse correspondence association.
		
		\begin{table}[h]
			\centering
			\begin{tabular}{c|c|c}
				\textbf{Step} & \textbf{Time} ($N=1$) & \textbf{Time} ($N=5$) \\ 
				\hline 
				Total & 154 min & 422 min \\ 
				\hline 
				Filter & 0.2\% & 0.4\% \\ 
				Optimize & 92.3\% & 96.9\% \\ 
				Depth & 7.5\% & 2.7\% \\ 
			\end{tabular} 
			\caption{Computation time (in minutes) needed for different steps to process the example in Fig.~\ref{fig:kihwan} depending on the number of iterations $N$ (see main paper for more details): Filtering of sparse correspondences, Joint hierarchical optimization and depth estimation; 
				file I/O not included.
			}
			\label{tab:performance}
		\end{table}
		
		\subsection{Camera Pose Estimation}
		\label{sec:pose}
\begin{figure}[h]
			\centering
			\includegraphics[width=\columnwidth]{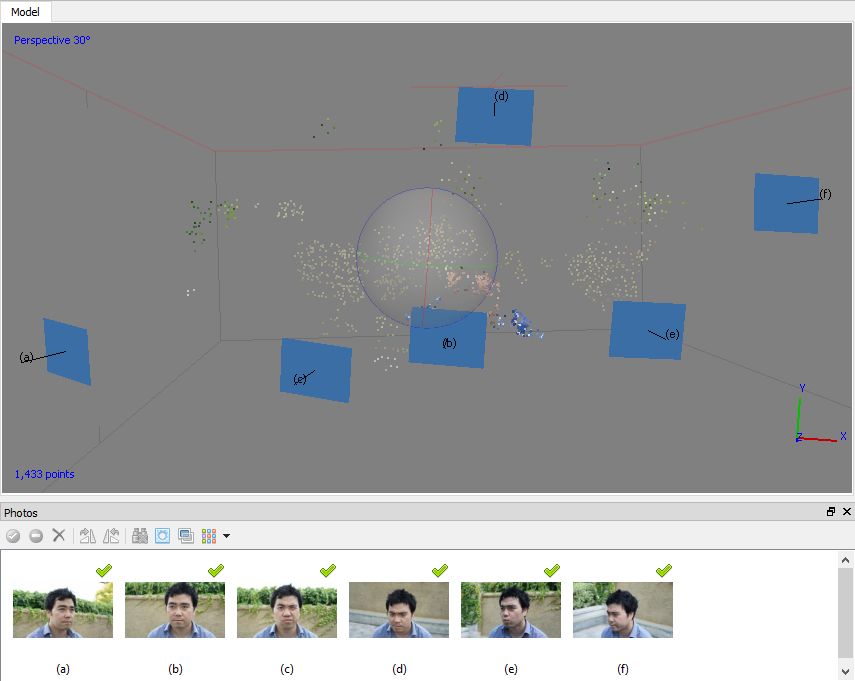}
			\caption{Screenshot of Agisoft PhotoScan}
			\label{fig:camera_pose}
		\end{figure}

		In Fig.~\ref{fig:camera_pose}, we show an example result for the estimated camera poses using Agisoft PhotoScan~\cite{photoscan}.
		As can be seen, the camera poses for the input images have been successfully recovered.
		
		\subsection{Canonical View Selection}
		\label{sec:canonical}
		To pick the canonical views, we analyze how many matches result in a faithful static reconstruction. 
		I.e., we triangulate each match (after doing the ratio test with $r~=~0.7$~\cite{lowe1999object}) and reject those with a reprojection error above 1 pixel.
		As can be seen in Table~\ref{tab:canon_view_select}, the image pair (a)-(b) dominates the ratio of static inliers.
		Therefore, our algorithm chooses these views to reconstruct the initial canonical surface.
		
		\begin{table}
			\begin{tabular}{cccccc}
				& (b) & (c) & (d) & (e) & (f) \\ 
				(a)  & \textbf{10.3\%} & 0.42\% & 0.00\% & 0.00\% & 0.83\% \\ 
				(b)  &  & 0.17\% & 0.24\% & 1.14\% & 0.00\% \\ 
				(c)  &  &  & 0.00\% & 0.00\% & 0.65\% \\ 
				(d)  &  &  &  & 0.28\% & 0.88\% \\ 
				(e)  &  &  &  &  & 0.26\% 
			\end{tabular} 
			\caption{Confusion matrix for static inlier ratio for all 2-view combinations (see Fig.~\ref{fig:kihwan}).}
			\label{tab:canon_view_select}
		\end{table}
		
		\section{Additional Results}
		\label{sec:results}		
		
\begin{figure}[h]
			\centering
			\subfloat[COLMAP]{\includegraphics[trim=11cm 6cm 8cm 5cm, clip=true,width=.49\columnwidth]{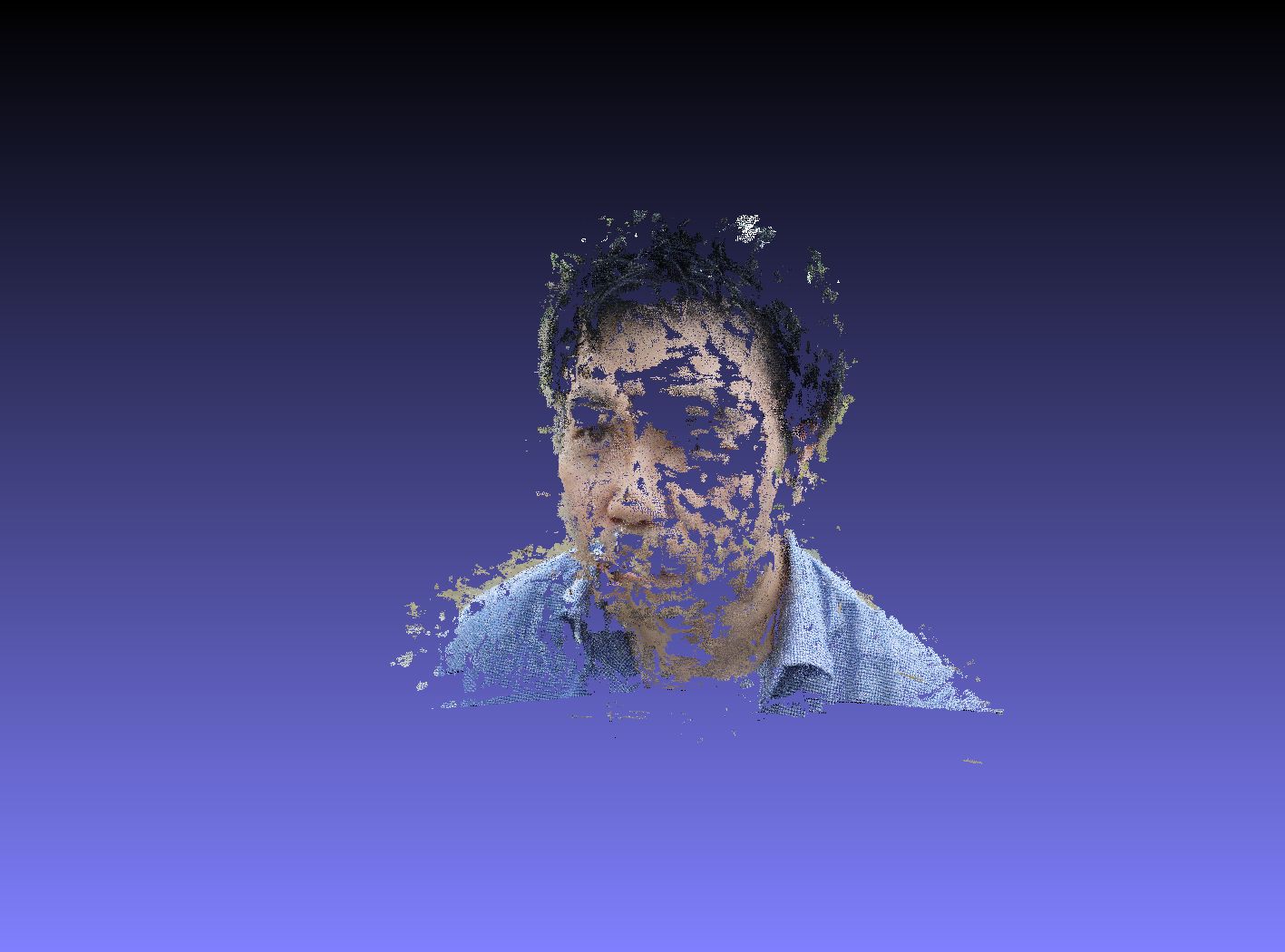}}
			\hfill
			\subfloat[Ours]{\includegraphics[trim=11cm 6cm 8cm 5cm, clip=true,width=.49\columnwidth]{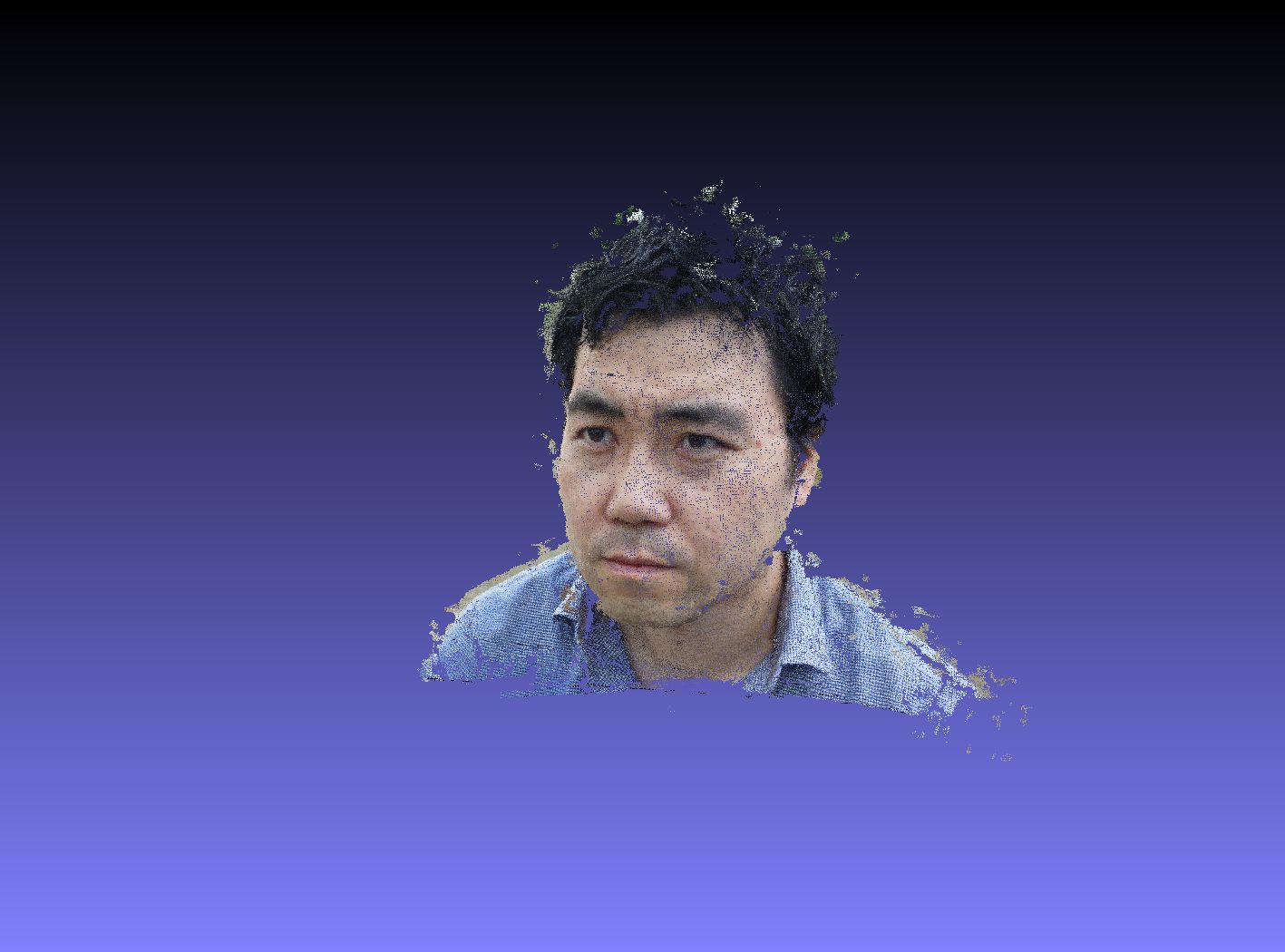}}
			\caption{Comparison with COLMAP \cite{schoenberger2016mvs}: a state-of-the-art MVS algorithm for static scenes fails to reconstruct images undergoing non-rigid motion (Fig.~\ref{fig:kihwan}).}
			\label{fig:comp_colmap}
		\end{figure}

		In Fig.~\ref{fig:comp_colmap}, we show an example result comparing our algorithm with a state-of-the-art MVS approach that performs best in a recent survey \cite{Schoeps2017CVPR}.
		As can be seen, the geometry of the deforming region can not be reconstructed successfully, if the method assumes static geometry.
		
	\end{appendices}
	
\end{document}